\documentclass[article,authoryear,jmlmc]{beg_32}

\usepackage[hang]{footmisc}
\fancypagestyle{plain}{
  \fancyhf{}
  \fancyhead[R]{\small {\it \jname}}
  \fancyfoot[R]{\small\bf\thepage}
}

\renewcommand{\dmyy}{26}
\renewcommand{\myyear}{2026}
\renewcommand{\today}{}

\newcommand\datastatement{\section* {DATA AVAILABILITY}}

\usepackage{booktabs}
\usepackage{mathtools}
\usepackage{soul}
\usepackage{color}
\usepackage{cuted}
\usepackage{makecell}
\usepackage{multicol}
\usepackage{placeins}
\usepackage{algorithmic}
\usepackage{pifont}
\usepackage{url}

\usepackage{lineno}

\SetKwComment{Comment}{$\triangleright\;$}{}

\begin{document}
\raggedbottom

\newcommand{\rLtwo}{Rel.$\,L_2\,$}
\newcommand{\overLtwo}{Overall Rel.$\,L_2\,$}
\newcommand{\E}{\mathbb{E}}

\volume{}
\title{Diagnosing Epistemic Uncertainty in Neural Operators: Effects of Initialization, Discretization, and Function Sampling}
\titlehead{Diagnosing Epistemic Uncertainty in Neural Operators}
\authorhead{T.Y. Yen, Y. Fu, P. Roy, G. Lin, \& M. Penwarden}

\corrauthor[1]{Tian Yu Yen}
\author[2]{Yucheng Fu}
\author[3]{Pratanu Roy}
\author[4]{Guang Lin}
\author[1,5]{Michael Penwarden}

\corremail{tyen@sandia.gov}
\corraddress{Sandia National Laboratories, Albuquerque, New Mexico, USA}
\address[1]{Sandia National Laboratories, Albuquerque, New Mexico, USA}
\address[2]{Pacific Northwest National Laboratory, Richland, Washington, USA}
\address[3]{Lawrence Livermore National Laboratory, Livermore, California, USA}
\address[4]{Purdue University, West Lafayette, Indiana, USA}
\address[5]{University of Utah, Salt Lake City, Utah, USA}

\dataO{}
\dataF{}

\abstract{
Neural operators provide efficient surrogates for mappings between function spaces, but their reliability cannot be characterized by the performance of a single trained model. Existing neural operator uncertainty quantification approaches often characterize predictive uncertainty without separately identifying the contributions of individual sources of epistemic uncertainty. Finite sampling of the training function space, output discretization, and model initialization can produce substantial variation among learned operators even when the underlying physical problem is deterministic. We systematically investigate these sources of epistemic uncertainty across a diverse set of operator learning architectures. Across our experiments, we train approximately 46,000 neural operator models and construct ensembles over training datasets, model initializations, and output discretizations to characterize distributions of relative \(L_2\) error rather than individual trained realizations. We find substantial architecture-dependent differences in epistemic uncertainty, with model initialization often accounting for the largest proportion of variance in relative \(L_2\) error. The ranked performance of neural operator architectures can invert as the number and spatial distribution of output samples change, with the best architectures under sparse or randomly sampled observations differing from those at higher-resolution or structured discretizations. Furthermore, ensembles that reliably characterize aggregate operator error can provide unreliable estimates of error quantiles for individual test functions. These results highlight the importance of distinguishing the sources of epistemic uncertainty when evaluating neural operator reliability and establishing robust benchmarking practices.}


\maketitle

\newpage
\section{Introduction}
\label{sec:introduction}
Neural operators learn mappings between function spaces and have become increasingly useful as surrogate models for partial differential equations and other parameterized scientific systems \citep{kovachki2023neural,lu2021learning}. Once trained, these models can evaluate new inputs substantially faster than conventional numerical solvers, making them attractive for applications that require repeated forward evaluations, including design optimization  \citep{shukla2024deep,zang2026design,kou2025neural,ramezankhani2025advanced}. However, the performance of a neural operator is not uniquely determined by its architecture. Different samples of the training function space, different discretizations of those functions, and stochasticity in model training can produce substantially different learned operators for the same underlying problem. Consequently, benchmarking an architecture using a single trained realization and a fixed sampling configuration can obscure both the variability of its performance and regimes where its relative performance compared to other architectures changes. Prior work has demonstrated that similar average errors across sampled tasks can conceal substantial differences in worst-case approximation accuracy, leading to misleading comparisons of neural architectures
\citep{penwarden2024kolmogorov}.

These effects are especially relevant when neural operators are embedded within downstream decision-making. In design optimization, quantities of interest derived from predicted solution fields are used to compare candidate designs, yet uncertainty in those quantities of interest is often not considered explicitly. A design with the most favorable predicted objective may lie in a region where the surrogate is highly uncertain, while a slightly less favorable design may be supported by substantially more reliable predictions. Epistemic uncertainty is therefore relevant not only to the predicted field, but also to the quantities of interest that ultimately determine the design decision. Similar considerations arise when selecting the neural operator architecture itself. An architecture that achieves the lowest error under favorable training conditions may also be brittle, exhibiting large variability or complete training failures in low-data or poorly resolved regimes. Another architecture may have a lower ceiling in best-case accuracy while producing substantially more consistent models across training datasets, initializations, and discretizations. Without benchmarking these distributions, architecture selection can therefore favor a model based on an unusually successful realization or a sampling regime that is not representative of the intended application. A more informative benchmark should therefore characterize the distribution of model performance across the training and sampling conditions relevant to the intended application, rather than reporting a single realization at a single discretization.

Assessing neural operator reliability requires distinguishing
between two questions: (i) how accurately an operator performs
on average across the test distribution, and (ii) how accurately
it predicts a specific test function. The first characterizes uncertainty in aggregate operator performance, while the second characterizes uncertainty in the approximation error for a specific input function. Reliable estimates of the former do not necessarily imply reliable estimates of the latter.

A growing body of work has developed uncertainty quantification (UQ) methods for neural operators, including Bayesian DeepONets \citep{LIN2023111713,garg2023vb}, randomized-prior ensembles \citep{yang2022randomized}, ensemble Kalman inversion \citep{pensoneault2025eki}, robust Bayesian formulations \citep{lone2026alpha}, conformal prediction \citep{moya2025conformalized}, and approximate Bayesian treatments of Fourier and more general neural operators \citep{weber2024uncertainty,magnani2025linearization}. These approaches primarily characterize the total predictive uncertainty after a neural operator has been trained, i.e., the sum of epistemic sources and aleatoric sources combined \citep{psaros2023uqsciml}. Here, we instead examine how different sources of epistemic variability introduced during training affects the resulting learned operator. We focus on finite sampling of the input-function distribution, discretization and sampling of the observed output functions, and stochasticity associated with model training. These sources are distinct from aleatoric uncertainty, which describes variability remaining after conditioning on the information available to the model \citep{hullermeier2021aleatoric}.

We investigate these effects across a diverse set of operator-learning architectures, including Deep Operator Networks (DeepONets) \citep{lu2021learning}, DeepONets using shallow universal polynomial network (SUPN) trunks \citep{morrow2025supn}, Fourier neural operators (FNOs) \citep{li2021fourier,kovachki2023neural}, and kernel neural operators (KNOs) \citep{lowery2026kernel}. Across the study, we train approximately 46,000 neural operator models over different training datasets, model initializations, output discretization levels, and sampling strategies. The experiments use operators for which reference solutions can be evaluated directly, allowing variability in the learned surrogate to be studied without introducing uncertainty from an approximate reference solver.


The principal findings of this study are:
\begin{itemize}
    \item{\textbf{Epistemic uncertainty is architecture-dependent.}
    Model initialization often accounts for the largest proportion of variance in \overLtwo, although its contribution depends on architecture and output discretization.}
    \item{\textbf{Overall performance does not imply reliable individual predictions.}
    Ensembles can accurately characterize aggregate operator error while providing unreliable uncertainty estimates for individual input functions.}
    \item \textbf{Output discretization changes architecture rankings.}
    Architectures that perform best on dense, structured grids may be less accurate under sparse discretizations, while randomized output sampling can improve accuracy
    and consistency with the right architecture.
    \item{\textbf{Ensembles only over initializations can underestimate sources of uncertainty.}
    Reliable uncertainty quantification requires accounting for training dataset variability, particularly when assessing individual predictions.}
\end{itemize}

These experiments are intended as a diagnostic study of epistemic uncertainty in neural-operator training and benchmarking rather than as a new predictive uncertainty-quantification method. The resulting distributions provide a basis for interpreting neural-operator benchmarks and for determining which sources of epistemic uncertainty should be represented when neural operators are used in downstream scientific and engineering applications.

\section{Background}
\label{sec:background}

\subsection{Operator Learning Architectures}
Neural operators approximate mappings between function spaces, but their inputs and outputs must be represented through finite discretizations. Although neural operators can be evaluated on discretizations different from those used during training, their accuracy still depends on the resolution and distribution of the sampled points. The following architectures handle these discretizations differently, motivating their comparison in this study.

\subsubsection{Deep Operator Networks (DeepONets)}

Deep Operator Networks (DeepONets, sometimes abbreviated DONs in this paper) approximate nonlinear operators using separate branch and trunk networks \citep{lu2021learning}, following the operator universal approximation framework of \citet{10.1109/72.392253}. Let
\begin{align}
\mathcal{G}:\mathcal{U}\rightarrow\mathcal{V}
\end{align}
denote the target operator, where $\mathcal{U}$ and $\mathcal{V}$ are spaces of input and output functions respectively. For an input function $u\in\mathcal{U}$, the corresponding output is $v=\mathcal{G}(u)\in\mathcal{V}$. The input function is represented by its values at fixed sensor locations $\{\mathbf{x}_i\}_{i=1}^{m}$,
\begin{align}
\mathbf{u}
=
[u(\mathbf{x}_1),\ldots,u(\mathbf{x}_m)]^\top.
\end{align}

The branch network $\mathcal{B}_{\theta_B}$ maps these sampled input values to a vector of $N$ coefficients, while the trunk network $\mathcal{T}_{\theta_T}$ maps an output coordinate $\mathbf{y}$ to $N$ basis values. The DeepONet prediction at $\mathbf{y}$ is their inner product,
\begin{align}
\mathcal{G}(u)(\mathbf{y})
&\approx
\widehat{\mathcal{G}}_{\theta_B,\theta_T}(u)(\mathbf{y}) =
\left\langle
\mathcal{B}_{\theta_B}(\mathbf{u}),
\mathcal{T}_{\theta_T}(\mathbf{y})
\right\rangle =
\sum_{n=1}^{N}
[\mathcal{B}_{\theta_B}(\mathbf{u})]_n
[\mathcal{T}_{\theta_T}(\mathbf{y})]_n.
\end{align}

Here, $\theta_B$ and $\theta_T$ denote the trainable parameters of the branch and trunk networks, respectively. Because the trunk network is evaluated directly at $\mathbf{y}$, the predicted output can be evaluated at coordinates that do not coincide with the output locations used during training.

\subsubsection{DeepONets with a SUPN Trunk}
DeepONet does not require the trunk or branch networks to be an MLP. Alternative trunk representations can be used such as POD-DeepONet which replaces the learned trunk with basis functions obtained from proper orthogonal decomposition \citep{lu2022comprehensive}. In this work, we instead replace the standard MLP trunk with a Shallow Universal Polynomial Network (SUPN) \citep{morrow2025supn}. SUPNs replace deep hidden layers with a shallow network built from learned polynomial expansions followed by nonlinear activation. This reduces the number of trainable parameters and the size of the optimization problem, and \cite{morrow2025supn} reports lower sensitivity to initialization and lower variability than standard deep networks on the tested approximation problems.

For a multidimensional coordinate $\mathbf{y}$, the SUPN polynomial expansion is built from tensor-product Chebyshev polynomials indexed by a multi-index set $\Lambda\subset\mathbb{N}_0^D$. For the DeepONet trunk, we use
\begin{align}
\mathcal{T}_{\theta_T}(\mathbf{y})
=
\sum_{r=1}^{R}
\mathbf{c}_r
\tanh\left(
\sum_{\mathbf{m}\in\Lambda}
a_{r,\mathbf{m}}
T_{\mathbf{m}}(\widetilde{\mathbf{y}})
\right),
\qquad
\mathbf{c}_r\in\mathbb{R}^{N},
\end{align}
where
\begin{align}
T_{\mathbf{m}}(\widetilde{\mathbf{y}})
=
\prod_{d=1}^{D}
T_{m_d}(\widetilde{y}_d).
\end{align}
Here, $\widetilde{\mathbf{y}}$ denotes the output coordinate rescaled to the canonical Chebyshev domain, $T_{m_d}$ is the one-dimensional Chebyshev polynomial of degree $m_d$, and $\mathbf{c}_r$ produces the $N$-dimensional trunk representation required by the DeepONet inner product. We use the SUPN trunk to test whether an architecture variant can meaningfully change the epistemic uncertainty of DeepONets. We refer to \citet{morrow2025supn} for the full SUPN construction and implementation details.

\subsubsection{Fourier Neural Operators}

Fourier neural operators (FNOs) learn operators by applying nonlocal transformations in Fourier space \citep{li2021fourier,kovachki2023neural}. The input function is first lifted pointwise to a higher-dimensional latent representation. Each Fourier layer applies a Fourier transform, multiplies a retained set of Fourier modes by learned complex-valued weights, transforms the result back to physical space, and combines it with a learned pointwise linear transformation. A final pointwise network maps the latent representation to the output field.

A Fourier layer can be written as
\begin{align}
v^{(\ell+1)}(\mathbf{x})
=
\sigma\left(
W^{(\ell)}v^{(\ell)}(\mathbf{x})
+
\mathcal{F}^{-1}
\left[
R^{(\ell)}
\mathcal{F}\left(v^{(\ell)}\right)
\right](\mathbf{x})
\right),
\end{align}
where $R^{(\ell)}$ contains the learned weights applied to the retained Fourier modes and $W^{(\ell)}$ is a learned pointwise linear transformation. Standard FNOs evaluate these spectral operations on structured grids, so their behavior depends on grid resolution and the retained Fourier modes. For randomly sampled outputs, the scattered observations must be converted to the structured grid required by the standard FNO; therefore, the resulting error reflects both the learned operator approximation and the interpolation used to construct that grid. We describe this procedure in Section~\ref{sec:fno_kno_interpolation}.

\subsubsection{Kernel Neural Operators (KNO)}

Kernel neural operators (KNOs) learn operators using explicit kernel integral transformations rather than Fourier convolution \citep{lowery2026kernel}. The input function is first lifted to a higher-dimensional latent representation, after which repeated kernel layers combine a learned pointwise linear transformation with a nonlocal integral operator. A kernel layer can be written as
\begin{align}
v^{(\ell+1)}(\mathbf{x})
=
\sigma\left(
W^{(\ell)}v^{(\ell)}(\mathbf{x})
+
\int_{\Omega}
K_{\theta_K}^{(\ell)}(\mathbf{x},\mathbf{z})
v^{(\ell)}(\mathbf{z})\,d\mathbf{z}
\right),
\end{align}
where $K_{\theta_K}^{(\ell)}$ is a learned matrix-valued kernel and $W^{(\ell)}$ is a learned pointwise linear transformation.

The kernel integral is evaluated numerically using quadrature. Unlike standard FNOs, KNOs are not restricted to a uniform Fourier grid because the quadrature points can be chosen independently of the observation locations. In our implementation, interpolation is still required when the locations at which the latent function is known differ from the quadrature or output locations used by the kernel layer. This differs from the FNO case, where interpolation is required specifically to construct the structured grid needed for the Fourier transform. We describe the interpolation procedures used for both FNOs and KNOs in Section~\ref{sec:fno_kno_interpolation}.

We include KNOs because they provide a nonlocal operator architecture with a different treatment of discretization than FNOs. This allows us to study how model performance changes when the output functions are sampled on regular or random point sets. In this work, we use the nonstationary generalized spectral mixture (NSGSM) kernel. We refer to \citet{lowery2026kernel} for the full continuous and discretized KNO formulations and the NSGSM kernel construction.

\subsection{Sources of Uncertainty and Error in Neural Operators}

Uncertainty in neural operators can arise from several sources, and the distinction between aleatoric and epistemic uncertainty depends on the information
available to the model \citep{gruber2023sources, hullermeier2021aleatoric}. Epistemic uncertainty arises from incomplete knowledge and may be reduced through additional data, denser observations, or improved training. Aleatoric uncertainty describes variability
that remains after conditioning on the available information, such as unresolved stochastic forcing or measurement noise.  The classification therefore depends on what is included in the neural operator input. For example, a stochastic coefficient that is explicitly provided as an input may still define a deterministic solution operator, while unresolved stochastic forcing or measurement noise may remain aleatoric.

Table~\ref{tab:uq_components} separates the sources that affect neural operator predictions into bias, epistemic variance, and aleatoric variance. In this work, we focus on three reducible sources of epistemic error and variability: finite sampling of the operator space, discretization and sampling of the function spaces, and stochasticity in model training. Finite sampling changes which input-output function pairs are available during training. Discretization limits the information available about each function, while random sampling of the domain also changes where that information is observed. Training stochasticity includes parameter initialization, mini-batch ordering, and other randomness in the optimization procedure. The bias terms in Table~\ref{tab:uq_components} should be interpreted separately from this variability.

\begin{center}
\captionof{table}{Uncertainty Quantification Components of Neural Operator Training}
\label{tab:uq_components}
\begin{tabular}{ccc}
\toprule
\textbf{Type} & \textbf{Quantity} & \textbf{Source} \\
\bottomrule

\multirow{3}{*}{Bias}
& Approximation Error & Model Expressivity  \\\cline{2-3}
& Optimization Suboptimality & \makecell{Not at Global Optima} \\\cline{2-3}
& Problem Formulation & \makecell{Ill-Posedness / Non-Uniqueness} \\\cline{1-3}

\multirow{7}{*}{\makecell{Epistemic \\ (variance)}}
& Sampling of Operator Space & Dataset Generation \\\cline{2-3}
& \multirow{2}{*}{Discretization of Input Space} & Training Set \\\cline{3-3}
& & Testing Set \\\cline{2-3}
& \multirow{2}{*}{Discretization of Output Space} & Training Set \\\cline{3-3}
& & Testing Set \\\cline{2-3}
& \multirow{2}{*}{Training Stochasticity} & Parameter Initialization \\\cline{3-3}
& & Mini-Batches \\\cline{1-3}

\multirow{3}{*}{\makecell{Aleatoric \\ (variance)}}
& Unresolved Stochasticity & Unobserved Random Inputs \\\cline{2-3}
& Latent Variability & Unobserved Random Processes \\\cline{2-3}
& \multirow{2}{*}{Data Quality} & Noise \\\cline{3-3}
& & Simulation Fidelity \\
\bottomrule
\end{tabular}
\end{center}

\section{Methodology}


\subsection{Metrics of Performance and Analyzing Uncertainties}
In this Section, we briefly mathematically formalize and discuss the two types of uncertainty questions posed in the introduction:
\begin{enumerate}
    \item How well does the trained operator perform in general, across many of the examples upon which the operator will be deployed?
    \item How will the trained operator perform on a specific example or use case?
\end{enumerate}
This formalization will lead to our approach to examining the sources of uncertainty involved in operator learning.

We measure approximation error using the relative
$L_2$ norm,
\begin{align}\label{eq:relL2}
\text{\rLtwo}&= \frac{||\hat{v} - v||_{L_2}}{||v||_{L_2}}
\end{align}
where $v=\mathcal{G}(u)$ is the target function and $\hat{v}$ its approximation. While the relative $L_2$ error is appropriate for assessing global solution accuracy, application-specific quantities of interest may require more localized error metrics.

We examine variation in \rLtwo arising from model initialization, finite training function sampling, and output discretization. Let $\theta_0\sim \mathcal{D}_\Theta$ denote the random initialization of a learned operator $\mathcal{G}_\theta$, and let $S_n=\{(u^i,v^i)\}_{i=1}^{n}$ be a training dataset drawn from the input-function distribution $\mathcal{D}_\mathcal{U}$, with $v^i=\mathcal{G}(u^i)$. The expectations $\E_{\theta_0}[\text{\rLtwo}]$ and $\E_{S_n}[\text{\rLtwo}]$ characterize the effects of initialization and training dataset sampling, respectively. Both affect the approximation $\hat{v}$ through training, whereas the reference solution $v$ remains unchanged.

The effect of structured domain discretizations is not a random source of uncertainty.
However, we hypothesized that, contrary to the promises of super-resolution, sparse discretizations would magnify the effects of other stochastic uncertainties in the performance of trained operators.
Note, that we assumed in all cases that the input functions $u$ were well-resolved: only the output functions $v$ suffer from limited available information.
This assumption is motivated by design optimization applications where controlled inputs are well-described and a significant cost is involved in collecting fine-scale evaluations of output functions (i.e., costly finite-element solves).

Alternatively, we examined the effect of random domain discretizations of the output training functions $v$.
In this case, we assume each function $v^i$ is randomly sampled at points in its domain, $y^i_{1},\ldots,y^{i}_{D}\sim\mathcal{D}_\mathcal{Y}$, where $\mathcal{Y}$ is the domain of functions $v\in\mathcal{V}$.
Thus we describe the joint effect of the sampling of functions and random discretizations as $\E_{S_n\times Y_D}$.

For any given $u\in\mathcal{U}$, we measure the specific \rLtwo (and its corresponding uncertainties) with equation \eqref{eq:relL2}.
However, if we want to compare the overall performance of a model across all examples, then we should compute the expected value of \rLtwo with respect to the distribution $\mathcal{D}_\mathcal{U}$.
In other words,
\begin{align}\label{eq:ovrelL2}
    \text{\overLtwo} = \E_v\left[\frac{||\hat{v}-v||_{L_2}}{||v||_{L_2}}\right], \; v \sim \mathcal{D}_{\mathcal{G}(U)}
\end{align}
where $\mathcal{D}_{\mathcal{G}(U)}$ is the push-forward of $\mathcal{D}_U$ through operator $\mathcal{G}$.

With this formalism, we can ask questions such as:
\begin{enumerate}
    \item What is the probability that the \overLtwo is less than some criteria $\delta$ given a specific training set $S_n$?
\begin{align}
    \Pr_{{\theta_0}\sim{\mathcal{D}_\Theta}}[\ \E_v\left[\text{\rLtwo}\right]<\delta\ \mid S_n]
\end{align}    
    \item Given a specified function $v=\mathcal{G}(u)$ for $u\in\mathcal{U}$, what is the probability that the average model initialization will have \rLtwo less than $\delta$ when trained on a randomly selected training set (with fixed domain discretization)?
\begin{align}
    \Pr_{S_n \sim \mathcal{D}_{S}} [\  \E_{\theta_0}\left[ \text{\rLtwo} \right] <\delta\ \mid v]
\end{align}    
where $\mathcal{D}_{S}$ is the distribution of a set of $n$ randomly sampled functions.
\end{enumerate}
\begin{figure}[H]
    \centering
    \centering
    \includegraphics[width=.8\linewidth]{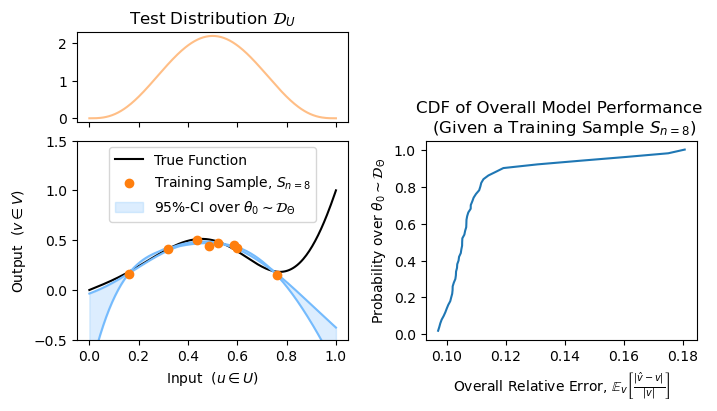}
    \caption{Shows a 95\% confidence interval from an ensemble of model initializations trained on the same training sample. The overall performance of each model, after training, is computed by taking the expectation of the relative error with respect to the test distribution (top). The empirical CDF of the expected relative error over the ensemble of models is then computed (right).}
    \label{fig:uqA}
\end{figure}
\begin{figure}[h]
    \centering
    \includegraphics[width=0.8\linewidth]{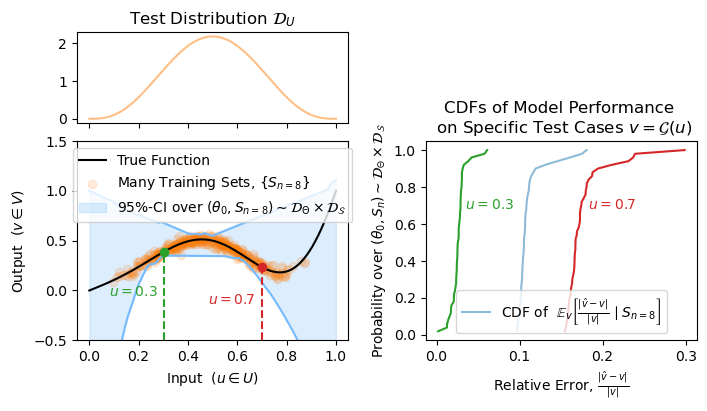}
    \caption{Shows a 95\% confidence interval from an ensemble of model initializations trained on multiple training samples. The average relative error is computed with respect to the ensemble of models at $u=0.3$ (green) and $u=0.7$ (red). The empirical CDF of this average relative error is then computed over the ensemble of datasets at $u=0.3$ and $u=0.7$ (right). Note that the point-wise probabilities provide significantly different information than the overall performance of an ensemble models trained on a single dataset (blue).}
    \label{fig:uqB}
\end{figure}

Figures~\ref{fig:uqA} and~\ref{fig:uqB} illustrate the distinction using ensembles of neural networks. In Figure~\ref{fig:uqA}, models trained on the same eight-point dataset achieve low aggregate error with high probability, despite larger errors near the boundaries of the input domain where observations are less frequent. Figure~\ref{fig:uqB} instead examines errors for specific inputs across different training datasets. At $u=0.3$, the average model is likely to have relatively small error, whereas at $u=0.7$, errors exceeding 15\% are common. These differences illustrate why aggregate error distributions can conceal substantial uncertainty for individual predictions, particularly where training samples insufficiently resolve the target function.

We consider operator learning problems with analytically
evaluated reference solutions, allowing us to evaluate
prediction errors without introducing numerical solver
errors. To characterize epistemic uncertainty, we train
ensembles across independently sampled training datasets
and model initializations. We analyze the resulting error
distributions using summary statistics and cumulative
distribution functions (CDFs), estimated through kernel
density estimation (KDE) and numerical integration.

\subsection{Proportion of Variance}\label{sec:variance_decomp}
To quantify the relative contributions of model initialization and finite sampling of the function space to variation in \overLtwo, we perform a sum-of-squares variance decomposition. Let $\mu_\text{\overLtwo}$ denote the mean \overLtwo across all model initializations and training datasets. For each factor, we compute
\begin{align}
    SS_{\text{factor}} = n_{i\neq j}\sum_j \left(\mu_{j}-\mu_\text{\overLtwo}\right)^2
\end{align}
where $\mu_j$ are the conditional means keeping the contributing factor $j=1,\ldots,n_{\text{factor}}$ constant and $n_{i\neq j}$ is the number of samples used to compute the conditional mean $\mu_j$.
We then compared the SS of each factor compared to the total sum of squares to determine the proportion of variance in \overLtwo explained by finite training datasets versus model initializations.
This procedure is equivalent to computing the eta-squared value in an analysis of variance (ANOVA), with model initialization and training dataset as factors.

\subsection{FNOs and KNOs Interpolation Error on Random Discretizations}
\label{sec:fno_kno_interpolation}

To accommodate random output discretizations, we interpolate sampled function values onto the structured grids of the highest fidelity as required by FNOs, following the approach of \cite{Lanthaler2024-gf}. For KNOs, linear interpolation maps predicted values from quadrature points to the randomly sampled output locations. This necessarily introduces interpolation errors, which we acknowledge as additional contribution to the approximation error.

\section{Computational Experiments}
\label{sec:numerical_results}

\subsection{Nonlinear Transform on Polynomials}\label{sec:poly1d}
\subsubsection{Experiment Setup}
Consider the following operator T(u)
\begin{align}
    T(u) = \alpha \frac{du}{dx} + \beta u^2
\end{align}
with $\alpha=0.5$ and $\beta=2$. For this problem, we assume that the input function domain $\mathcal{U}$ is $\mathcal{U}:=\{u\in P_5[-1,1]: \|u\|_{L_2}=1\}$. Then the operator $T(u)$ maps polynomials $u(x)$ of degree at most 5 onto polynomials $v(x)$ of degree at most 10.

\textbf{Discretization}: we assume that $n_D=150$ evenly spaced points sufficiently discretizes the input and output spaces. We assume we have complete knowledge of the input space, and thus discretize all input functions with the maximum number of points. We then examine the effect of discretizing the output space with $n_D=\{10,25,50,75,100,125,150\}$ points.
Alternatively, we compare the effect of independent random sampling $n_D$ points uniformly from $[-1,1]$. That is, for each function $u$, we randomly sample a new set of $n_D$ points for that function. We hypothesized that this might improve training at low fidelity sampling rates because every training function would contain information from different randomly selected parts of the domain.
As a limitation, we note that FNOs require a uniform output discretization for computing the fast Fourier transform.
Thus, for FNOs, we first perform linear interpolation of the random points onto a uniform sampling grid of size $n_D$ and train the FNO to fit approximate function values on the interpolated uniform grid.
For KNOs, linear interpolation was applied to interpolate the predicted output function values from the quadrature points to the randomly sampled points.
In lower sample regimes, we expected these approximations to contribute significantly to the \overLtwo error.

\textbf{Dataset Generation}: we investigate the effect of finite sampling the function space $P_5[-1,1]$. We sample $n_s=250$ functions from $\{u\in P_5[-1,1]: \|u\|_{L_2}=1\}$ using Legendre polynomials as the basis and uniformly sampled coefficients from the unit sphere (normalized appropriately such that $\|u\|_{L_2}=1$). We generate $k=1,\ldots,50$ such datasets, training each model on each dataset. Thus, each training dataset can be described as $\mathcal{D}_{k,n_D}=\{(u^i_{150},v^i_{n_D})\}$, where $k$ is the dataset number and $n_d$ is the discretization condition described above.

\textbf{Model Initialization:} for each model type we initialize the parameters of $n_m=15$ models using standard randomization procedures for the parameters of each operator type. Initializations were saved so that each initialization could be trained independently on each dataset $\mathcal{D}_{k,n_d}$, $k=1,\ldots,50$.

\textbf{Training Procedures:} All architectures were trained
using Adam followed by L-BFGS refinement under a common
optimization protocol. The complete training configuration,
parameter counts, and computational costs are provided in
~\ref{app:poly_training}
(Table~\ref{tab:poly1d-times}).

\subsubsection{Overall Performance Results}
\begin{figure}[H]
    \centering
    \includegraphics[width=1\linewidth,keepaspectratio]{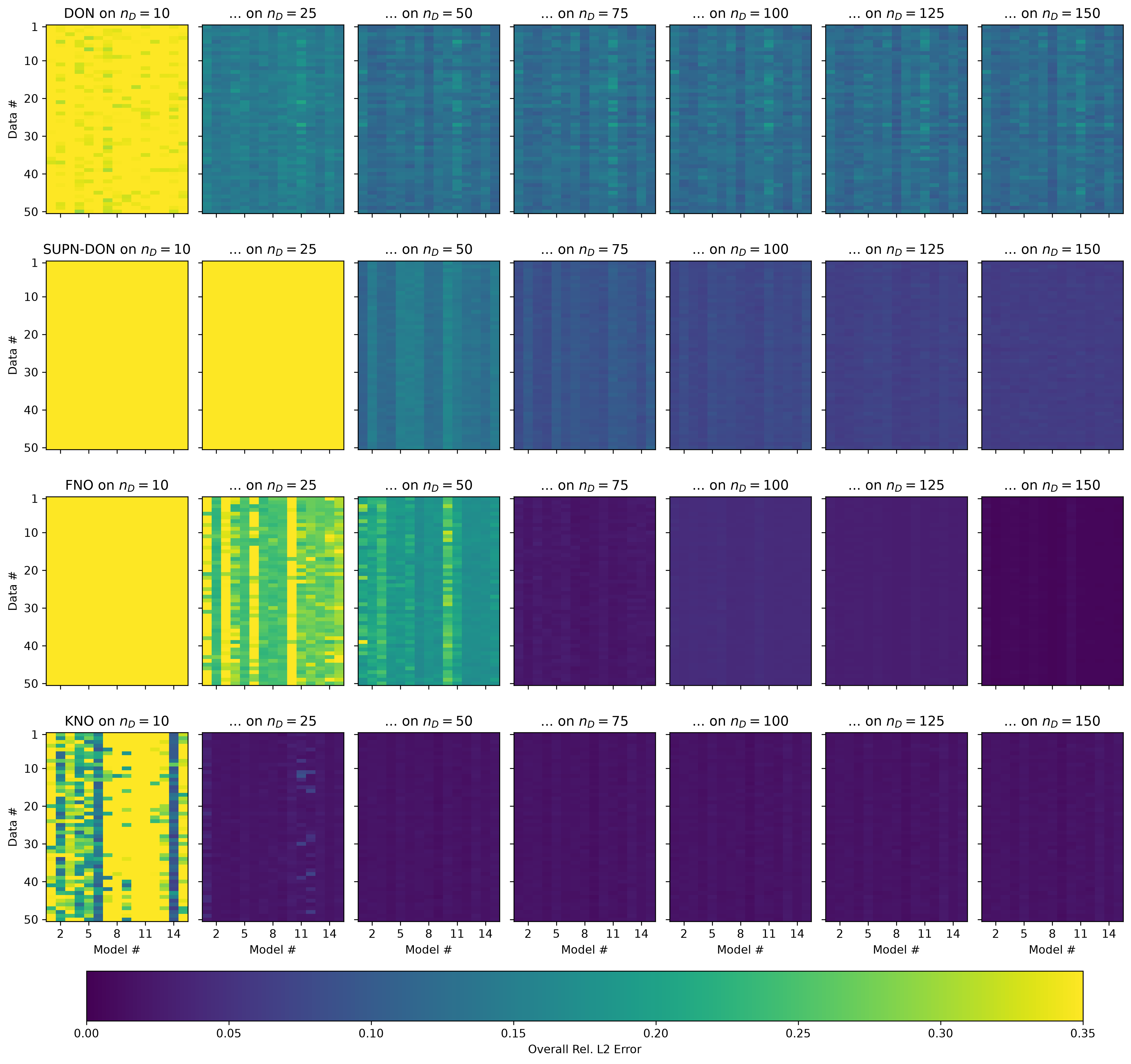}
    \caption{Heatmap of the \overLtwo for Models trained on Uniform Discretizations. Each pixel represents the \overLtwo of a specific model initialization trained on a specific training dataset. In general, increasing the size $n_D$ of the discretization improves \overLtwo for all models. For most architectures and grids, the heatmaps show vertical stripes: this suggests that some model initializations (columns) tend to have similar error across all datasets (rows) and one model initializations can differ significantly from another.}
    \label{fig:poly1d_unif}
\end{figure}

\begin{figure}[H]
    \centering
    \includegraphics[width=1\linewidth,keepaspectratio]{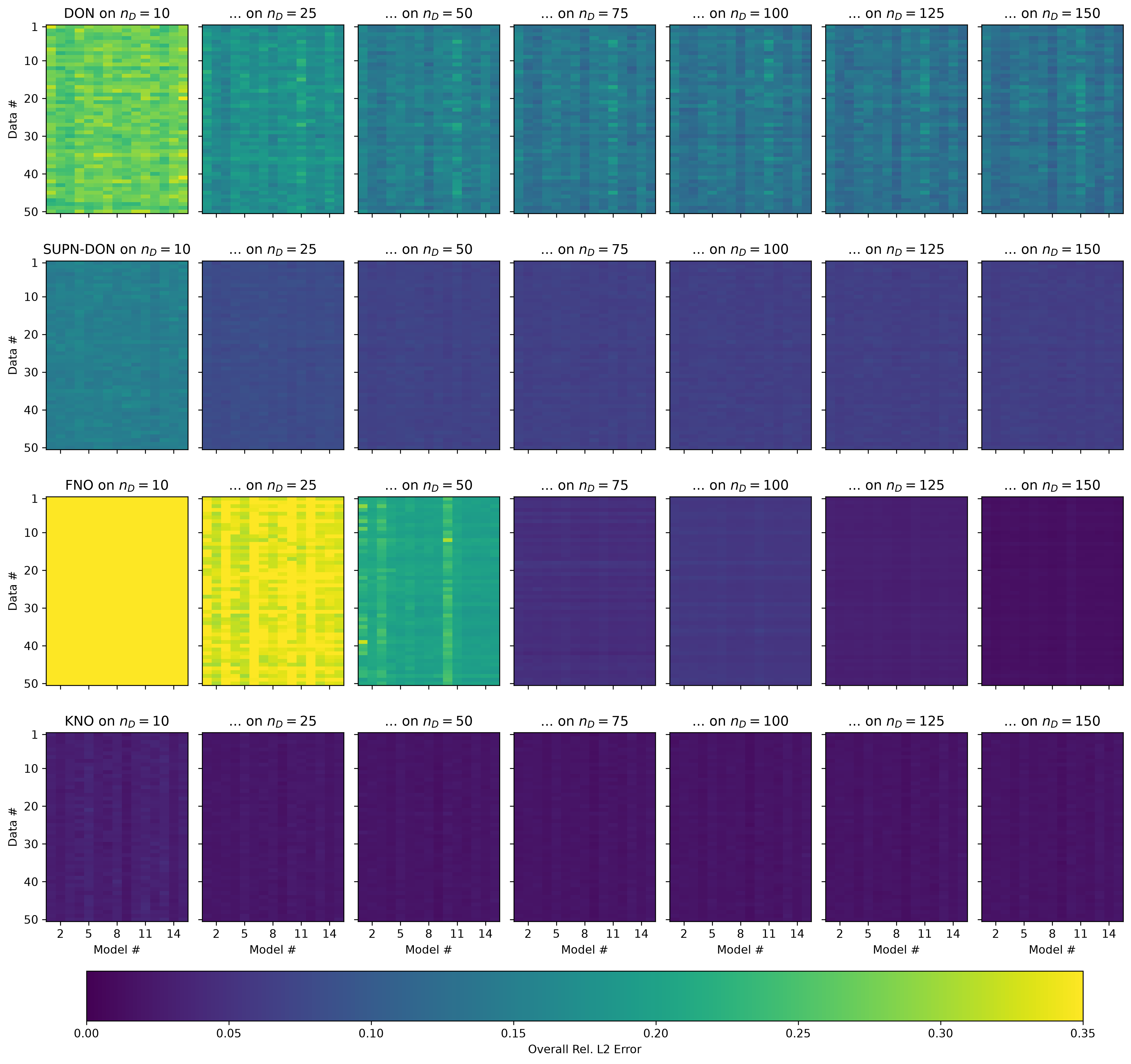}
    \caption{Heatmap of the \overLtwo for Models trained on Random Discretizations. Each pixel represents the \overLtwo of a specific model initialization trained on a specific training dataset. In general, increasing the size $n_D$ of the discretization improves \overLtwo for all models. In the small $n_D$ regime, we see that SUPN DONs and KNOs have significantly better performance than other models (and significantly better performance than on uniform discretizations). Model initialization still has a visible effect (i.e., pixels in a given column tend to be correlated), however this is much less pronounced than on uniform discretizations.}
    \label{fig:poly1d_rand}
\end{figure}

\begin{figure}[H]
\centering
\includegraphics[width=1\linewidth,keepaspectratio]{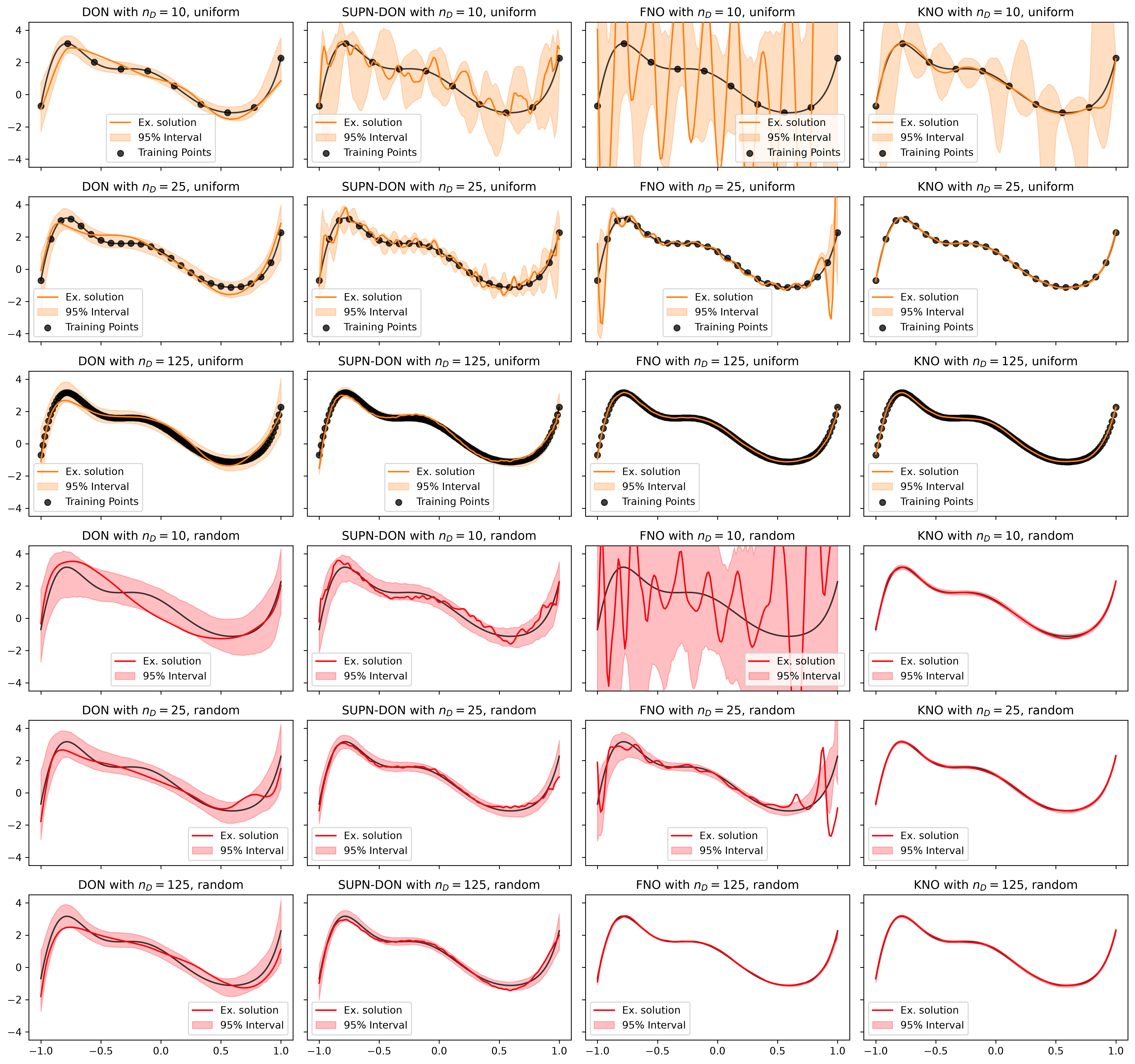}
\caption{Shows a comparison of different operator types on a test example function after training. First three rows show the model predictions for models trained on a uniform discretization (fixed training points are shown) where the bottom three rows show the model predictions for models trained on a random discretization (note: training points not shown since they differ for each training sample). Interval represents a 95\% quantiles across model initializations and training sets. } \label{fig:poly1d-ex-combo}
\end{figure}

Figures~\ref{fig:poly1d_unif} and \ref{fig:poly1d_rand} show the \overLtwo performance of each model type on a test set of functions. Each pixel represents one model initialization trained on one training set. In Figure~\ref{fig:poly1d_unif}, we see that on uniform discretizations, FNOs and KNOs tend to have the smallest error at high discretizations, with FNOs achieving the lowest mean overall relative $L_2$ error when the training and test discretizations coincide ($n_D=150$; see Table~\ref{tab:poly1d-times} in ~\ref{app:poly_training}). At extremely low discretizations, DONs and KNOs generally perform better on this example than SUPN DONs and FNOs. Looking at Figure~\ref{fig:poly1d-ex-combo}, we see FNOs approximately interpolate the discretization points but there are not enough training discretization points to accurately resolve the frequency behavior of the exact solution between training points. KNOs interpolate the training points on the uniform grid, but because of the initial assumptions on the learned kernel, the solution between grid points is much more regular than FNOs. Similarly, SUPN DONs have initialized basis functions that interpolate the values at the training grid points well, but are insufficiently constrained between grid points, whereas regular DONs have much more regularized basis functions that trade accuracy at the training points with a more smooth approximation everywhere.

\begin{figure}[H]
    \centering
    \includegraphics[width=1\linewidth,keepaspectratio]{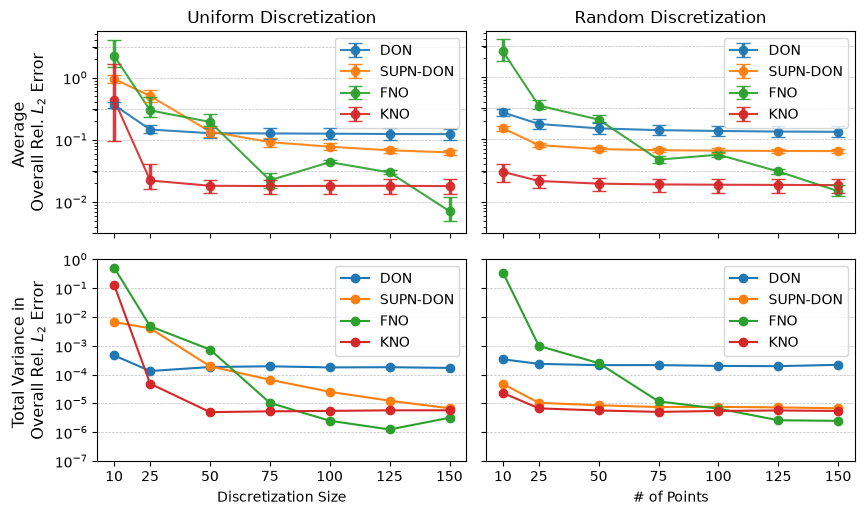}
    \caption{The top row shows the average of \overLtwo errors across model initializations and datasets with respect to discretization and discretization type; the error bars represent 95\% quantiles of the \overLtwo. \overLtwo converges as $n_D$ increases, with FNOs and KNOs performing the best. the bottom row shows the total variation of the \overLtwo (across model initializations and training datasets) for each model type is shown at each discretization level. SUPN DONs and KNOs show a significant reduction in the total variance with random discretizations compared to uniform discretizations.}
    \label{fig:pol1d-conv}
\end{figure}

Figure~\ref{fig:pol1d-conv} summarizes the convergence behavior with respect to the different discretizations $n_D$.
Model performance tends to not improve substantially after $n_D=75$ on uniform grids, with the notable exception of FNOs. We note that FNOs have a strong bias with respect to the training grid, i.e., a model with the same training and test grid performs much better than when the training and test grids are different. These results are consistent with the results shown in \cite{gao2025discretization}, where FNOs were found to have strong discretization mismatch errors when training and test grids were significantly different.

On random discretizations, we see that KNOs, DONs, and SUPN DONs show marked improvements in performance with low number of sample points per function (i.e., $n_D$ is small). SUPN DONs and KNOs in particular showed an improvement of almost an order of magnitude with $n_D=10$ (see Figure~\ref{fig:pol1d-conv}).
This makes sense: because every randomized training function has a different set of training points, KNOs and DONs (regular and SUPN) must perform better throughout the domain and cannot merely improve predictions at a fixed set of the domain during training.
FNOs, however, did not benefit from randomized sampling of the domain because the random samples had to be interpolated onto a regular grid for training--the accuracy of this interpolation determining the accuracy of the FNO. Figure~\ref{fig:poly1d-ex-combo} illustrates how the random discretization during training constrains the SUPN DONs and KNOs throughout the domain on a specific test example. Note that as $n_D$ increases, the performance of all models trained on random discretizations converges to the performance of models trained on uniform discretizations (see Figure~\ref{fig:pol1d-conv}); the error due to discretization size becomes negligible compared to other factors.

In summary, FNOs and KNOs achieved better results on this example for an increased computational cost--with the caveat that the discretization of the training grid is chosen appropriately.
When the number of discretization points was significantly limited ($n_D = 10$), DONs tended to perform the best on uniform grids, likely due to their implicit assumption of function smoothness.

\subsubsection{Model Variability and Sources of Uncertainty}
Figure~\ref{fig:pol1d-conv} shows the total variance in \overLtwo across training datasets and model initializations.
In general, the total variance in the \overLtwo measure of model performance tends to decrease as the discretization size increases, suggesting that models tend to converge to solutions with similar performance at higher discretizations.
The exception is regular DONs, which have the similar variability in their performance regardless of discretization size.
Figure~\ref{fig:poly1d-ex-combo} and \ref{fig:pol1d-conv} suggest possible explanations: DONs quickly converge to an over-smoothed solution whose errors are small everywhere.
Since the basis has an implicitly strong assumption about the regularity of solutions, this leads to more consistent solutions across discretizations and thus an equivalently consistent variability in the \overLtwo error.

\begin{figure}[H]
    \centering
    \includegraphics[width=1\linewidth,keepaspectratio]{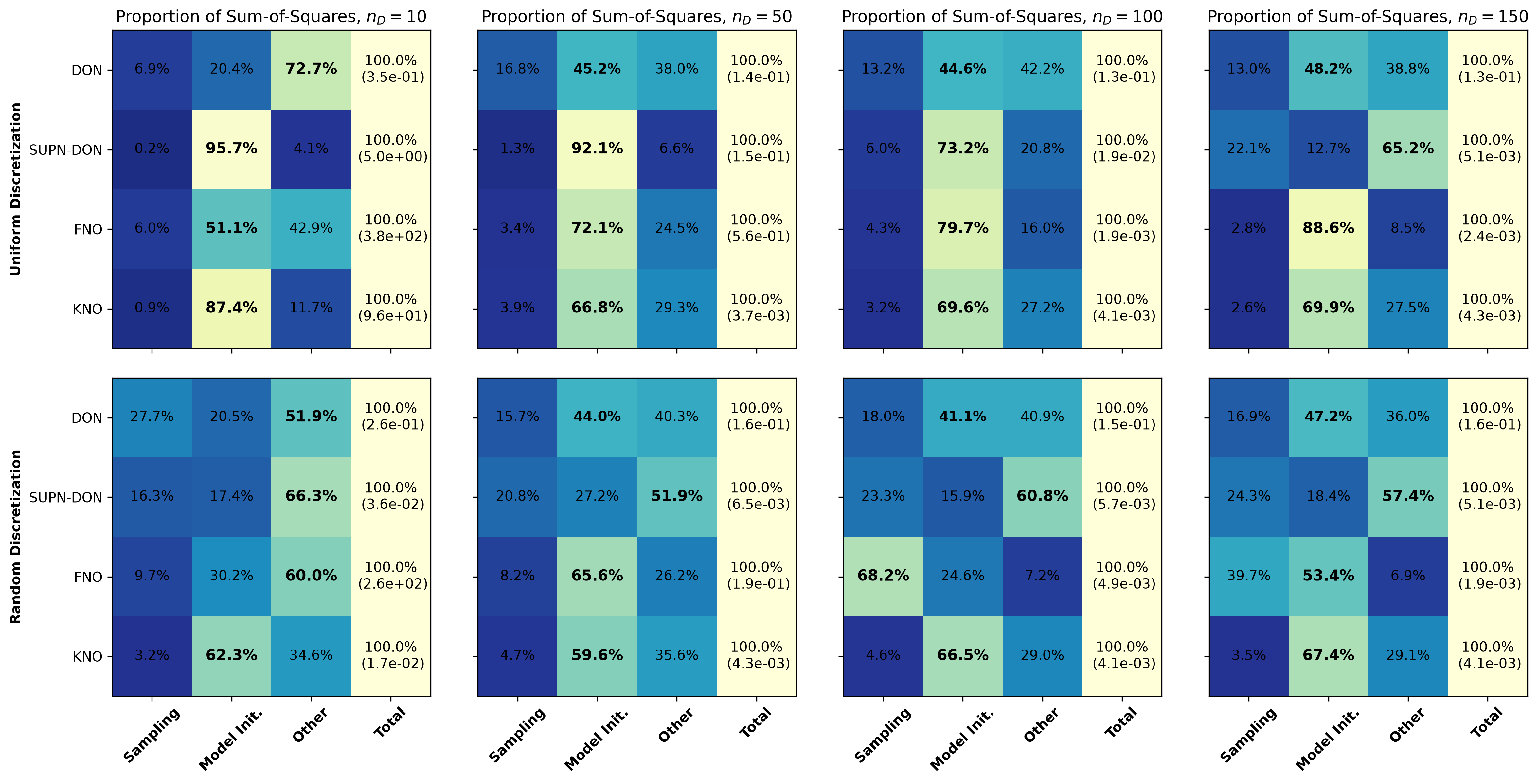}
    \caption{Proportion of sum of squares variance in the \overLtwo at a select set of discretizations for the Nonlinear Polynomial example, broken down by model type, sampling set, and model initializations. On uniform discretizations, model initialization tends to dominate the proportion of variance except in the case of DONs at $n_D=10$ and SUPN DONs at $n_D=150$. On random discretizations, sampling tends to have a larger contribution to the SS at lower discretizations compared to uniform discretizations, which is likely due to the effect of different randomized discretizations per sample set. Note that the ``Other'' effects is the remaining residual sum-of-squares, which includes interaction effects between sampling and model initialization as well as stochastic optimization.}
    \label{fig:poly1d_propvar}
\end{figure}

Figure~\ref{fig:poly1d_propvar} shows a numeric breakdown of the proportion of variance in the \overLtwo using the ratios of the sums of squares as discussed in Section~\ref{sec:variance_decomp}.
Model initialization tends to dominate the variability in \overLtwo, especially on uniform discretizations.
This is also represented in Figure~\ref{fig:poly1d_unif} by the long vertical bands, where a single model initialization tends to have consistently higher or lower error profile across all datasets.
For this example, finite sampling (of functions) had only a small effect on the variability of the model, suggesting that a training set of $n_s=250$ functions lead to similar overall performance. 

Note that for random discretizations, the variability due to sampling includes the effect of the randomly sampled discretization as well, thus the contribution to variability in \overLtwo increases compared to uniform discretizations.
We also see that for SUPN DONs and KNOs, using a randomized discretization lowers the total variability (recall FNOs do not benefit because of the linear random sample-to-grid interpolation) and reduces the proportion of variability due to model initialization, though model initialization remains a significant factor.
We caveat this observation by noting that the sum-of-squares proportion of variance only captures the independent main effects of the sampling and model initialization--especially for random discretizations, it is possible that there are interaction effects between sampling and model initialization which are included in the ``other'' or residual explained variance.

Finally, we highlight a main difference between the behaviors of FNOs and SUPN DONs on a uniform discretization: as $n_D$ increases, the contribution of model initialization to the total of variance increases for FNOs and decreases for SUPN DONs.
For SUPN DONs, Figure~\ref{fig:poly1d-ex-combo} provides a simple explanation: as mentioned earlier, solutions from SUPN DONs tend to interpolate the solution at training grid points with strong precision, leaving the solution \emph{between} grid points to be dominated by the effects of model initializations.
When the training grid $n_D$ is the same as the test grid ($n_D=150$), model initialization does not dominate the variance because the SUPN DONs have interpolated the function solutions on the output test grid very well.
However, FNOs at low discretization suffer from aliasing errors and Gibbs-like phenomena, which contribute to the variation in the \overLtwo error regardless of model initialization.
As $n_D$ increases, the overall variability of FNOs significantly decreases and we observe only small differences between model predictions, with model initialization growing to be the dominant contributor.

\subsubsection{Consequences for Ensemble UQ}\label{sec:poly1dUQ}
Initialization ensembles are commonly used to estimate
uncertainty in neural operator predictions. However,
model initialization represents only one source of
epistemic uncertainty, and ensembles trained on a single
dataset may not capture the effects of finite training
function sampling. Given that initialization accounts
for a substantial proportion of the variance in overall
relative $L_2$ error in our experiments, we investigate
whether initialization ensembles can reliably estimate
quantiles of the overall relative $L_2$ error distribution.

\begin{figure}[H]
    \centering
    \includegraphics[width=1.0\linewidth,keepaspectratio]{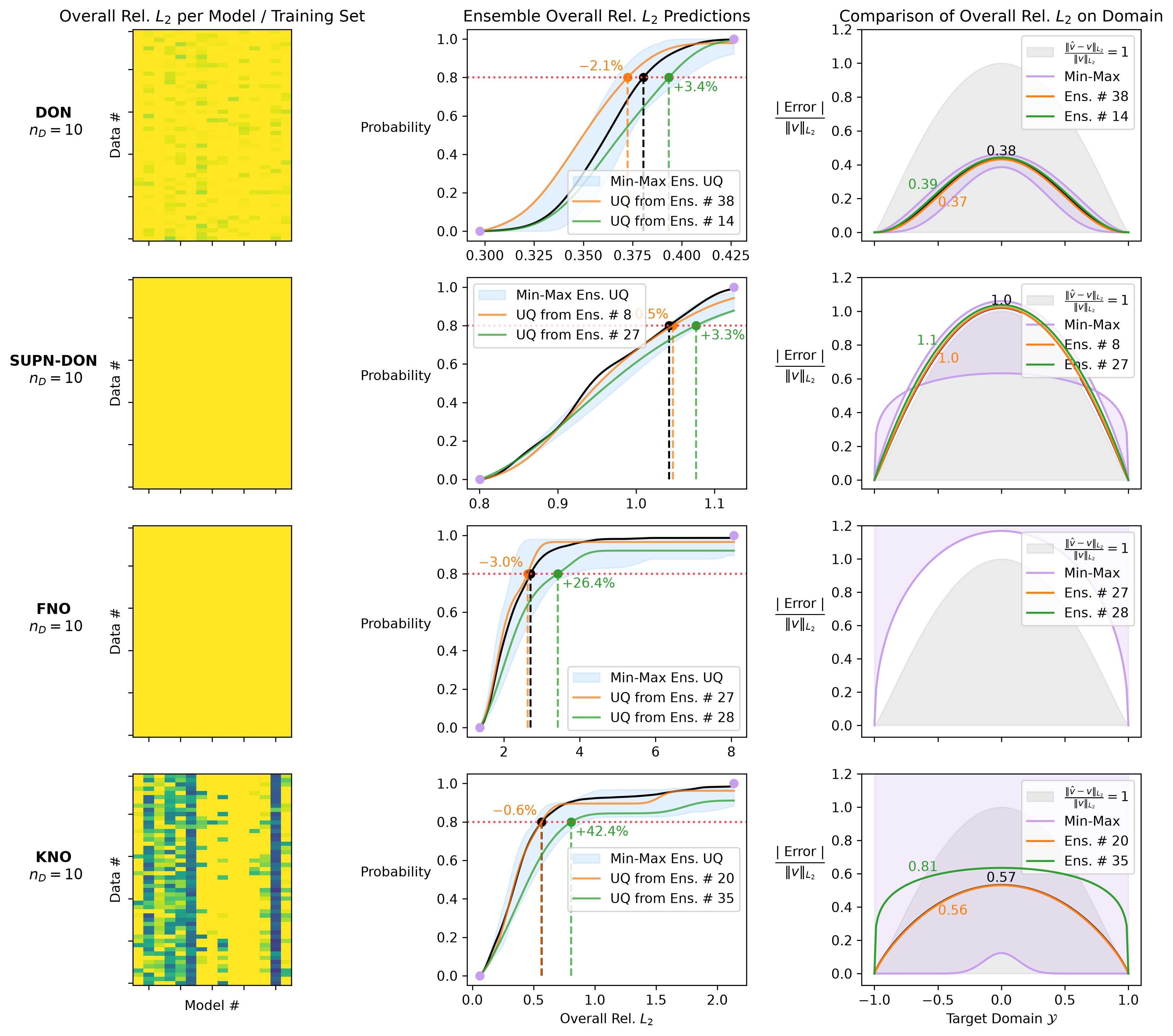}
    \caption{The \textbf{left column} shows the heatmap of the \overLtwo on the test set for each model given initialization and training dataset (as in Figure~\ref{fig:poly1d_unif}) on a uniform discretization, $n_D=10$. The \textbf{middle column} shows CDFs of the \overLtwo error: the true observed CDF computed over all models trained on all datasets (black), CDFs computed from two separate ensembles trained on single datasets (orange and green), and the min-max CDFs estimated from an ensemble (blue region). The plotted ensemble members were selected as the lower 10\% (orange) and upper 90\% quantile estimators of an 80\%-quantile prediction of the \overLtwo.
    The \textbf{right column} interprets the differences in 80\%-quantile predictions of the \overLtwo by comparing it to \overLtwo$=1$ and the min-max \overLtwo observed across all models trained on all datasets (purple region). The \overLtwo is visualized as the integral (i.e., area) under a beta distribution defined over the domain $\mathcal{Y}$. The two predictions of the 80\% quantile of the \overLtwo from the previously plotted ensemble CDFs (orange and green) are shown and compared to the true 80\% quantile from the full empircal CDF (black).}
    \label{fig:poly1d_uqOverL2-ld}
\end{figure}

\begin{figure}[H]
    \centering
    \includegraphics[width=1.\linewidth,keepaspectratio]{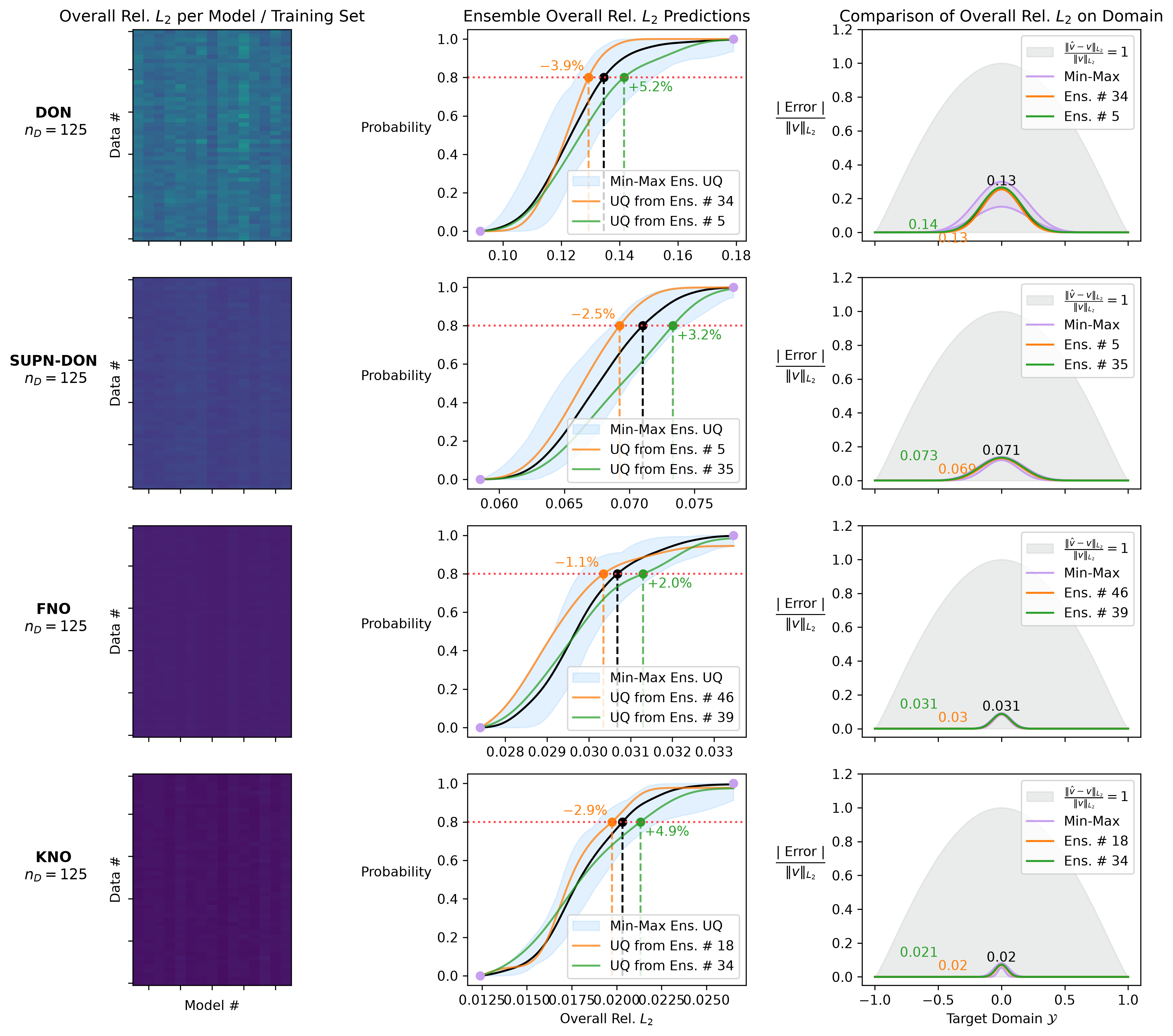}
    \caption{Ensemble estimates of the overall relative $L_2$
error distribution, as in Figure~\ref{fig:poly1d_uqOverL2-ld}, but with a uniform
output discretization of $n_D=125$ instead of $n_D=10$.}
    \label{fig:poly1d_uqOverL2-hd}
\end{figure}

Figure~\ref{fig:poly1d_uqOverL2-ld} and \ref{fig:poly1d_uqOverL2-hd} illustrate and interpret the quality of UQ estimates of the \overLtwo \\ provided by using an ensemble trained on a given dataset with a uniform discretization, $n_D=10$ and $n_D=125$, respectively.
We provide similar plots for all discretizations in~\ref{app:A}.
For each dataset $k=1,\ldots,50$, we utilize the same ensemble of $n_m=15$ models to estimate a CDF of the \overLtwo.
We compare this to the true observed CDF estimated using all models trained on all datasets.

At higher discretization levels (e.g., $n_D \geq 75$),
most ensemble estimates did not deviate from the
empirical CDF by more than 10\% for most confidence
levels. Figure~\ref{fig:poly1d_uqOverL2-hd}
illustrates this behavior at $n_D=125$, where
initialization ensembles trained on individual datasets
provide estimates of the 80\% \overLtwo quantile
that agree well with the empirical quantile across
all training datasets. In other words, training an ensemble of models on a single dataset is sufficient to answer the question; ``What \overLtwo error do we feel confident the model will achieve 80\% of the time?''

However, for $n_D=10$, Figure~\ref{fig:poly1d_uqOverL2-ld} shows that ensembles of FNOs and KNOs tend to overestimate the appropriate \overLtwo quantiles, while ensembles of DONs and SUPN DONs still provide appropriate UQ estimates most of the time.
In other words, an ensemble of FNOs and KNOs trained on a single dataset is unlikely to provide robust quantile estimates for the \overLtwo, even though model initialization contributes significantly to the variability in the \overLtwo error.
One explanation is that some model initializations perform significantly worse on specific datasets, leading the ensemble to overestimate the quantile values of \overLtwo for FNOs and KNOs with $n_D=10$.

For the sake of brevity, we only illustrate the UQ on two uniform discretizations. Plots for all uniform discretizations are provided in~\ref{app:A}.
Generally, we observed that ensemble UQ estimates of \overLtwo error with random discretizations are equivalent or better than on uniform discretizations.

\begin{figure}[H]
    \centering
    \includegraphics[width=1.\linewidth,keepaspectratio]{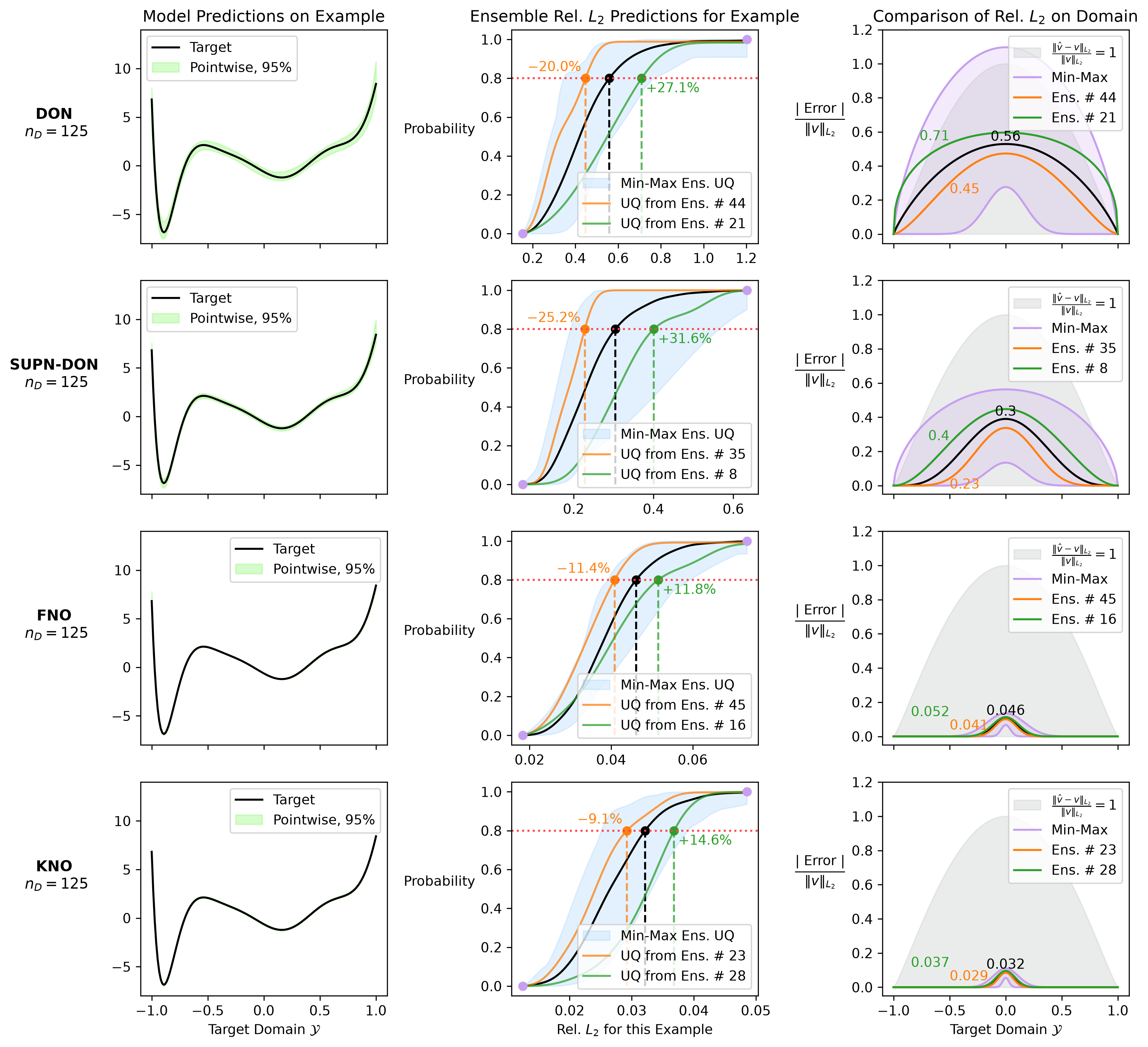}
    \caption{The \textbf{left column} shows an output target function for a \emph{specific} randomly selected example from the test set: the green shaded region represents a pointwise 95\%-quantile taken across all models trained on all datasets with a uniform discretization ($n_D=125$). The \textbf{middle column} shows CDFs of the \emph{specific} Rel. $L_2$ error (i.e., the Rel. $L_2$ for this specific example): the true observed CDF computed over all models trained on all datasets (black), CDFs computed from two separate ensembles trained on single datasets (orange and green), and the min-max CDFs estimated from an ensemble (blue region). The plotted ensemble members were selected as the lower 10\% (orange) and upper 90\% quantile estimators of an 80\%-quantile prediction of the \emph{specific} Rel. $L_2$ for this example. The \textbf{right column} interprets the differences in 80\%-quantile predictions of the \emph{specific} Rel. $L_2$ by comparing it to a Rel. $L_2=1$ and the min-max \emph{specific} Rel. $L_2$ (for this example) observed across all models trained on all datasets (purple region). The Rel. $L_2$ is visualized as the integral (i.e., area) under a beta distribution defined over the domain $\mathcal{Y}$. The two predictions of the 80\% quantile of the specifc Rel. $L_2$ from the previously plotted ensemble CDFs (orange and green) are shown and compared to the true 80\% quantile from the full empircal CDF (black).}
    \label{fig:poly1d_uqSPL2-hd}
\end{figure}

While ensemble UQ estimates are robust for the \overLtwo error, i.e., the model performance overall, ensemble UQ estimates for specific examples was much more lackluster.
Recall that the \overLtwo is the expectation of the \emph{specific} $L_2$ of each individual example with respect to the distribution of the examples $v\sim \mathcal{D}_{\mathcal{G}(U)}$ in the test set.
The uncertainty of model performance in this average sense is much easier to quantify than on a per example basis.
This is evident in Figure~\ref{fig:poly1d_uqSPL2-hd}.
In the first column, we see an example target function $v$ from the test set as well as the pointwise 95\%-quantiles computed across all models trained on all datasets with the given discretization ($n_D=125$).
In the second column, we see the observed CDF computed as before except for Rel. $L_2$ of the example target function shown rather than the \overLtwo.
For the example shown, the ensemble CDF estimates can vary significantly depending on the training dataset the ensemble was trained on.
Even for larger $n_D=125$ and higher accuracy models (i.e., FNOs and KNOs), a single ensemble UQ prediction could potentially vary between 10\% and 15\% of the true quantile $L_2$ value.

To assess whether this behavior extends beyond the
example in Figure~\ref{fig:poly1d_uqSPL2-hd},
Table~\ref{tab:poly1d_uq80ex} reports the average
proportion of test functions for which ensemble
estimates of the 80\% error quantile deviate from
the empirical quantile by more than 10\% or 30\%.
Substantial deviations occur for many test functions,
with results across discretizations provided in ~\ref{app:poly_relL24ex}.

\begin{table}[H]
\caption{Shows 80\%-quantile accuracies for ensemble model estimates on uniform discretizations ($n_D=125$), measured by the average (across datasets) of the proportion of examples where the Rel. $L_2$ quantile ensemble estimate differed from the observed value (across all datasets and models) by more than 10\% or 30\%.}
\label{tab:poly1d_uq80ex}
\begin{tabular}{lcccc}
\toprule
 & $n_D$ & \makecell{ Average Proportion of Examples \\ where 80\%-quantile \\Estimate differs by $>$ 10\%\ } & \makecell{ Average Proportion of Examples \\ where 80\%-quantile \\ Estimate differs by $>$ 30\%\ } \\
\midrule
DON & 125 & 0.647 & 0.156 \\
\cline{1-4}
SUPN-DON & 125 & 0.645 & 0.154 \\
\cline{1-4}
FNO & 125 & 0.111 & 0.003 \\
\cline{1-4}
KNO & 125 & 0.445 & 0.021 \\
\bottomrule
\end{tabular}
\end{table}

The fact that there is significant variation in the UQ estimates of the specific Rel. $L_2$ for ensembles trained on different datasets should come as no surprise.
Ensembles trained on different datasets may be biased to perform better on certain examples compared to others, even as the overall performance across the distribution of test examples remains relatively stable.
However, severity of such over / under estimates must necessarily be context dependent.
While under-estimating the Rel. $L_2$ error by more than 20\% for a specific example may be concerning statistic, if the overall model precision is well below the threshold of importance, such differences may not matter.
In the right column of Figure~\ref{fig:poly1d_uqSPL2-hd}, we see that even though the 80\%-quantile estimates for FNOs and KNOs deviate by around 10\%, a comparison relative to the Rel. $L_2=1$ makes such differences in UQ appear trivial.
however, for DONs and SUPN DONs, where the magnitude of the Rel. $L_2$ is much higher, differences of 20 or 30\% have a much more visually apparent effect.

Our key observation is that using initialization
ensembles to make strong statistical claims about
the relative $L_2$ error of specific input functions
can be unreliable at best and misleading at worst.
While ensemble UQ generally provides consistent
estimates of \overLtwo, its error quantile estimates
for individual input functions can vary substantially
across training datasets, even at higher discretization
levels. For high-accuracy models, it may be more
appropriate to make general statements about
\overLtwo relative to application-specific accuracy
requirements than to infer confidence in the error
of individual predictions.

\subsection{2D nonlinear Allen-Cahn equation} \label{sec:ac_results}
Consider the following stationary PDE problem
\begin{align}\label{eq:acpde}
\lambda \left(\frac{\partial^2v}{\partial x^2} + \frac{\partial^2 v}{\partial y^2} \right) - v(v^2-1) &= u(x,y), \; (x,y) \in [-1,1] \times [-1,1] \\
v|_{\partial\Omega} &= 0
\end{align}
where $\lambda = 0.1$ and we assume Dirichlet boundary conditions. 

For this example, we assume that the solutions $v(x,y)$ of interest take the form 
\begin{align}\label{eq:acsol}
    v(x,y) = B(x,y)\cdot P_\omega(x,y)
\end{align}
where $\omega\sim U[0,\pi]^4$ is a random variable and $B(x,y)$ and $P_\omega(x,y)$ are defined by:
\begin{align}
    B(x,y)&=(x-1)(x+1)(y-1)(y+1)\\
    \begin{split}
        P_\omega(x,y)&=\sin(\omega_1(\pi x))\sin(\omega_2 (\pi y)) + \cos\left(\omega_2 (\pi x)^2\right)\sin\left(\omega_3 (\pi y)^2\right)\\
        &\quad\quad\quad +\sin\left(\frac{\omega_3}{\pi} (\pi x)^3\right)\cos\left(\frac{\omega_4}{\pi} (\pi y)^3\right) .
    \end{split}
\end{align}

Our goal is to obtain an operator which approximates the solution $v$ given the input forcing function $u$, i.e., to learn a nonlinear operator $\mathcal{G}$ such that
\begin{equation}
    \mathcal{G}(u)=v
\end{equation}
for all $u\in\mathcal{F}$. For the purposes of this synthetic example, the class of forcing functions $\mathcal{F}$ was formed by sampling $\omega\sim U[0,\pi]^4$ and taking the appropriate derivatives. Representative forcing functions and their corresponding
solutions are provided in~\ref{app:ac_examples}
(Figure~\ref{fig:ac_examples}).

\textbf{Discretization:} All input functions $u$ were discretized using a uniform grid in  $[-1,1]\times[-1,1]$ with 128 grid points per dimension. The output functions $v$ were discretized with uniform grids at $n_D=\{8,16,32,64,128\}$ in each dimension. For random discretizations, as in the previous example, points were sampled uniformly (with $n_D\times n_D$ samples) in the domain for DeepONet-style networks. 
For KNOs and FNOs, the randomly sampled points were interpolated onto a uniform style grid utilizing linear interpolation.

\textbf{Dataset Generation:} we sample $n_s=500$ training pairs $(u,v)$ by first sampling the $\omega$, generating $v$ from \eqref{eq:acsol}, and then evaluating the differential equation in \eqref{eq:acpde} to find $u$ at the discretization points. We generate $k=1,\ldots,10$ such datasets. As in our previous example, each training dataset can be described as $\mathcal{D}_{k,n_D}=\{(u^i_{128\times128},v^i_{{n_D}\times {n_D}})\}$, where $k$ is the dataset number and $n_D$ is the discretization condition described above.

\textbf{Model Initialization:} for each model type we initialize the parameters of $n_m=10$ models using standard randomization procedures for each operator type. Initializations were saved so that each initialization could be trained independently on each dataset $\mathcal{D}_{k,n_D}$.

\textbf{Training Procedures:} for this example, DON, SUPN DON and KNOs were all trained with the SOAP optimizer~\cite{vyas2025soap}.
SOAP was not compatible with FNOs because FNOs require the implementation of backpropagation with complex numbers (used in the Fourier transform).
Instead, FNOs were optimized with regular Adam and a cosine-annealing learning rate scheduler.
All models were trained with a maximum of 5000 epochs and an initial learning rate of $\text{lr}=2e{-3}$.
The average timing of one training epoch using a standard Adam optimizer at the highest discretization is shown in Table~\ref{tab:acmod-times}.

\subsubsection{Overall Performance:}
\begin{figure}[H]
    \centering
    \includegraphics[width=1\textwidth,keepaspectratio]{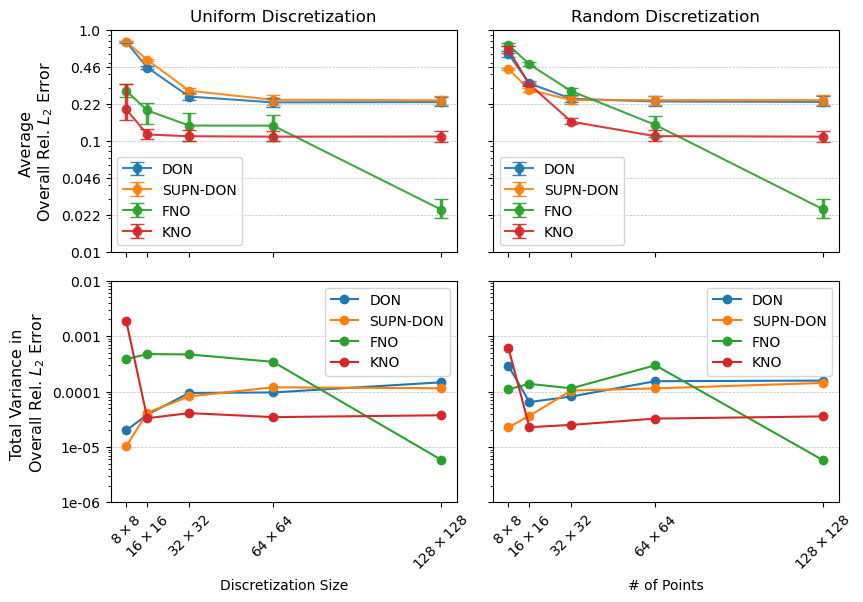}
    \caption{The top row shows the average of \overLtwo errors across model initializations and datasets with respect to discretization and discretization type; the error bars represent 95\% quantiles of the \overLtwo. \overLtwo converges as $n_D$ increases, with FNOs and KNOs performing the best. On random discretizatins $n_D=8$, SUPN DONs perform better than FNOs and KNOs, but not as well as FNOs and KNOs trained on uniform discretizations with the same number of points. The bottom row shows the total variation of the \overLtwo (across model initializations and training datasets) for each model type is shown at each discretization level. The total variation in the \overLtwo for KNOs significantly drops when $n_D$ changes from $8$ to $16$, whereas it curiously increases for DON and SUPN DONs as $n_D$ increases.}
    \label{fig:ac_conv}
\end{figure}

Figure~\ref{fig:ac_conv} shows the convergence of models in the \overLtwo with respect to discretization sizes across datasets and model initializations.
Note that one of the FNO models trained on a uniform discretization, $n_D=32$, did not converge and is excluded from the results shown in Figure~\ref{fig:ac_conv}.
On uniform discretizations, FNOs far outperformed DeepONet style networks on all discretization levels, even on small discretization grids like $n_D=8$.
This is likely due to the periodic behavior of the input forcing functions and the corresponding outputs $(u,v)$, for which the Fourier transformation / basis of FNOs may be especially optimal.
Representative predictions under uniform and random
discretizations are provided in~\ref{app:ac_examples}
(Figures~\ref{fig:ac_unifEX} and~\ref{fig:ac_randEX}).
These examples illustrate spatial differences in the
predicted solution fields that are not captured by
aggregate relative $L_2$ errors.

On random discretizations, DeepONets perform better on smaller sample sizes, with FNOs becoming comparable on sample sizes of $32\times32$.
Because FNOs do not have an algorithmic approach for random discretizations, the error in their performance is bounded by the accuracy of the interpolation of the randomly sampled points onto a uniform grid, which can be especially poor with small samples.
In general, we note that DeepONets with SUPN trunks performed slightly better compared to DeepONets with a standard MLP trunk on smaller samples.
In addition, DeepONets trained on randomized discretizations performed much better than when trained on uniform discretizations in the low-sample size regime.
We hypothesize that random sampling of the domain prevents DeepONets from overfitting grid points on coarse grids, allowing for better generalization to the test set, similar to the polynomial example.

\subsubsection{Proportion of Variance:}
\begin{figure}[H]
    \centering
    \includegraphics[width=1.\textwidth,keepaspectratio]{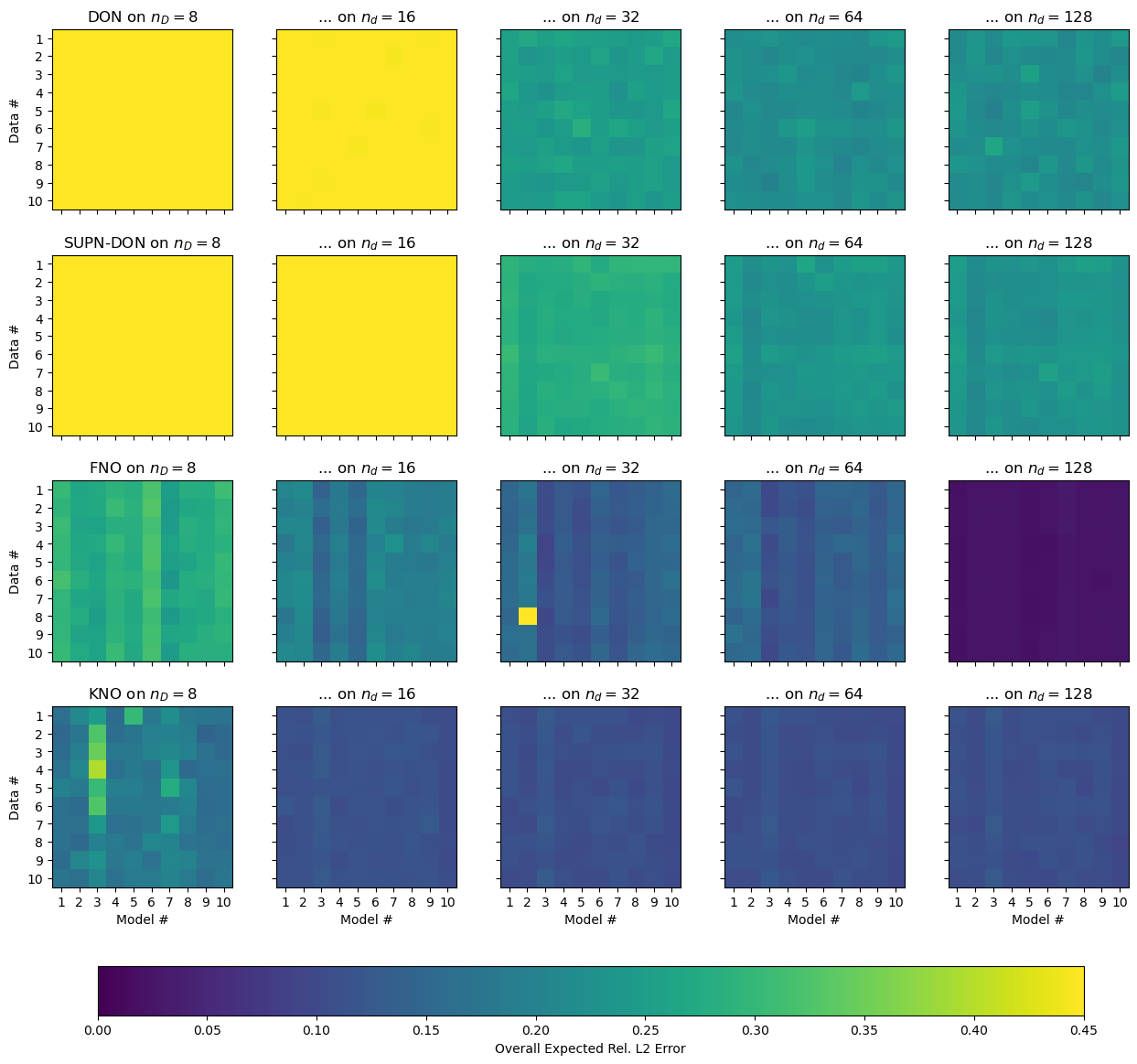}
    \caption{Heatmap of the \overLtwo for Models trained on Uniform Discretizations for Allen Cahn example. Each pixel represents the \overLtwo of a specific model initialization trained on a specific training dataset. In general, increasing the size $n_D$ of the discretization improves \overLtwo for all models. With respect to model initializations, FNOs have similar vertical striping behavior observed on the polynomial example: model initializations (columns) tend to have more similar error across all datasets (rows). However, this behavior is not as apparent for DONs and SUPN DONs.}
    \label{fig:ac_unifL2}
\end{figure}

\begin{figure}[H]
    \centering
    \includegraphics[width=1.\textwidth,keepaspectratio]{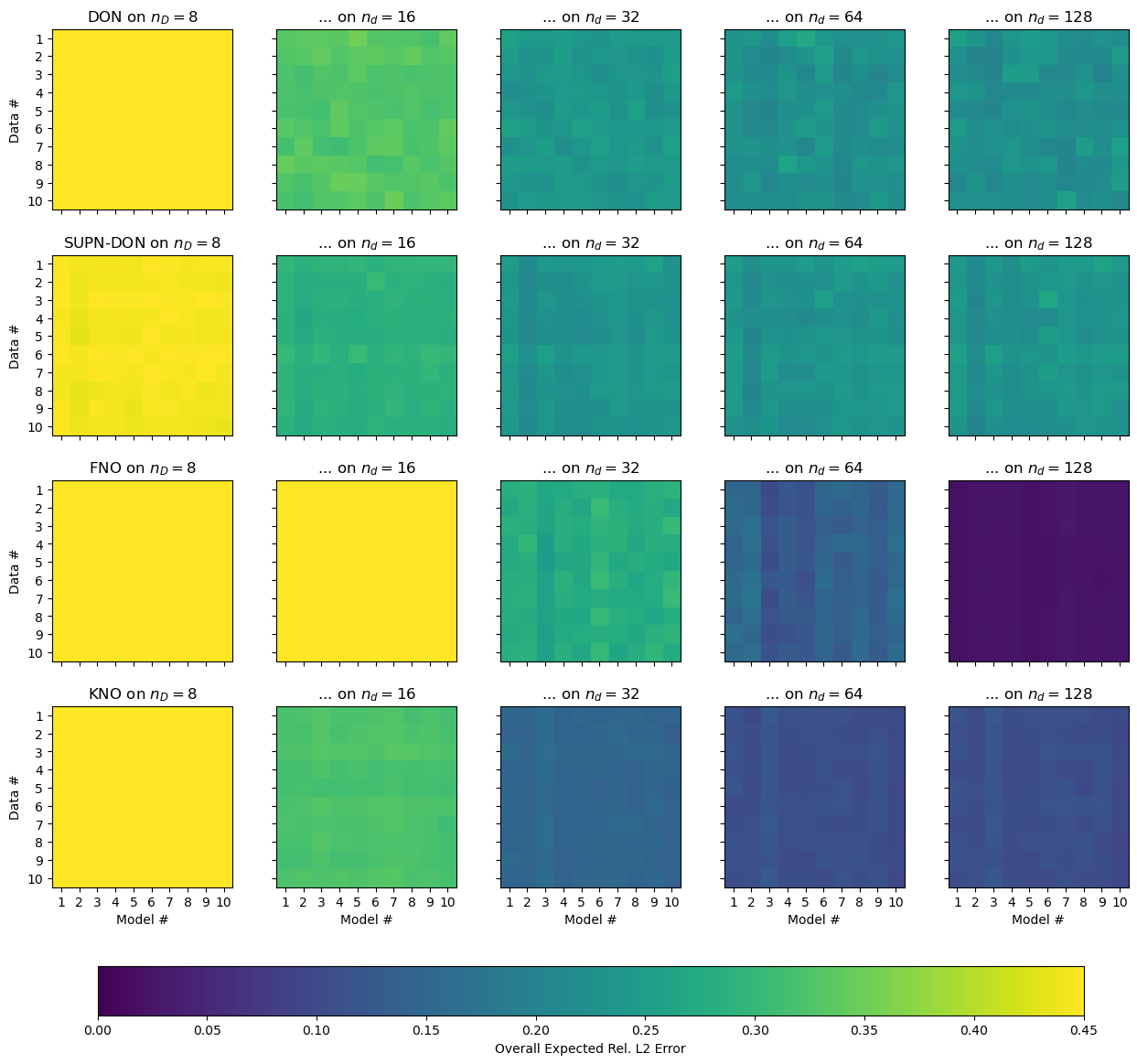}
    \caption{Heatmap of the \overLtwo for Models trained on Random Discretizations for Allen Cahn example. Each pixel represents the \overLtwo of a specific model initialization trained on a specific training dataset. In general, increasing the size $n_D$ of the discretization improves \overLtwo for all models.}\label{fig:ac_randL2}
\end{figure}

Figure~\ref{fig:ac_unifL2} and \ref{fig:ac_randL2} shows the \overLtwo error for each type of operator model with different initializations trained on different datasets.
We notice, as in the previous example, that FNOs tend to have much more variation across model initializations versus across training datasets (columns differ from each other more than rows).
As before, we compute the sum-of-squares proportion of variance to quantitatively confirm this observation.
Note that the FNO model initialization \# 2 trained on dataset \# 8 ($n_D=32$) did not converge; we excluded the corresponding model initialization (i.e., column \# 2 in Figure~\ref{fig:ac_unifL2}) from the proportion of variance analysis to maintain the accuracy of the sum-of-squares variance decomposition.

\begin{figure}[H]
    \centering
    \includegraphics[width=1.\linewidth,keepaspectratio]{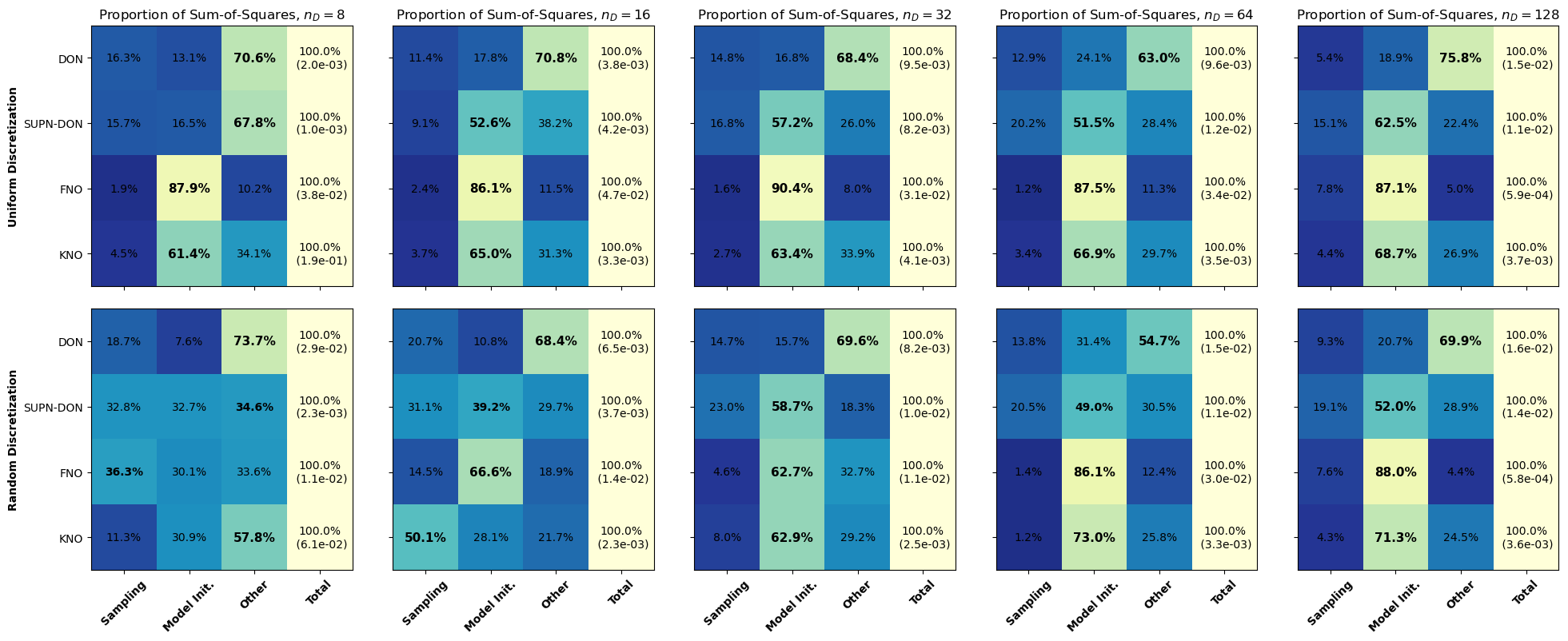}
    \caption{Sum-of-squares contributions to variance in \overLtwo
for the Allen--Cahn problem. Model initialization
dominates on uniform grids, except for DONs and
SUPN DONs at $n_D=8$. On random discretizations,
sampling contributes more at low resolutions, likely
due to variation in output locations. The ``Other''
category includes interactions between sampling and
initialization, as well as residual variation from
stochastic optimization.}
    \label{fig:acmod_propvar}
\end{figure}

The results for the proportion of sum-of-squares variance decomposition are shown in Figure~\ref{fig:acmod_propvar}.
As suspected, the variance in the overall performance of FNOs on uniform discretizations is dominated by the model initialization ($>80$\% on all uniform discretizations). 
The variance in the overall performance of DeepONets without a SUPN trunk tends to be dominated by factors other than the model initialization or the sampling dataset ($>$60\% on all uniform discretizations), while the variance for DeepONets with a SUPN trunk tends to be heavily influenced by the model initialization ($>$50\% except on $n_D=8$) but is also influenced by the data training sample as well as other factors.
On random discretizations, it is clear that the randomization of the domain plays a role in the overall model performance (especially with a smaller number of points), which increases the proportion of variance finite sampling plays for all models.
Nonetheless, model initialization still maintains a large contribution in the proportion of variance.

\begin{figure}[H]
    \centering
    \includegraphics[width=1.\linewidth,keepaspectratio]{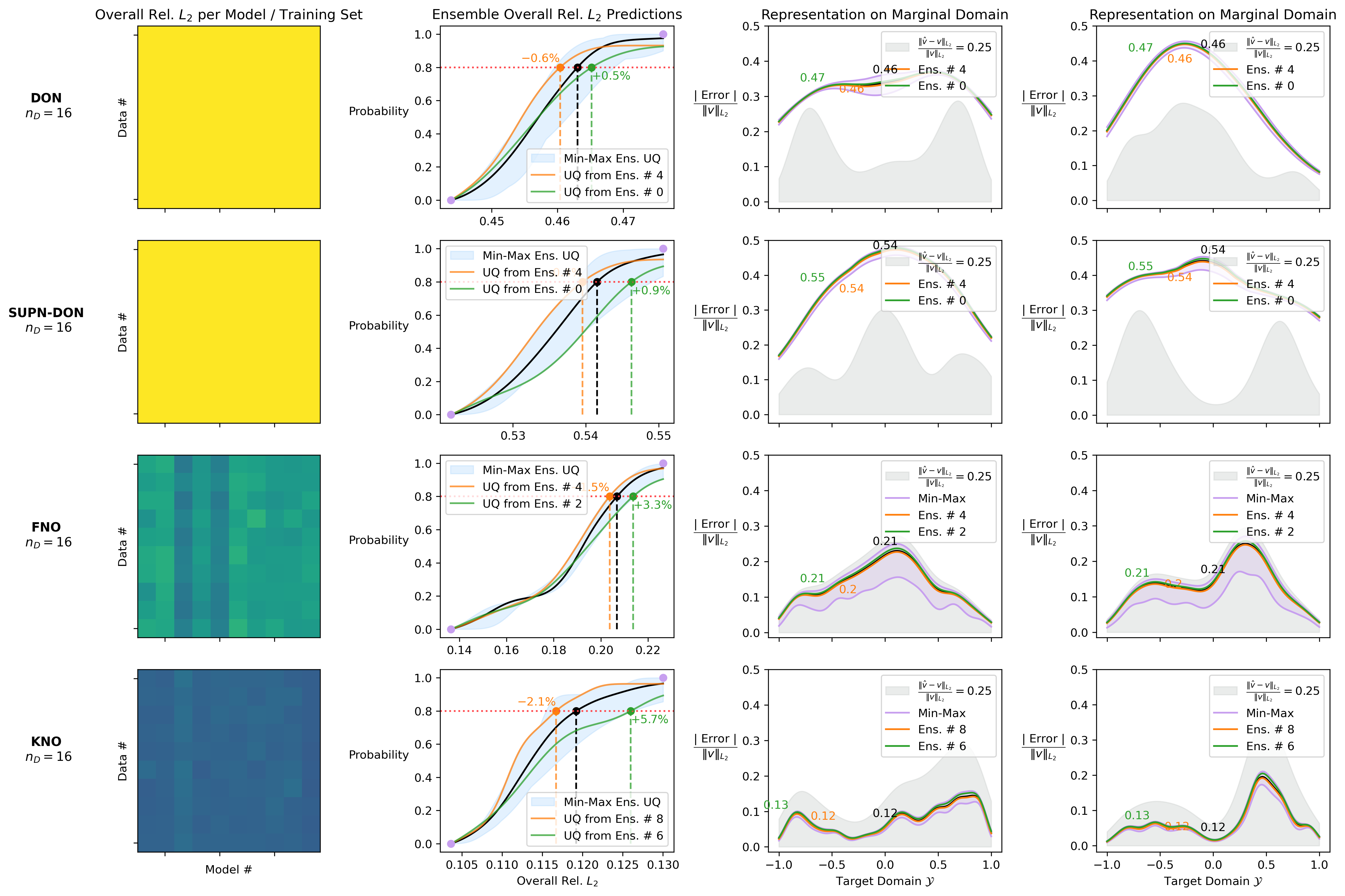}
    \caption{Ensemble estimates of the overall relative $L_2$
error distribution for the Allen--Cahn problem with
uniform discretization ($n_D=16$). The visualization
follows Figure~\ref{fig:poly1d_uqOverL2-ld}, except that the rightmost columns show
spatially marginalized error distributions along both
coordinate directions, compared against a benchmark
relative $L_2$ error of 0.25.}
    \label{fig:acmod_uqover}
\end{figure}

\subsubsection{Consequences for Ensemble UQ}\label{sec:acuq}

Figure~\ref{fig:acmod_uqover} compares ensemble estimates of the overall
relative $L_2$ error distribution for the Allen--Cahn
problem at $n_D=16$. As in the polynomial experiment,
initialization ensembles generally provide reliable
estimates of aggregate error quantiles. The main
exception occurs for KNOs at $n_D=8$, where estimates
vary substantially across training datasets (see~\ref{app:acuq}).

\begin{figure}[H]
    \centering
    \includegraphics[width=1\linewidth,keepaspectratio]{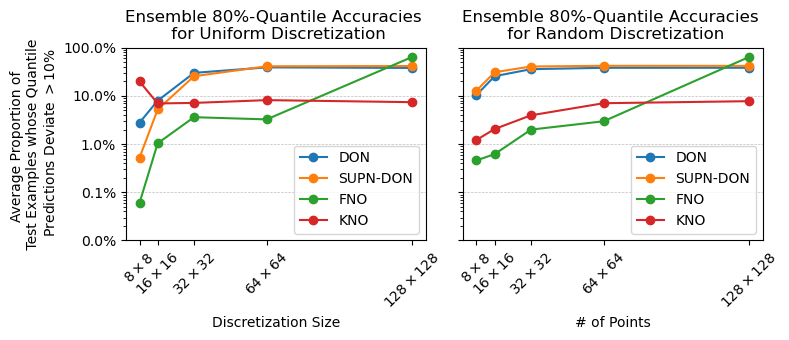}
    \caption{Shows estimates of the 80\%-quantile ensemble accuracies for the Allen Cahn example. For each ensemble, the average proportion of examples in the test set where the 80\%-quantile predictions differed by more than 10\% were computed. The Figure shows the results for all models on all discretizations $n_D$.}
    \label{fig:acmodqauntilequality}
\end{figure}

However, the use of ensembles for predicting quantiles for individual examples was more mixed.
Figure~\ref{fig:acmodqauntilequality} shows the average proportion of test set examples where specific 80\%-quantile estimates from an ensemble trained on one dataset differed from the observed 80\%-quantile by more than 10\%.
In general, compared to the polynomial example, the ensemble predictions of the 80\%-quantile for specific examples tended to be better.
On low-discretizations in particular, the proportion of 80\%-quantile predictions that deviated by more than 10\% from the observed value was usually less than 1 out of 10.
Increasing the discretization $n_D$ tended to increase the proportion of inaccurate quantile predictions, with significant increases for DON and SUPN DONs at $n_D=32$ and FNOs at $n_D=128$

\begin{figure}[H]
    \centering
    \includegraphics[width=1.\linewidth,keepaspectratio]{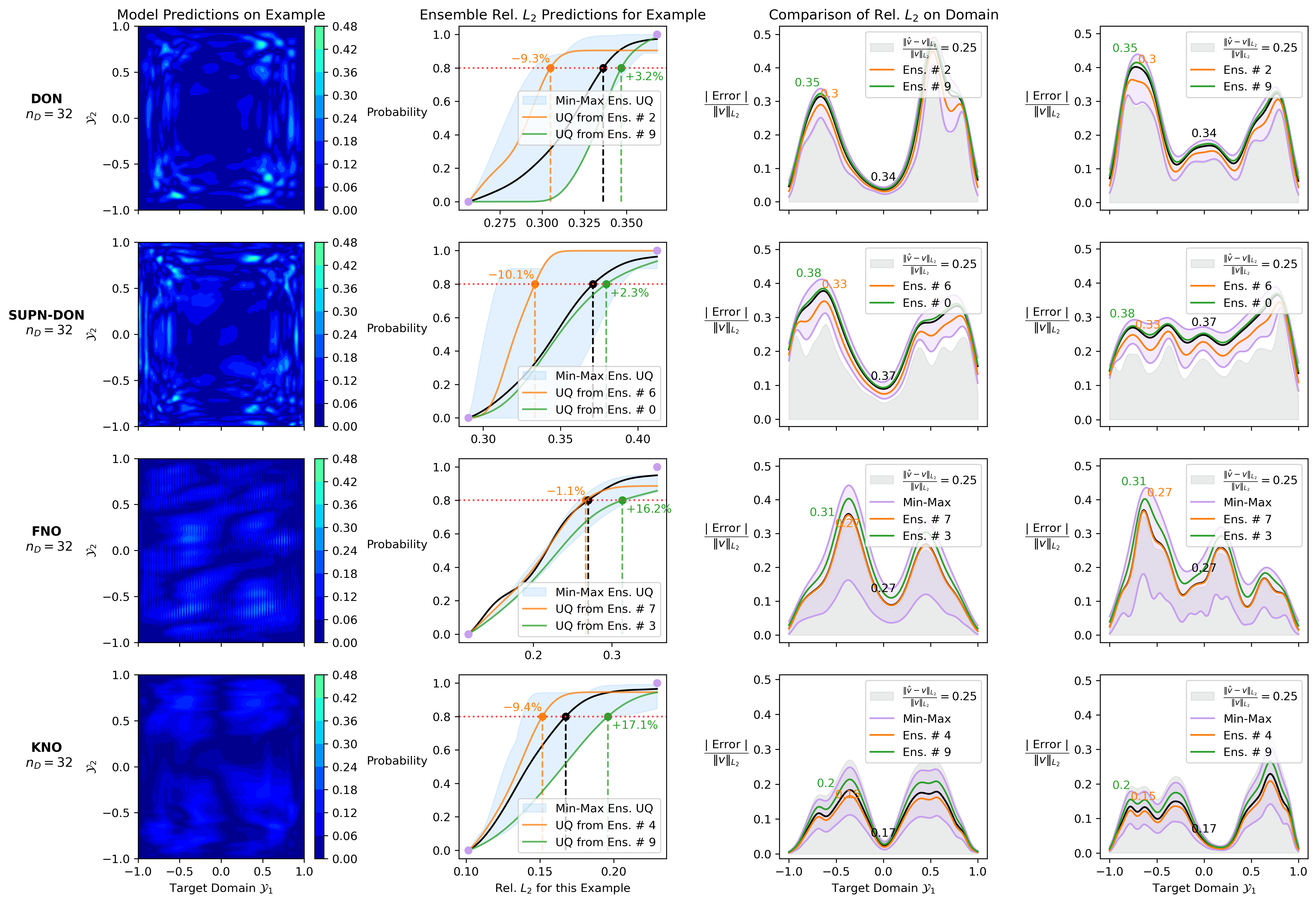}
    \caption{The \textbf{left column} shows the absolute pointwise error (relative to $\|v\|_{L_2}$) of a model prediction for a \emph{specific} randomly selected example from the test set. The model prediction was chosen to have a specific Rel. $L_2$ on this example $\approx$ the Rel. $L_2$ of the 80\%-quantile of all models on all datasets (i.e., the black dashed line in the secund column). The \textbf{second column} shows CDFs of the \emph{specific} Rel. $L_2$ error (i.e., the Rel. $L_2$ for this specific example): the true observed CDF computed over all models trained on all datasets (black), CDFs computed from two separate ensembles trained on single datasets (orange and green), and the min-max CDFs estimated from an ensemble (blue region). The plotted ensemble members were selected as the lower 10\% (orange) and upper 90\% quantile estimators of an 80\%-quantile prediction of the \emph{specific} Rel. $L_2$ for this example. The \textbf{last two columns} interpret the differences in 80\%-quantile predictions of the \emph{specific} Rel. $L_2$ by comparing it to a Rel. $L_2=0.25$ and the min-max \emph{specific} Rel. $L_2$ (for this example) observed across all models trained on all datasets (purple region). The Rel. $L_2$ is visualized as the integral (i.e., area) under a marginalized error distribution determined by integrating the error in one-dimension of the domain $\mathcal{Y}$. The two predictions of the 80\% quantile of the specific Rel. $L_2$ from the previously plotted ensemble CDFs (orange and green) are shown and compared to the true 80\% quantile from the full empirical CDF (black). The specific quantile predictions of the Rel. $L_2$ (i.e., the area under the curves) are labeled in the plots.}
    \label{fig:acmod_uqSPL2-ld}
\end{figure}

\begin{figure}[H]
    \centering
    \includegraphics[width=1.\linewidth,keepaspectratio]{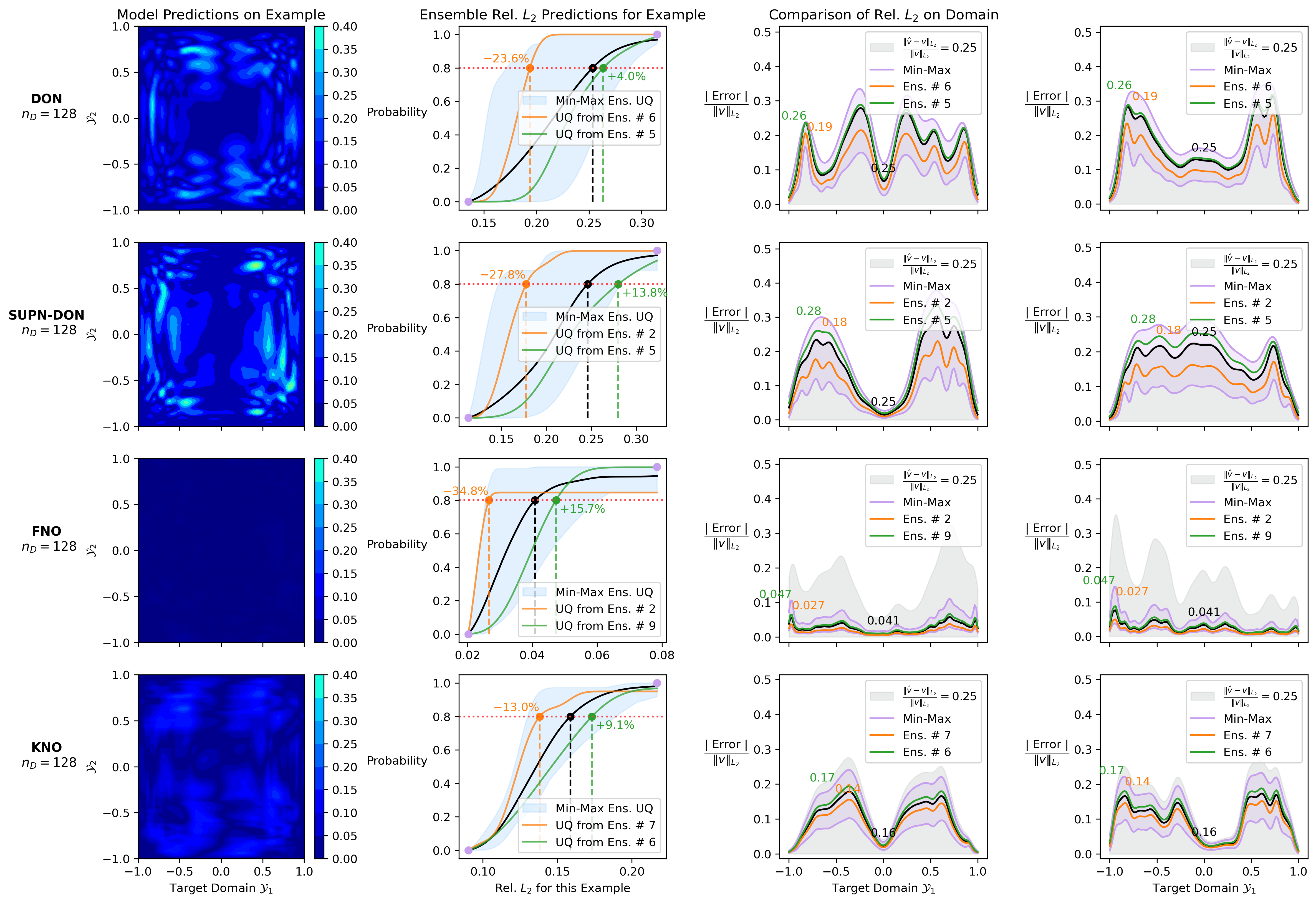}
    \caption{Ensemble estimates of the relative $L_2$ error
distribution for the same Allen--Cahn test example as
Figure~\ref{fig:acmod_uqSPL2-ld}, but with uniform discretization $n_D=128$
instead of $n_D=32$. Panel definitions and colors
follow Figure~\ref{fig:acmod_uqSPL2-ld}.}
    \label{fig:acmod_uqSPL2-hd}
\end{figure}

Figures~\ref{fig:acmod_uqSPL2-ld}
and~\ref{fig:acmod_uqSPL2-hd} illustrate
ensemble estimates of the relative $L_2$ error
quantile for the same Allen--Cahn test function
at $n_D=32$ and $n_D=128$, respectively.
The nonconverged FNO model at $n_D=32$
(initialization 2, dataset 8) was excluded.
The spatial error distributions illustrate
how discrepancies in ensemble quantile estimates
compare with a benchmark error of 0.25.

In Figure~\ref{fig:acmod_uqSPL2-hd}, with $n_D=128$, we see that utilizing an ensemble to make quantile inferences on the specific Rel. $L_2$ may be especially problematic for DONs and SUPN DONs: these ensembles sometimes severely underestimate the 80\% quantile to values far below $0.25$.
However, even though an ensemble of FNOs may also significantly underestimate the 80\% quantile, the magnitude of the errors for these models is so small compared to $0.25$ that the inaccurate prediction may not be important.

The key takeaway for us is that an ensemble approach to uncertainty provided reliable estimates for \overLtwo, in general.
However, for the assessment of the Rel. $L_2$ of \emph{specific} examples, the results are much more mixed.
Depending on the model type, discretization, training dataset, and the specific target example of interest, the ensemble estimator of 80\% quantiles could be signficantly different.
Our observations suggest that utilizing 80\% quantile estimates from ensembles trained on single datasets for an analysis of specific error on examples of interest should be conducted with caution.

\section{Conclusion}

The purpose of this study was to investigate the sources of uncertainty in the performance of learned operators, specifically to examine finite sampling of the function spaces, model initialization, and the effects of limited domain discretizations.
In both of our examples, we see that model initialization plays a significant role in the variability of overall model performance (as measured by the \overLtwo).
However, not all operator architectures are affected by model initializations in the same ways: indeed, different model architectures seem to induce different prior assumptions on the estimated operator, which lead to different forms of errors in predicted solution functions.
This was especially true when the training functions are weakly characterized by low sample discretizations.
On the other hand, we found that the differences between operators trained on different datasets (drawn from the same distribution) was much less of a factor in model training.
Our study does not rule out the significance of finite sampling of function spaces as a contributor to model performance.
Instead, it suggests that for the training / test distributions in our example, the limited sample size was large enough that overall model performance tended to generalize equivalently across training samples.

Where we did see a large influence of finite sampling of training functions is on ensemble measures of uncertainty.
Ensembles of trained models on the same dataset tended to provide similar measures of uncertainty in the overall model performance, i.e., a single ensemble trained on one dataset was a decent predictor of the overall performance of a randomly selected initialization trained on a different dataset.
However, using an ensemble to predict the performance of an operator on a specific example was much less reliable, especially in the polynomial example.
This makes sense: the measure of overall performance is the average of model performance over the entire set of functions, and each model initialization is optimized to perform well on all examples on average.
Thus, we should expect that for datasets which are comparable in an ``overall'' sense, a specific model initialization will behave similarly.
At the same time, different training datasets may contain different specific examples that a model initialization will over / under fit, leading to more significant variations between ensembles trained on different datasets for \emph{specific} examples.

Another major observation from our experiments is that the best performing model architecture is dependent on the problem and the training context.
When unstructured discretizations are required for training, FNOs and KNOs can exhibit poor performance if the method of interpolating the points to the kinds of structured grids required by the model architecture is poor.
On the other hand, SUPN DONs perform signficantly better if a separate randomized discretization for every sampled function is provided: the network is forced to regularize the representation of its predicted functions more effectively across the whole domain.

These findings demonstrate the importance of accounting
for multiple sources of epistemic uncertainty when
evaluating neural operators. Architecture selection
based on a single trained realization or discretization
may not reflect the accuracy and robustness of the
architecture under different training conditions.
Furthermore, initialization ensembles may provide
reliable estimates of aggregate operator error while
failing to characterize uncertainty in the relative
$L_2$ error for individual test functions. These
observations motivate more comprehensive benchmarking
practices and further investigation of uncertainty
quantification methods that explicitly account for
training function sampling, model initialization,
and discretization.

\acknowledgements
The research described in this paper was conducted under the auspices of Sandia National Laboratories (SNL, Contract No. DE-NA0003525), Lawrence Livermore National Laboratory (LLNL, Contract No. DE-AC52-07NA27344), and Pacific Northwest National Laboratory (PNNL, Contract No. DE-AC05--76RL01830) as part of the DOE Interlaboratory (IL) Laboratory Directed Research and Development (LDRD) Pilot Program. The LLNL LDRD project number is 25-ERD-052, and the release number is LLNL-JRNL-2026639. 
Any subjective views or opinions that might be expressed in the paper do not necessarily represent the views of the U.S. Department of Energy or the United States Government. The publisher acknowledges that the U.S. Government retains a non-exclusive, paid-up, irrevocable, worldwide license to publish or reproduce the published form of this written work or allow others to do so, for U.S. Government purposes. The DOE will provide public access to results of federally sponsored research in accordance with the DOE Public Access Plan.

\datastatement
The python scripts used to generate the data for the examples used in the computational experiments will be published in Zenodo, along with the $L_2$ losses of each of the models for every example on the corresponding test sets. Trained models and weights are available upon request.

\bibliographystyle{Bibliography_Style}
\bibliography{references}

\newpage
\appendix
\section{Training Configurations and Computational Costs}
\label{app:training}

This section provides additional implementation details and computational costs for the polynomial transformation and Allen--Cahn experiments. The main text describes the experimental design, while the configurations reported here specify the optimization procedures and computational requirements used for each architecture.

\subsection{Polynomial Transformation}
\label{app:poly_training}

All architectures were trained using 5000 Adam iterations with an initial learning rate of $10^{-3}$ and an exponential learning-rate scheduler with decay factor $\gamma=0.999$. Training was followed by L-BFGS refinement for a maximum of 5000 additional iterations, with gradient and function tolerances of $10^{-7}$ and $10^{-9}$, respectively.

Table~\ref{tab:poly1d-times} reports the number of trainable parameters, average training time per epoch, and mean overall relative $L_2$ error at the maximum output discretization.

\begin{table}[H]
\centering
\caption{Table of Training Timings on Polynomial Example with $n_D=150$. Average of \overLtwo taken across all model initializations / training sets.}
\label{tab:poly1d-times}
\begin{tabular}{llccc}
\toprule
 &  & Avg. Time per Epoch & \# of Params & Overall Rel. $L_2$ \\
\midrule
\multirow[t]{2}{*}{DON} & Uniform & 0.008 & 12030 & 0.123 \\
 & Random & 0.009 & 12030 & 0.132 \\
\cline{1-5}
\multirow[t]{2}{*}{SUPN-DON} & Uniform & 0.008 & 11040 & 0.063 \\
 & Random & 0.008 & 11040 & 0.065 \\
\cline{1-5}
\multirow[t]{2}{*}{FNO} & Uniform & 0.016 & 6633 & 0.007 \\
 & Random & 0.016 & 6633 & 0.015 \\
\cline{1-5}
\multirow[t]{2}{*}{KNO} & Uniform & 0.054 & 6745 & 0.018 \\
 & Random & 0.056 & 6745 & 0.018 \\
\cline{1-5}
\bottomrule
\end{tabular}
\end{table}

\subsection{Allen--Cahn Equation}
\label{app:ac_training}

DeepONets, SUPN-DeepONets, and KNOs were trained using SOAP, while FNOs were trained using Adam with cosine-annealing learning-rate scheduling. The different optimizer choice for FNOs resulted from limitations of the SOAP implementation used in this study with complex-valued operations. All architectures were trained for a maximum of 5000 epochs with an initial learning rate of $2\times10^{-3}$.

Table~\ref{tab:acmod-times} reports the trainable parameter counts, average training times per epoch, and mean overall relative $L_2$ errors at the maximum output discretization.

\begin{table}[H]
\centering
\caption{Table of Training Timings for the Allen-Cahn example with the full discretization $n_D=128\times128$. The timings are the average time per epoch: note that DON, SUPN-DON, and KNOs utilized the higher-order SOAP optimizer. Average of \overLtwo on the test set taken across all model initializations / training sets.}
\label{tab:acmod-times}
\begin{tabular}{llccc}
\toprule
 &  & Avg. Time per Epoch & \# of Params & \makecell{Overall Rel. $L_2$ \\ with $n_D=128$} \\
\midrule
\multirow[t]{2}{*}{DON} & Uniform & 0.016 & 2507585 & 0.223 \\
 & Random & 0.051 & 2507585 & 0.224 \\
\cline{1-5}
\multirow[t]{2}{*}{SUPN-DON} & Uniform & 0.017 & 3153217 & 0.232 \\
 & Random & 0.074 & 3153217 & 0.232 \\
\cline{1-5}
\multirow[t]{2}{*}{FNO} & Uniform & 0.078 & 7818449 & 0.024 \\
 & Random & 0.078 & 7818449 & 0.024 \\
\cline{1-5}
\multirow[t]{2}{*}{KNO} & Uniform & 0.018 & 20929 & 0.109 \\
 & Random & 0.021 & 20929 & 0.109 \\
\cline{1-5}
\bottomrule
\end{tabular}
\end{table}

\section{Representative Allen--Cahn Forcing Functions and Predictions}
\label{app:ac_examples}

\begin{figure}[H]
    \centering
    \includegraphics[width=0.75\textwidth,height=0.48\textheight,keepaspectratio]{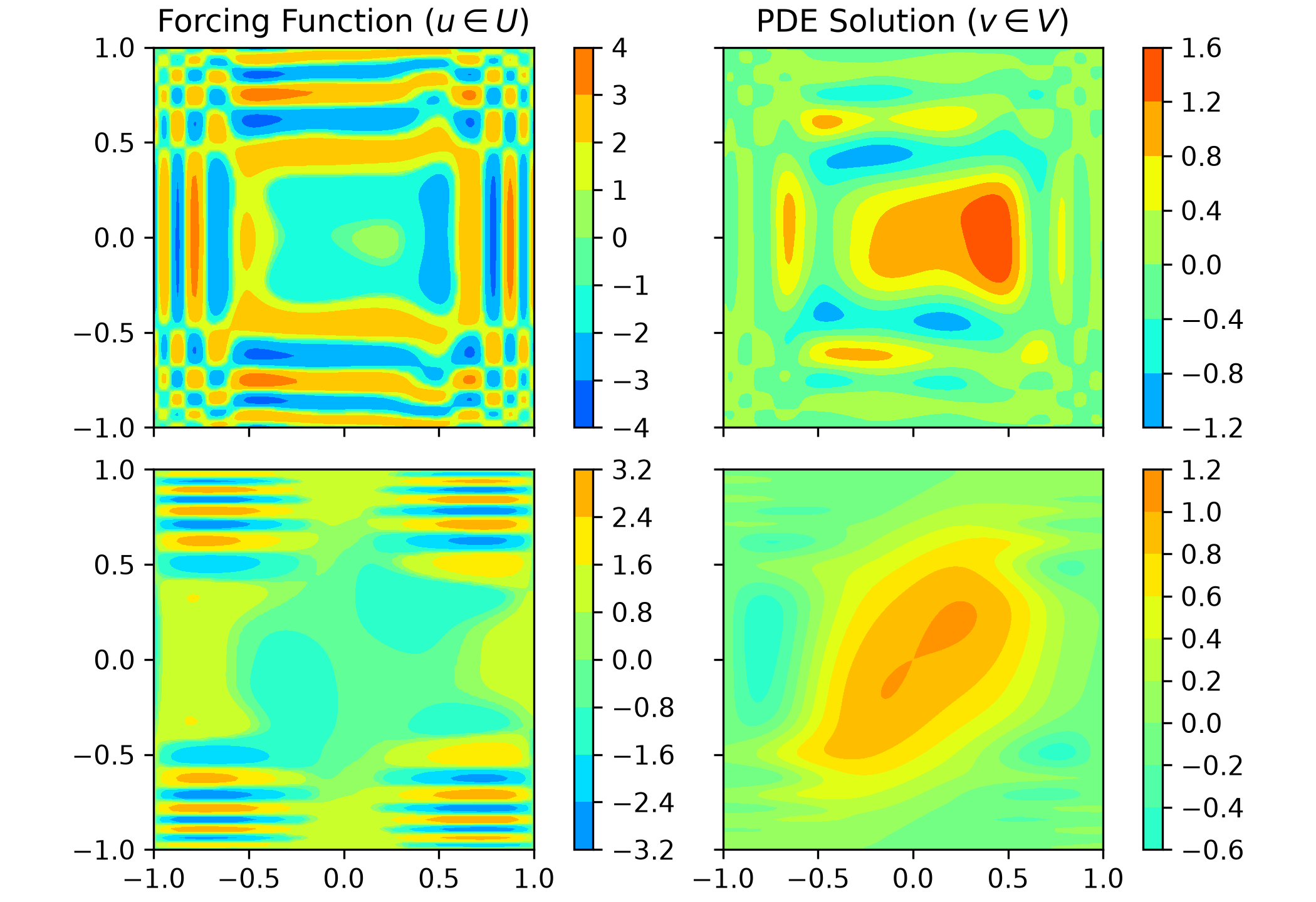}
    \caption{Shows two training examples and their corresponding solutions plots: the forcing functions $u\in U$ (left) and the PDE solutions $v\in V$ (right).}
    \label{fig:ac_examples}
\end{figure}

\begin{figure}[H]
    \centering
    \includegraphics[width=0.9\linewidth,keepaspectratio]{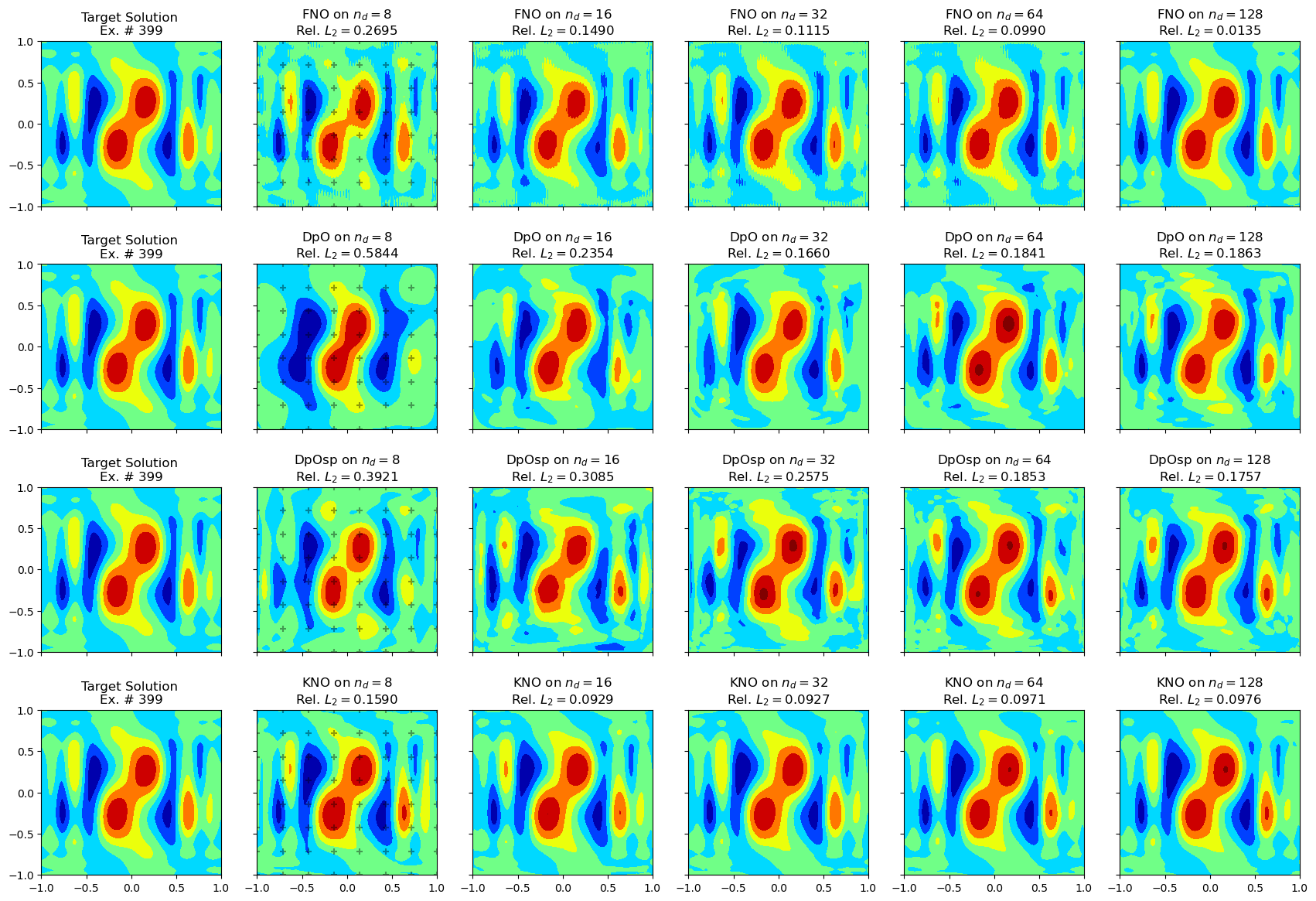}
    \caption{Representative Allen--Cahn solution predictions from DeepONet, SUPN--DeepONet, FNO, and KNO models trained on uniform output discretizations. The first column shows the target solution; subsequent columns show predictions at increasing training resolutions.}
    \label{fig:ac_unifEX}
\end{figure}

\begin{figure}[H]
    \centering
    \includegraphics[width=0.9\linewidth,keepaspectratio]{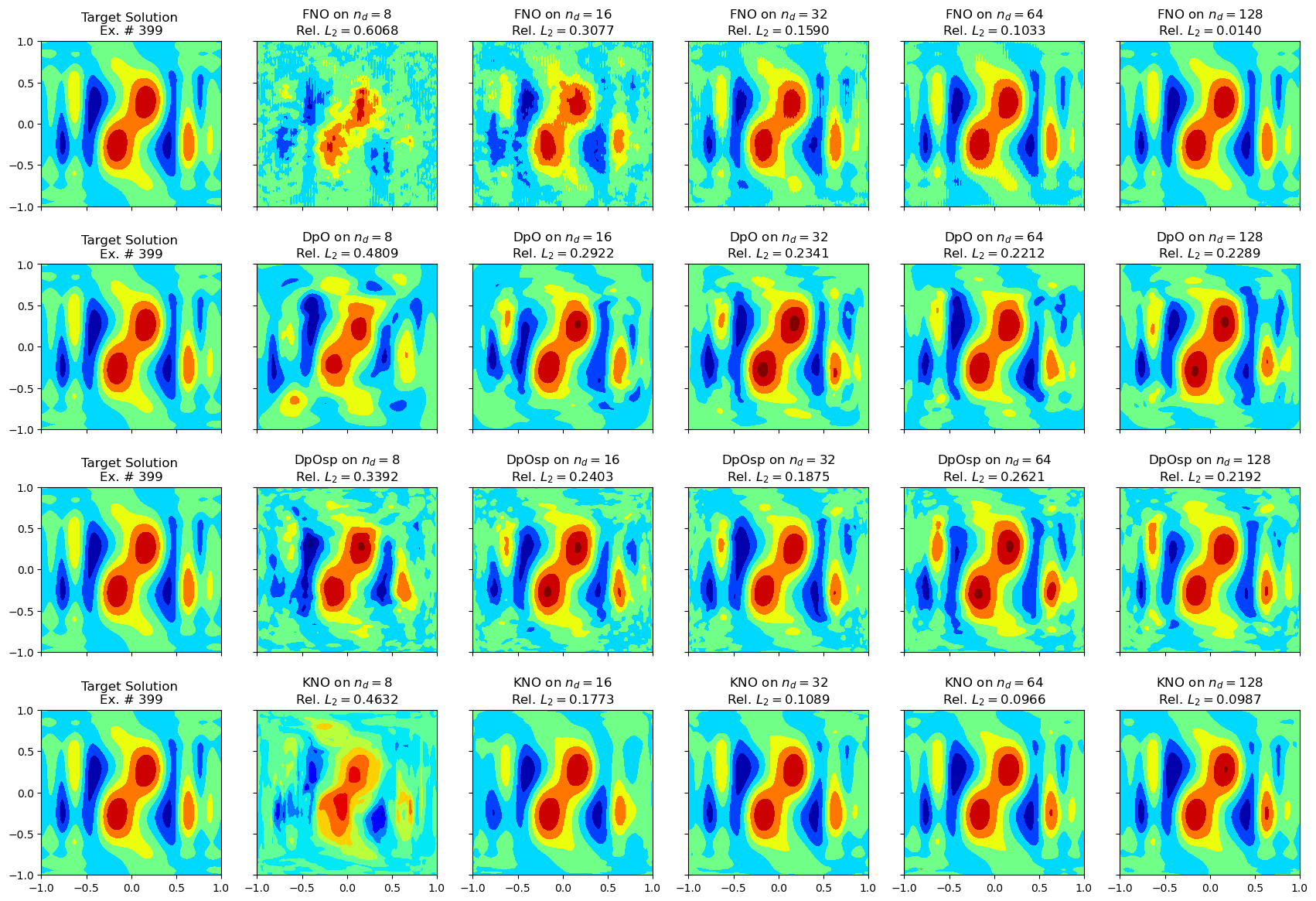}
    \caption{Representative Allen--Cahn solution predictions from DeepONet, SUPN--DeepONet, FNO, and KNO models trained on random output discretizations. The first column shows the target solution; subsequent columns show predictions at increasing training resolutions.}
    \label{fig:ac_randEX}
\end{figure}

This section provides representative forcing functions and solutions
for the nonlinear Allen--Cahn problem, along with predicted solution
fields from different neural operator architectures. Figure~\ref{fig:ac_examples}
illustrates the spatial structure of the generated forcing functions
and their corresponding solutions. Figures~\ref{fig:ac_unifEX}
and~\ref{fig:ac_randEX} compare neural operator predictions
across uniform and random output discretizations, respectively.
These examples supplement the aggregate accuracy and variability
results presented in Section~\ref{sec:ac_results}.

\section{Nonlinear Polynomial UQ Experiment}\label{app:A}

This section provides a comparison of UQ figures discussed in Section~\ref{sec:poly1dUQ} for all uniform discretizations.
These figures describe the robustness of ensemble UQ methods for both the \overLtwo and the relative $L_2$ for specific examples.

\subsection{Ensemble UQ for \overLtwo}
For each discretization, we compare the empirical cumulative distribution function (CDF) of the overall relative $L_2$
error across all trained models with CDFs estimated from initialization ensembles trained on individual datasets. The shaded region represents the range of ensemble CDFs,
while the selected curves illustrate the variability in estimates of the 80\% error quantile. The right column compares these quantile estimates with a reference error
threshold.

\begin{figure}[H]
    \centering \captionsetup{width=0.725\textwidth}\includegraphics[width=0.725\linewidth,keepaspectratio]{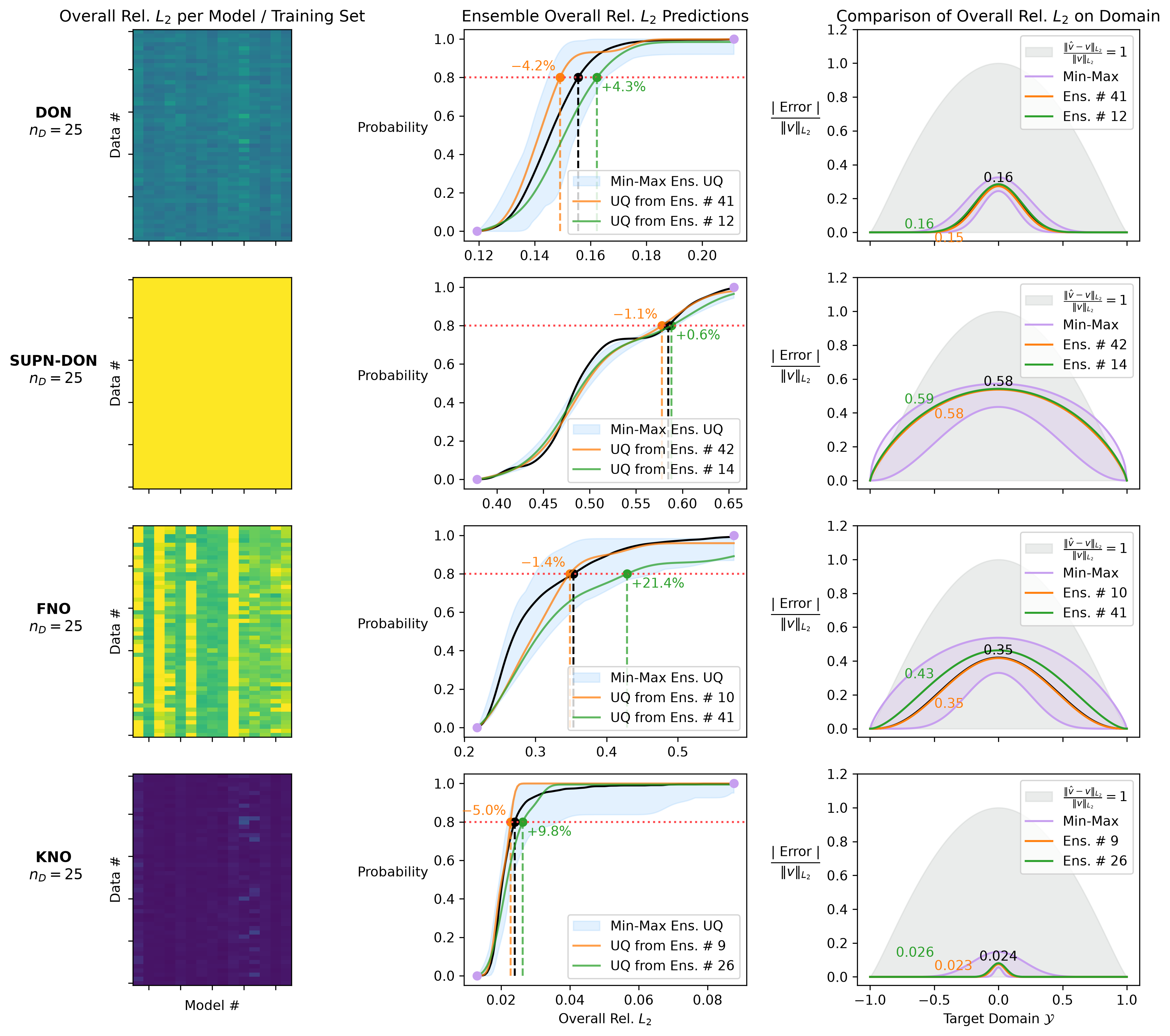}
    \caption{Ensemble estimates of the overall relative $L_2$ error distribution for the polynomial problem with $n_D=25$.}
    \label{fig:poly1d_uqOverL2-25}
\end{figure}

In general, the ensemble predictions for the \overLtwo were fairly well calibrated across all models trained on all uniform discretizations. The two scenarios where ensemble UQ predictions of the \overLtwo showed potentially large overestimates for the 80\%-quantile were FNOs and KNOs on discretizations with $n_D=10$ (Figure~\ref{fig:poly1d_uqOverL2-ld}) and $n_D=25$ (Figure~\ref{fig:poly1d_uqOverL2-25}).

\begin{figure}[H]
    \centering \captionsetup{width=0.725\textwidth}
\includegraphics[width=0.725\linewidth,keepaspectratio]{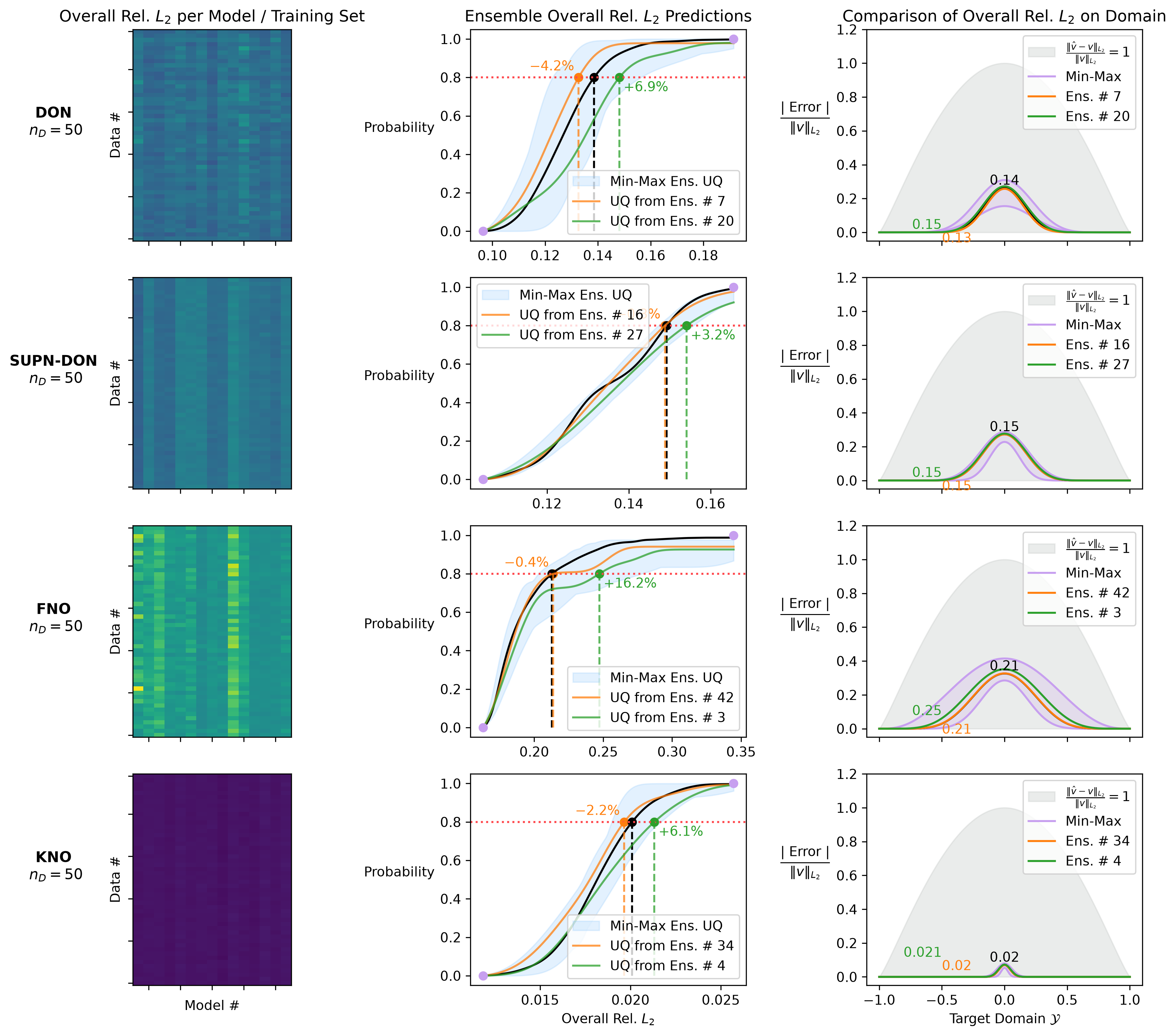}
    \caption{Ensemble estimates of the overall relative $L_2$ error distribution for the polynomial problem with $n_D=50$.}
    \label{fig:poly1d_uqOverL2-50}
\end{figure}

\begin{figure}[H]
    \centering \captionsetup{width=0.725\textwidth}\includegraphics[width=0.725\linewidth,keepaspectratio]{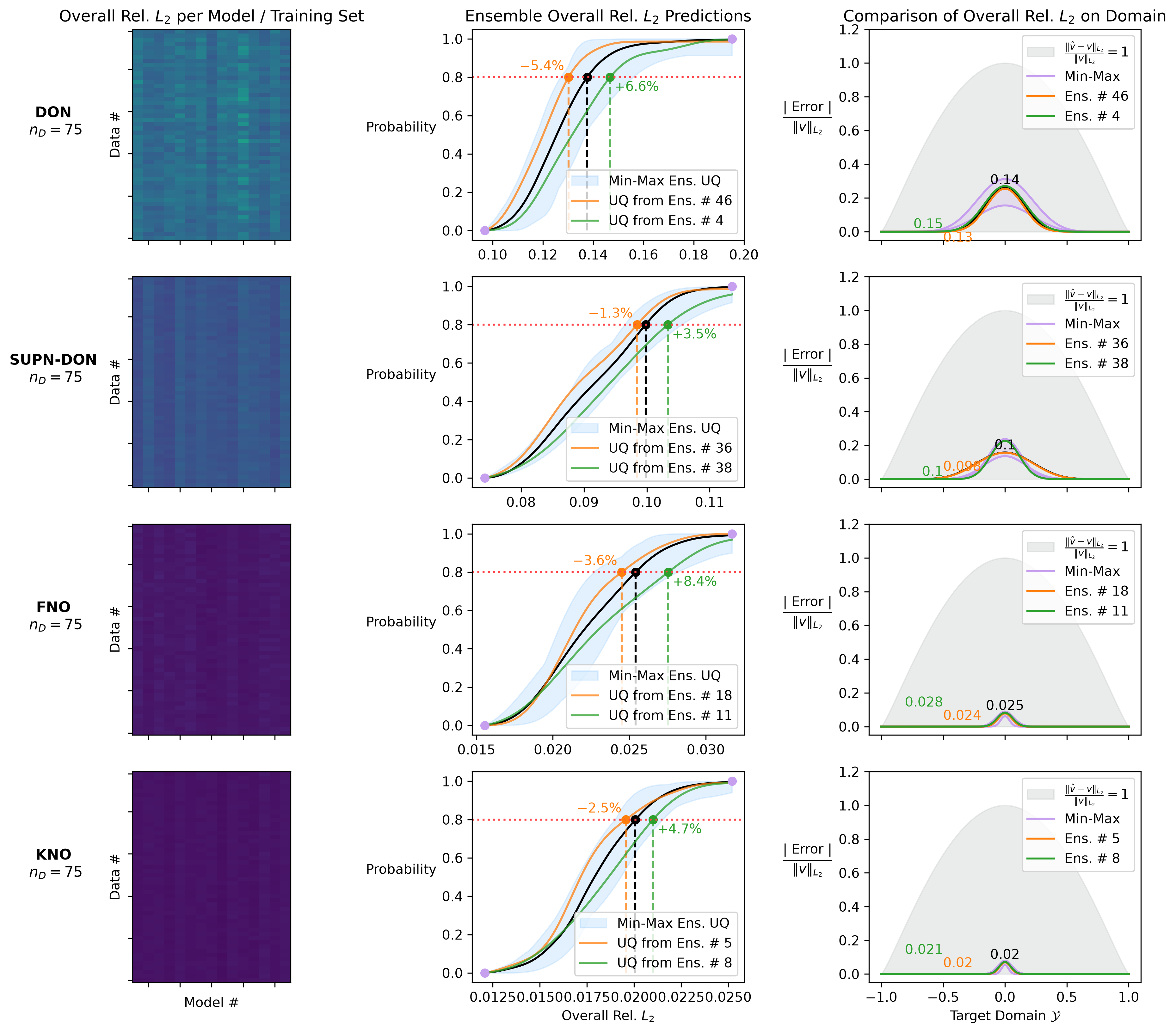}
    \caption{Ensemble estimates of the overall relative $L_2$ error distribution for the polynomial problem with $n_D=75$.}
    \label{fig:poly1d_uqOverL2-75}
\end{figure}

\begin{figure}[H]
    \centering
    \captionsetup{width=0.725\textwidth} \includegraphics[width=0.725\linewidth,keepaspectratio]{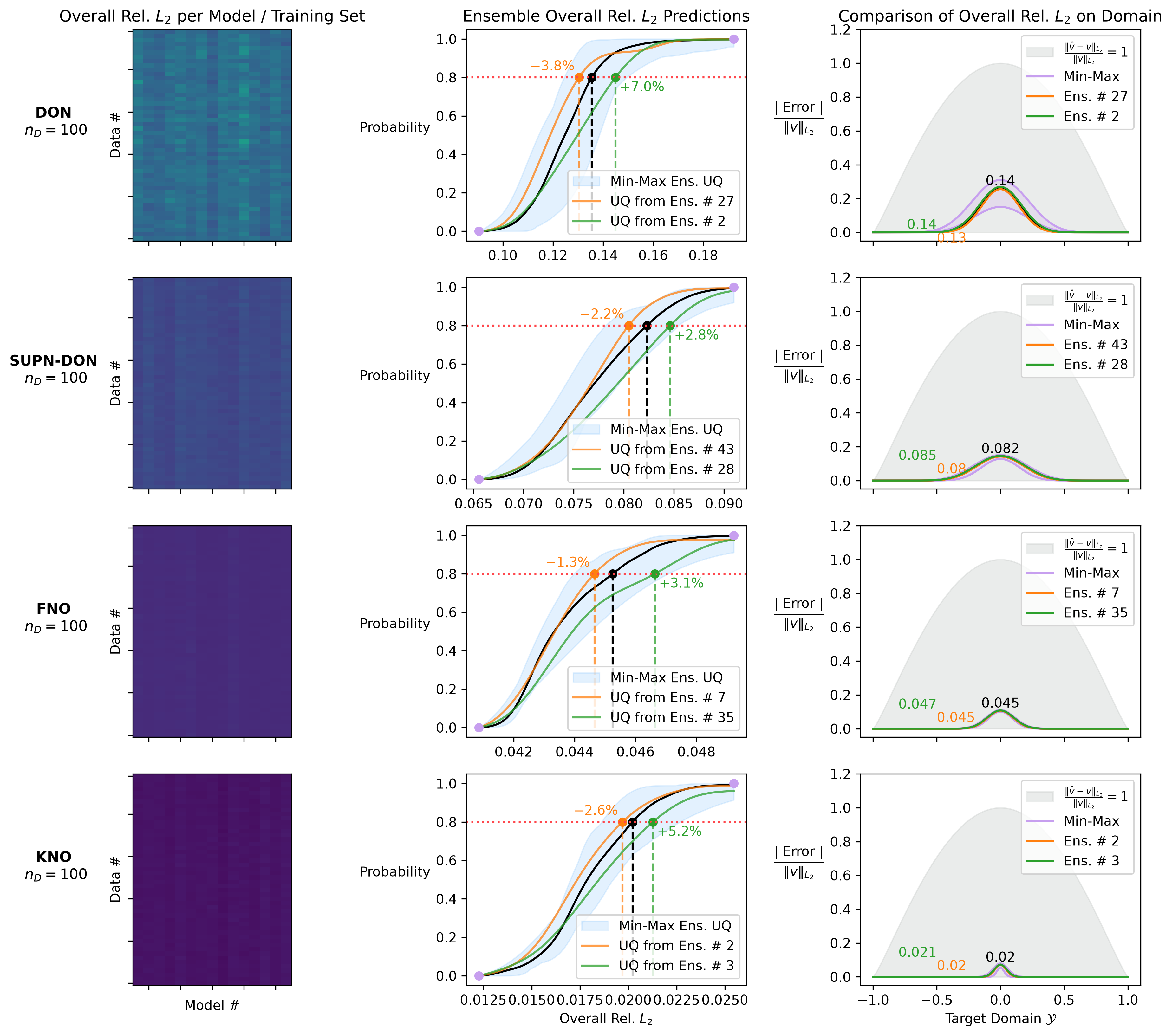}
    \caption{Ensemble estimates of the overall relative $L_2$ error distribution for the polynomial problem with $n_D=100$.}
    \label{fig:poly1d_uqOverL2-100}
\end{figure}

\begin{figure}[H]
    \centering
    \captionsetup{width=0.725\textwidth} \includegraphics[width=0.725\linewidth,keepaspectratio]{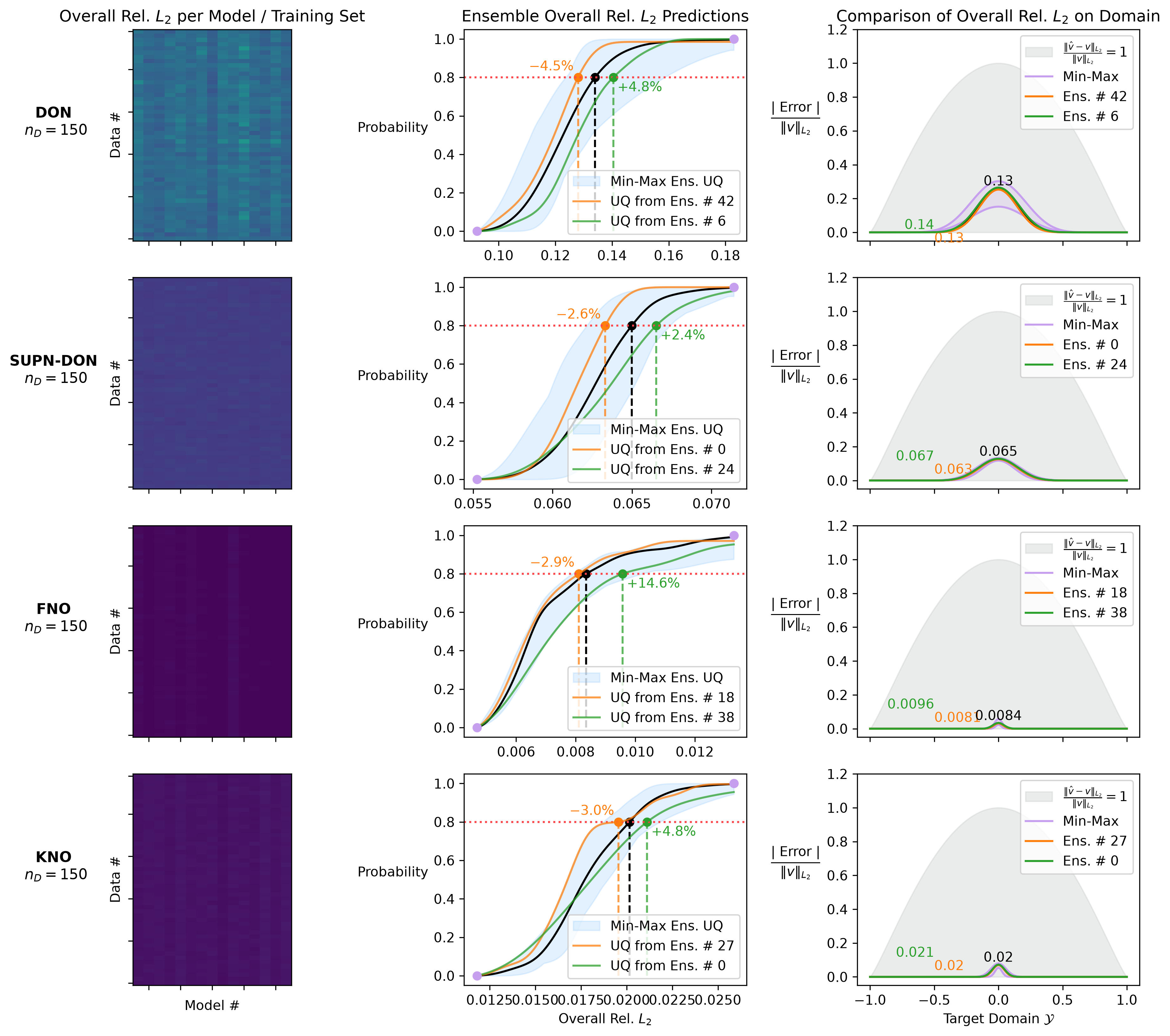}
    	\caption{Ensemble estimates of the overall relative $L_2$ error distribution for the polynomial problem with $n_D=150$.}
    \label{fig:poly1d_uqOverL2-150}
\end{figure}

\subsection{Ensemble UQ for Relative $L_2$ on a Specific Example}\label{app:poly_relL24ex}

\begin{figure}[H]
    \centering
    \includegraphics[width=1\linewidth,keepaspectratio]{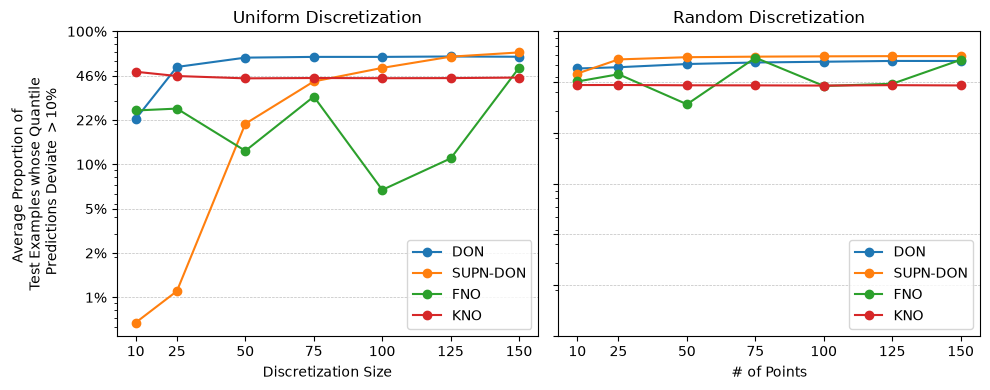}
	\caption{Fraction of test examples for which ensemble estimates of the 80\% relative $L_2$ error quantile differ from the empirical quantile by more than 10\%, across uniform and random discretizations.}
    \label{fig:poly1d_uq4ex_eval}
\end{figure}

In this Section, we show the corresponding Figures for the ensemble UQ estimates for the relative $L_2$ error on a specific example.
An ensemble approach to estimating the relative $L_2$ for a specific example tended to be much less robust than the \overLtwo.
As evidence for this claim, Figure~\ref{fig:poly1d_uq4ex_eval} shows the average proportion of examples where the ensemble UQ predictions of the 80\%-quantile devaiated from the actual 80\%-quantile by more than 10\% on both uniform and random discretizations.
With the exception of SUPN-DONs trained on a uniform discretization $n_D=10$ and $n_D=25$, a large proportion of the test set showed noticeable deviations in the accuracy of the 80\%-quantile ensemble predictions.

For a representative test function, we compare the
empirical distribution of relative $L_2$ errors across
all trained models with distributions estimated from
initialization ensembles trained on individual datasets.
The figures also show the predicted solution, pointwise
uncertainty intervals, and differences in the estimated
80\% error quantiles.

\begin{figure}[H]
    \centering
    \captionsetup{width=0.7\textwidth} \includegraphics[width=0.7\linewidth,keepaspectratio]{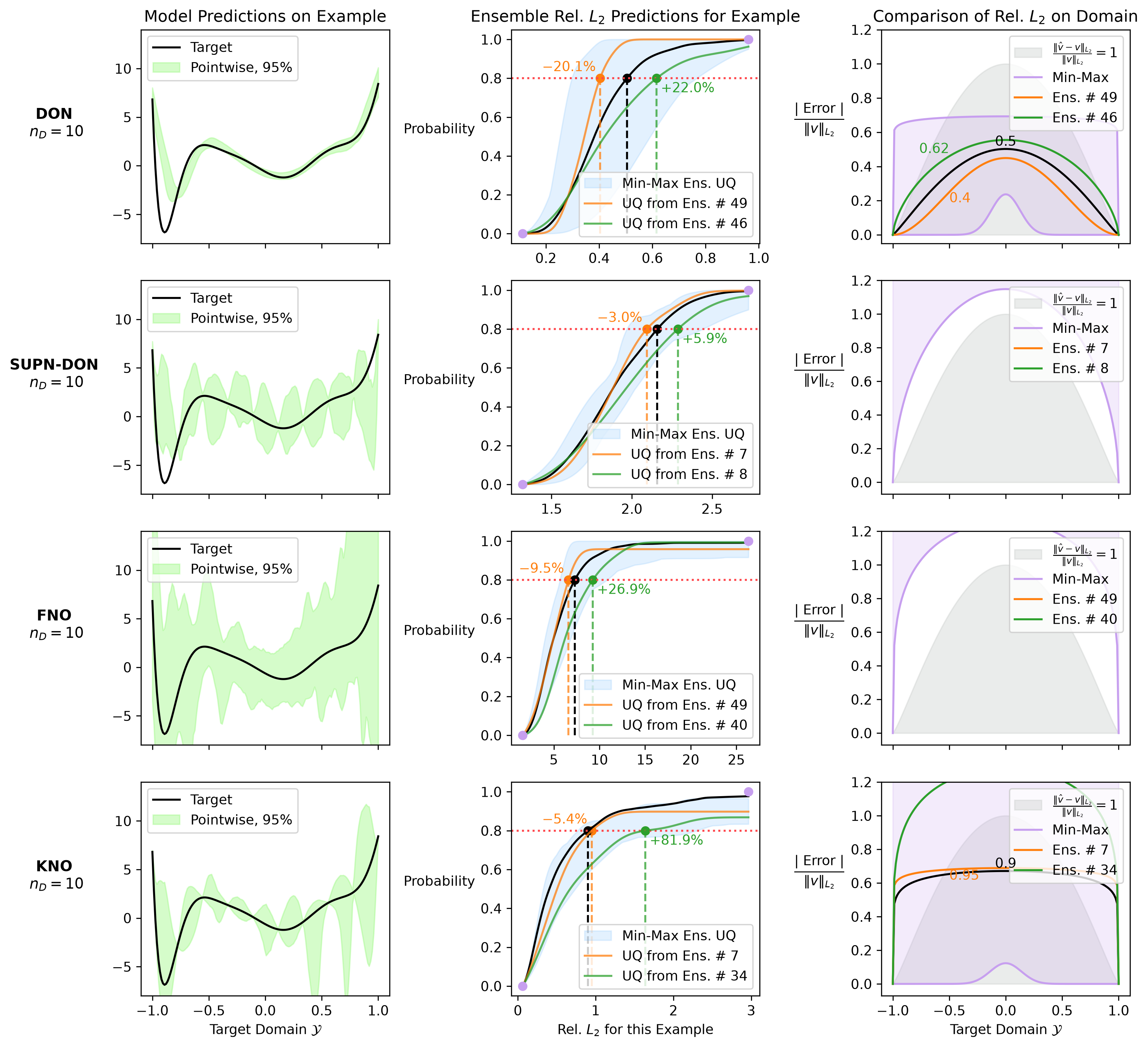}
    \caption{Ensemble uncertainty estimates for a representative polynomial test function with $n_D=10$.}
    \label{fig:poly1d_uqSPL2-10}
\end{figure}

\begin{figure}[H]
    \centering
    \captionsetup{width=0.7\textwidth} \includegraphics[width=0.7\linewidth,keepaspectratio]{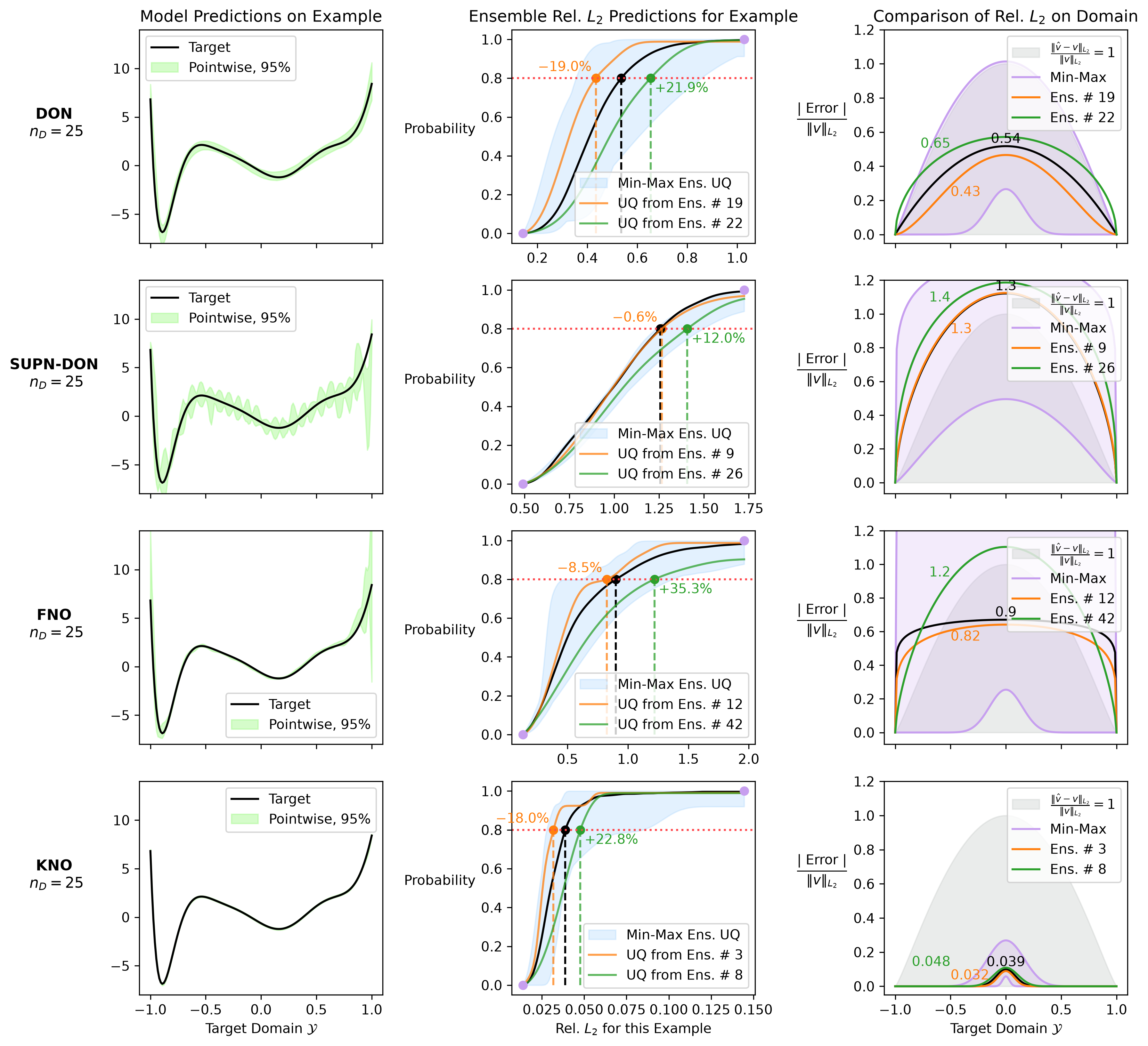}
    \caption{Ensemble uncertainty estimates for a representative polynomial test function with $n_D=25$.}
    \label{fig:poly1d_uqSPL2-25}
\end{figure}

\begin{figure}[H]
    \centering
    \captionsetup{width=0.7\textwidth} \includegraphics[width=0.7\linewidth,keepaspectratio]{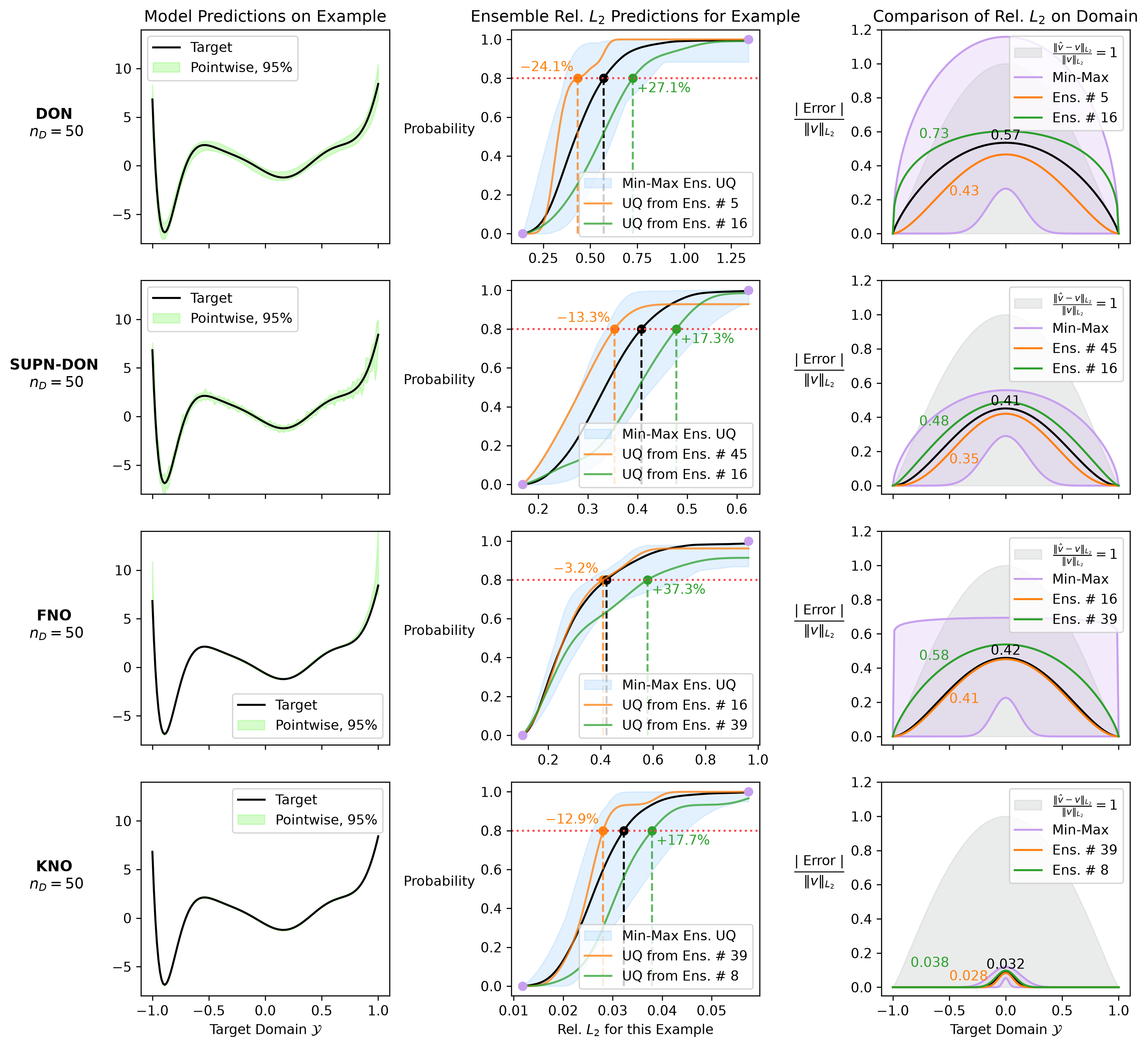}
    \caption{Ensemble uncertainty estimates for a representative polynomial test function with $n_D=50$.}
    \label{fig:poly1d_uqSPL2-50}
\end{figure}

\begin{figure}[H]
    \centering
    \captionsetup{width=0.7\textwidth} \includegraphics[width=0.7\linewidth,keepaspectratio]{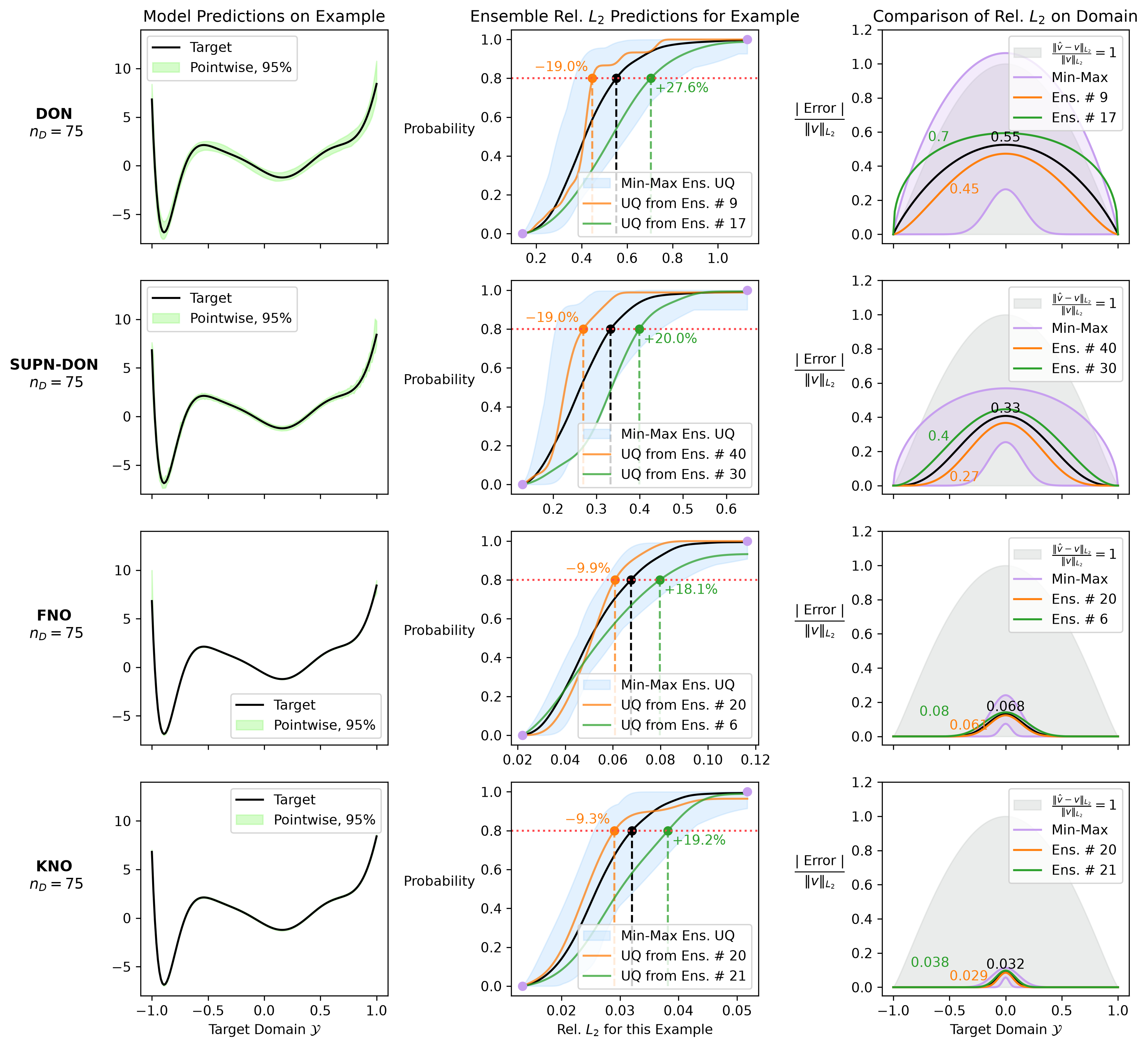}
    \caption{Ensemble uncertainty estimates for a representative polynomial test function with $n_D=75$.}
    \label{fig:poly1d_uqSPL2-75}
\end{figure}

\begin{figure}[H]
    \centering
    \captionsetup{width=0.7\textwidth} \includegraphics[width=0.7\linewidth,keepaspectratio]{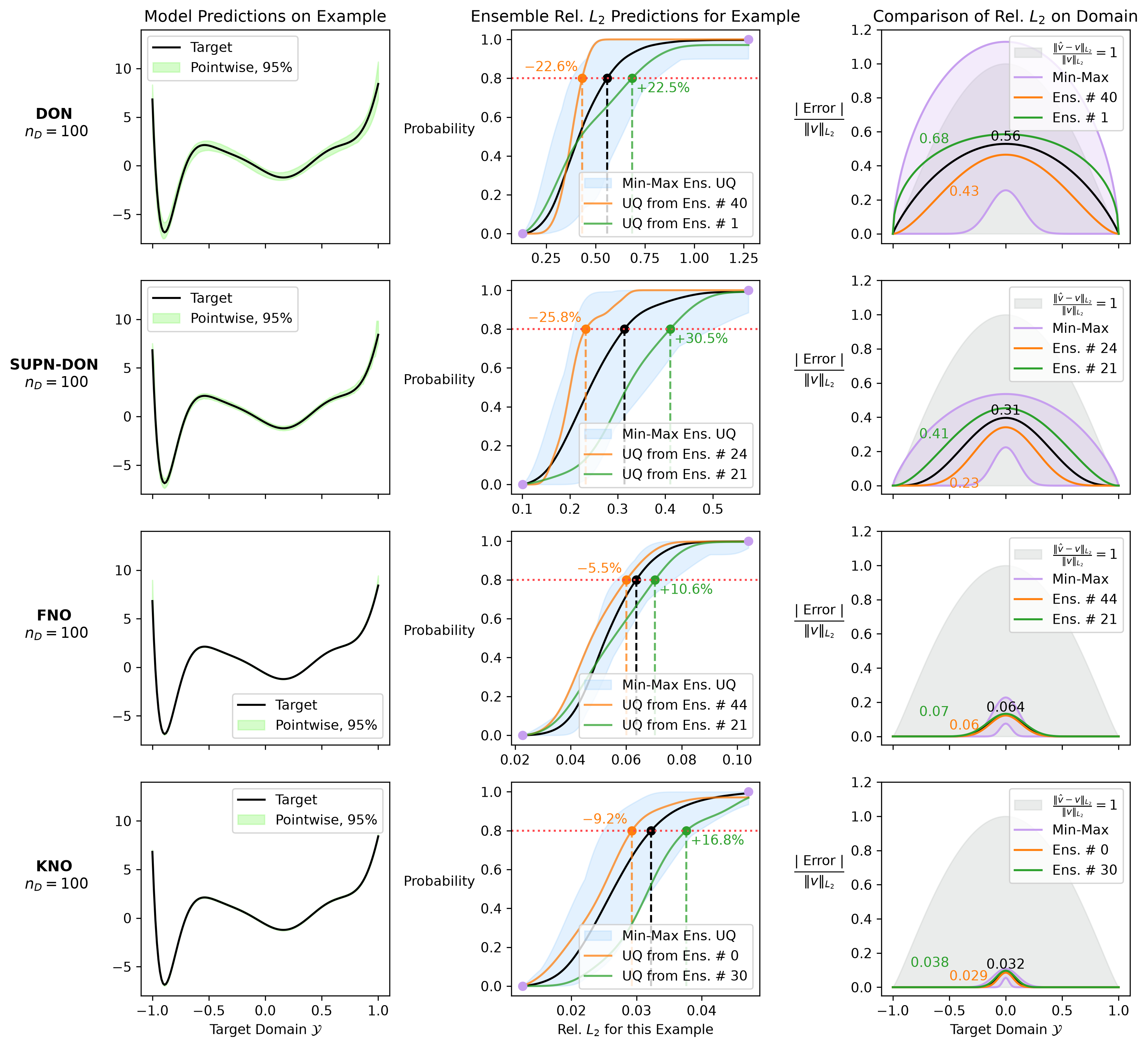}
	\caption{Ensemble uncertainty estimates for a representative polynomial test function with $n_D=100$.}
    \label{fig:poly1d_uqSPL2-100}
\end{figure}

\begin{figure}[H]
    \centering
    \captionsetup{width=0.7\textwidth} \includegraphics[width=0.7\linewidth,keepaspectratio]{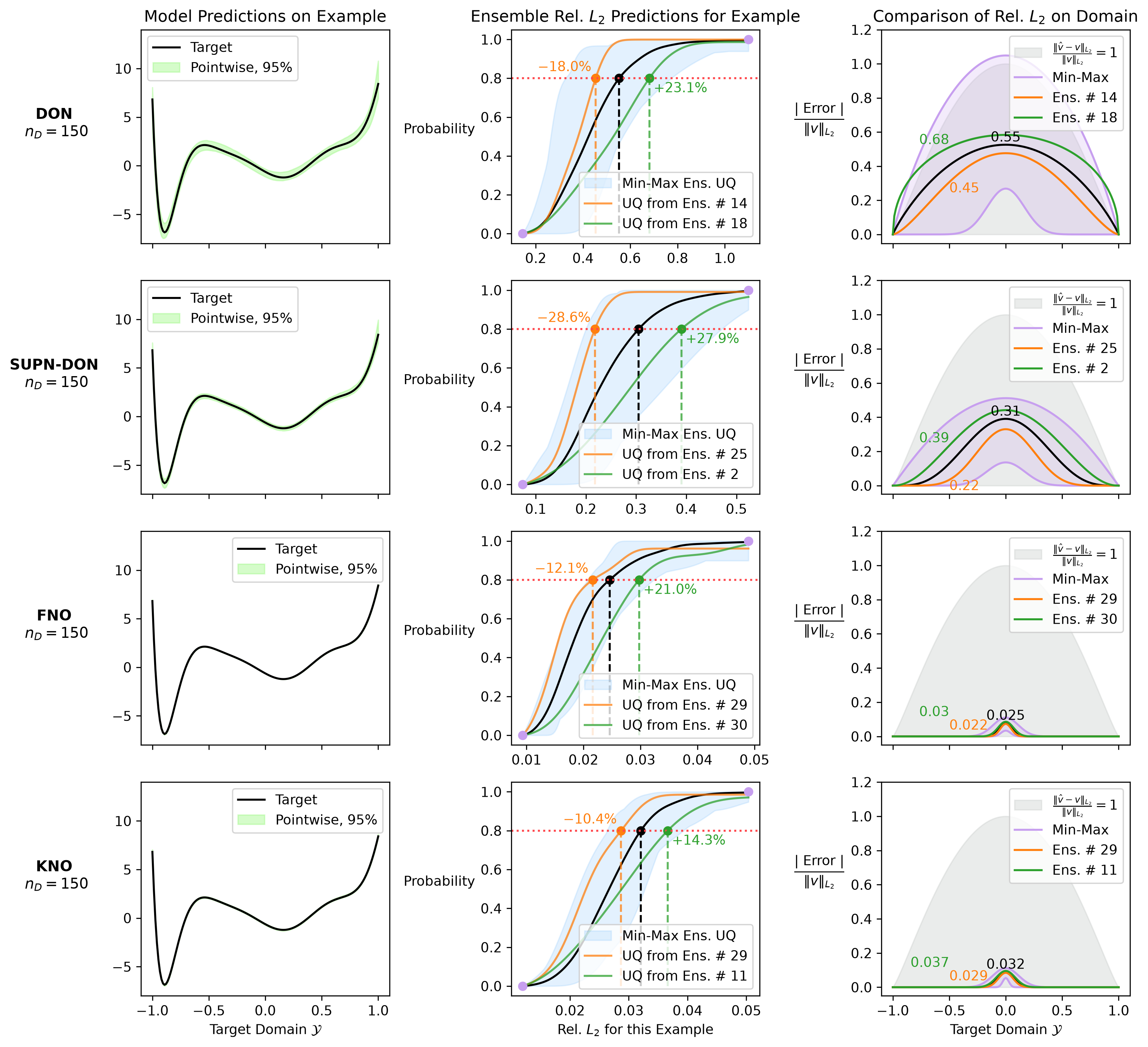}
    \caption{Ensemble uncertainty estimates for a representative polynomial test function with $n_D=150$.}
    \label{fig:poly1d_uqSPL2-150}
\end{figure}

\section{Allen-Cahn UQ Experiment}\label{app:acuq}
This section provides a comparison of UQ figures discussed in Section~\ref{sec:acuq} for all uniform discretizations.
These figures describe the robustness of ensemble UQ methods for both the \overLtwo and the relative $L_2$ for specific examples.

\subsection{Ensemble UQ for \overLtwo}
For each discretization, we compare the empirical cumulative
distribution function (CDF) of the overall relative $L_2$
error across all trained models with CDFs estimated from
initialization ensembles trained on individual datasets.
The shaded region represents the range of ensemble CDFs,
while the selected curves illustrate differences in estimates
of the 80\% error quantile. The right columns visualize
these quantile estimates relative to a benchmark overall
relative $L_2$ error of 0.25, using a representative
spatial error distribution.

In general, the ensemble predictions for the \overLtwo were fairly well calibrated across all models trained on all uniform discretizations except for KNOs on $n_D=8$. An ensemble of FNOs trained on $n_D=32$ also sometimes lead to overestimates of \overLtwo quantiles, even when removing the outlier model which did not converge.

\begin{figure}[H]
    \centering
    \captionsetup{width=0.9\textwidth} \includegraphics[width=0.9\linewidth,keepaspectratio]{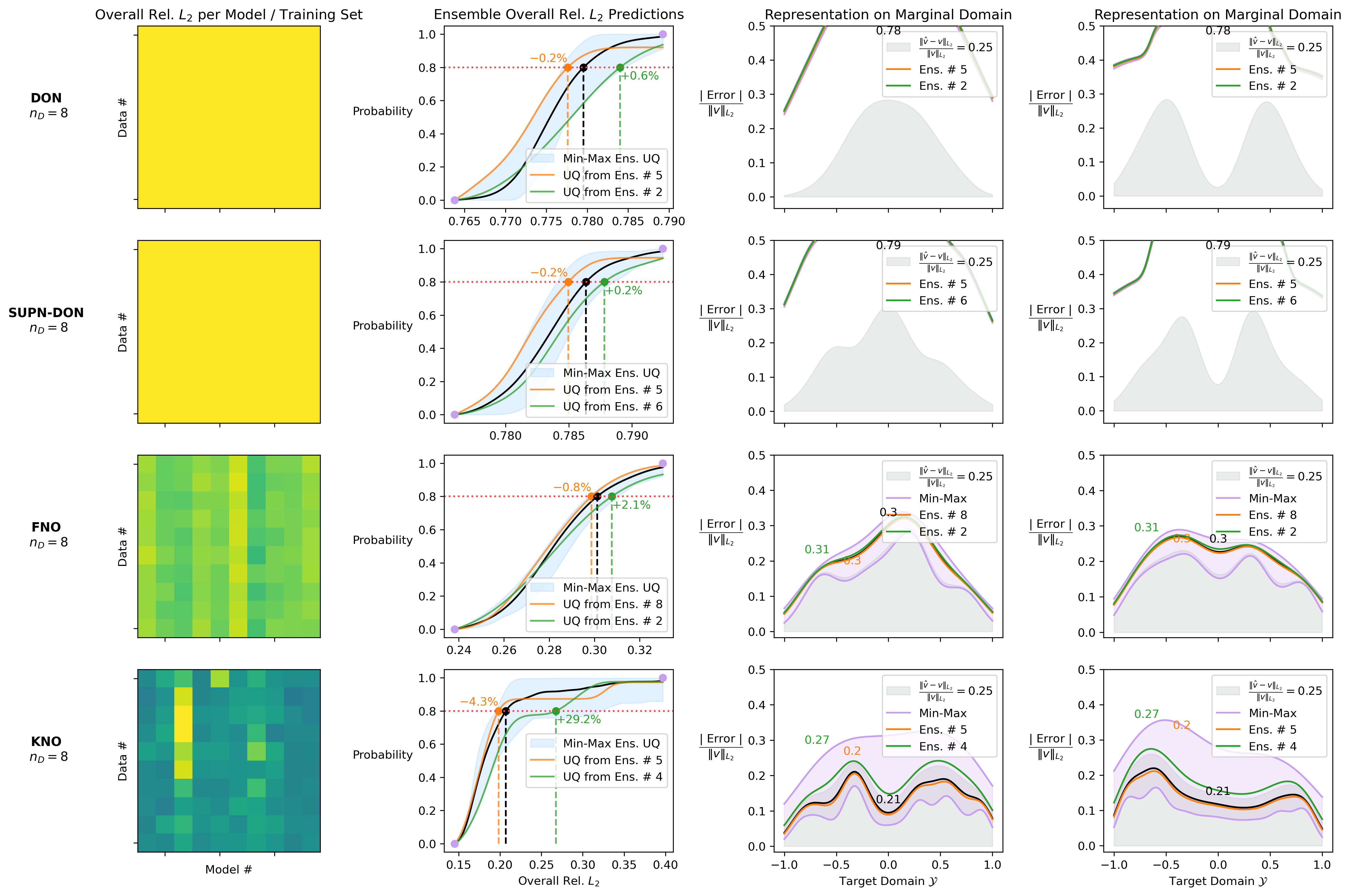}
    \caption{Ensemble estimates of the overall relative $L_2$ error distribution for the Allen--Cahn problem with $n_D=8$.}
    \label{fig:acmod_uqover_8}
\end{figure}

\begin{figure}[H]
    \centering
    \captionsetup{width=0.9\textwidth} \includegraphics[width=0.9\linewidth,keepaspectratio]{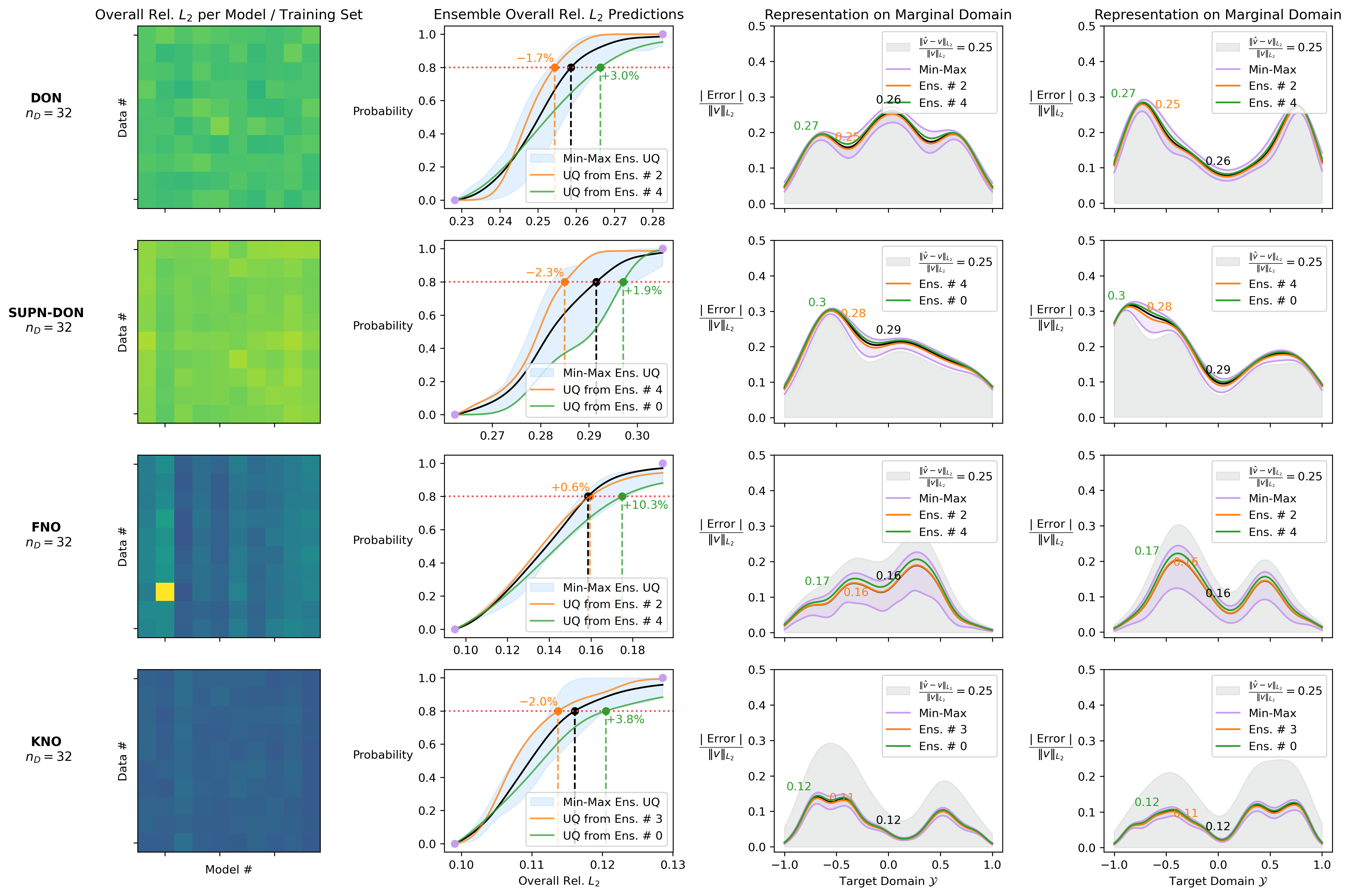}
	\caption{Ensemble estimates of the overall relative $L_2$ error distribution for the Allen--Cahn problem with $n_D=32$.}
    \label{fig:acmod_uqover_32}
\end{figure}

\begin{figure}[H]
    \centering
    \captionsetup{width=0.9\textwidth} \includegraphics[width=0.9\linewidth,keepaspectratio]{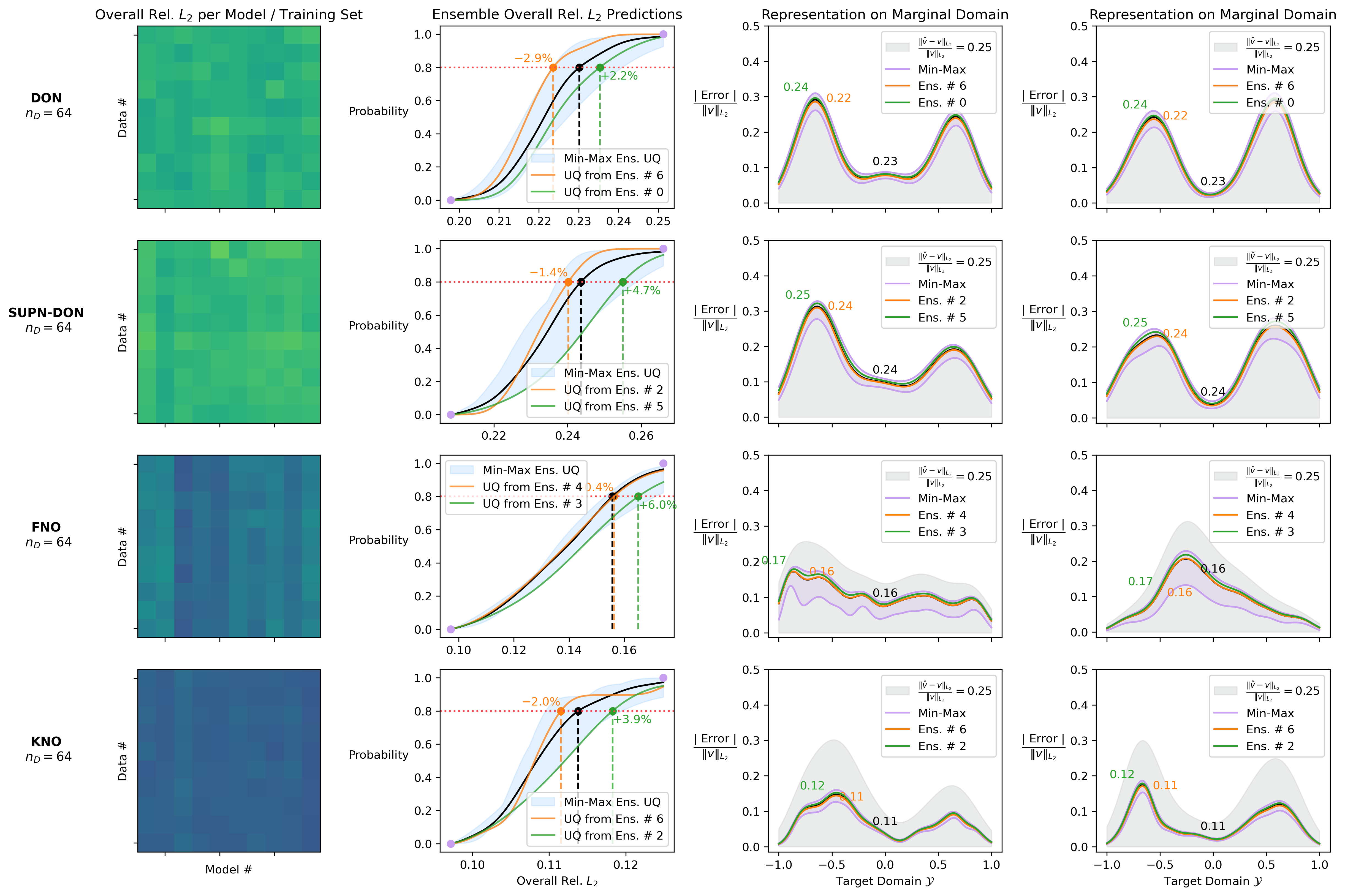}
	\caption{Ensemble estimates of the overall relative $L_2$ error distribution for the Allen--Cahn problem with $n_D=64$.}
    \label{fig:acmod_uqover_64}
\end{figure}

\begin{figure}[H]
    \centering
    \captionsetup{width=0.9\textwidth} \includegraphics[width=0.9\linewidth,keepaspectratio]{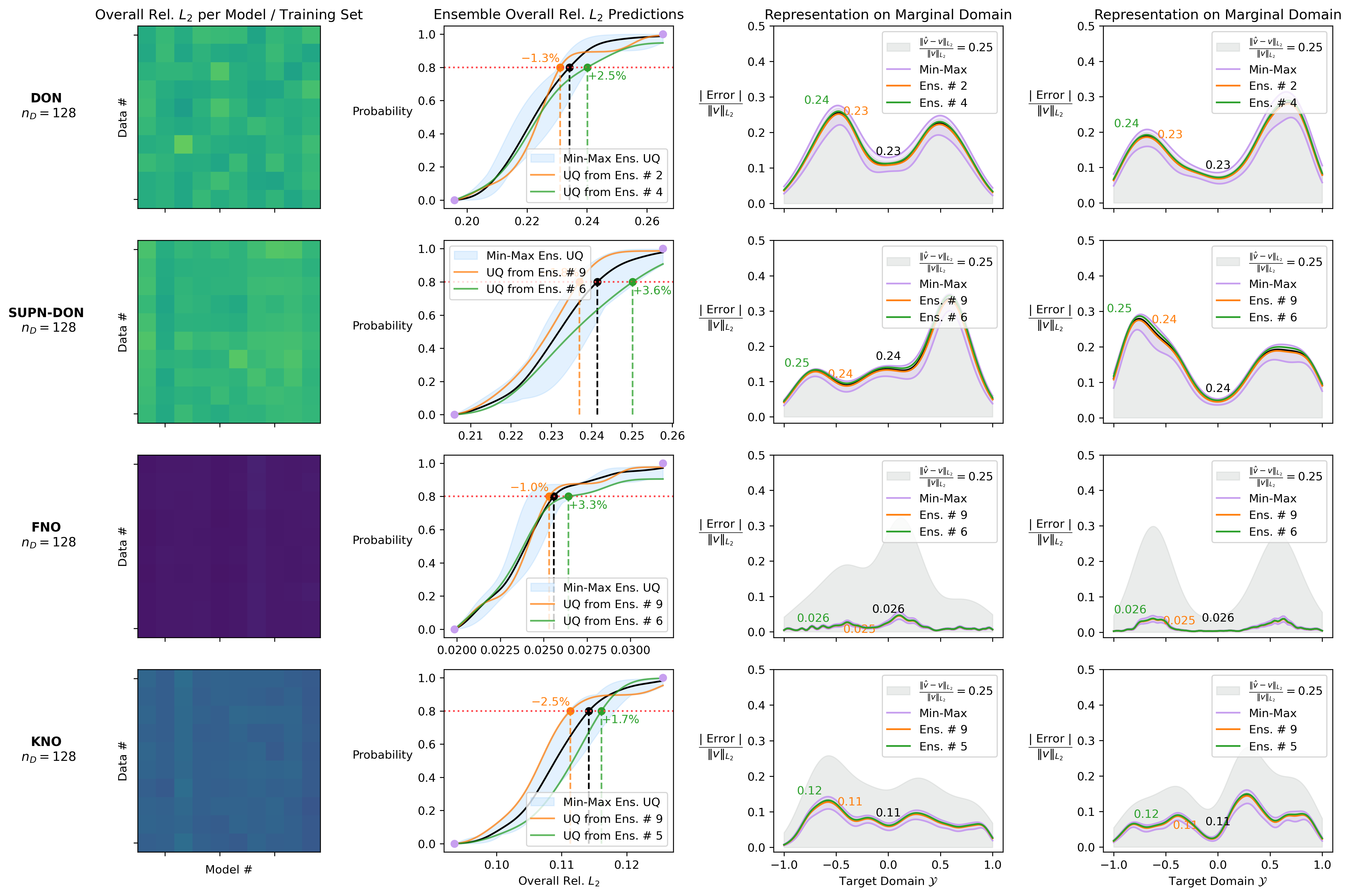}
	\caption{Ensemble estimates of the overall relative $L_2$ error distribution for the Allen--Cahn problem with $n_D=128$.}
    \label{fig:acmod_uqover_128}
\end{figure}

\subsection{Ensemble UQ for Relative $L_2$ on a Specific Example}
For a representative Allen--Cahn test example, we compare
the empirical cumulative distribution function (CDF) of
the relative $L_2$ error across all trained models with
CDFs estimated from initialization ensembles trained on
individual datasets. The left column shows the spatial
distribution of the absolute prediction error, normalized
by the $L_2$ norm of the reference solution. The shaded
region in the CDF plots represents the range of ensemble
estimates, while the selected curves illustrate differences
in estimates of the 80\% error quantile. The right columns
visualize these quantile estimates relative to a benchmark
relative $L_2$ error of 0.25, using spatially marginalized
error distributions.

The performance of Ensemble UQ for the Relative $L_2$ on a specific example was much more varied, as described by Figure~\ref{fig:acmodqauntilequality}.
For the specific example shown in Figures~\ref{fig:acmod_uqSPL2-ld} and \ref{fig:acmod_uqSPL2-hd}, we present the ensemble quantile plots for the other discretizations for comparison.
We see that ensembles of KNOs provide unreliable uncertainty estimates at $n_D=8$ and $n_D=32$ and decent results for $n_D=16$ and $n_D=64$.
Ensembles of DONs and SUPN DONs on the other hand, provide decent UQ for this example at $n_D=8$ and $n_D=16$, but the results for $n_D>16$ is increasingly inaccurate.
Note that since this example was randomly selected, it may not be representative of all samples in the test set.
However, it is a concerning feature of the ensemble UQ approach that for this example, the behavior of 80\%-quantile predictions for the relative $L_2$ is not consistent across samples and discretizations.

\begin{figure}[H]
    \centering
    \captionsetup{width=0.9\textwidth} \includegraphics[width=0.9\linewidth,keepaspectratio]{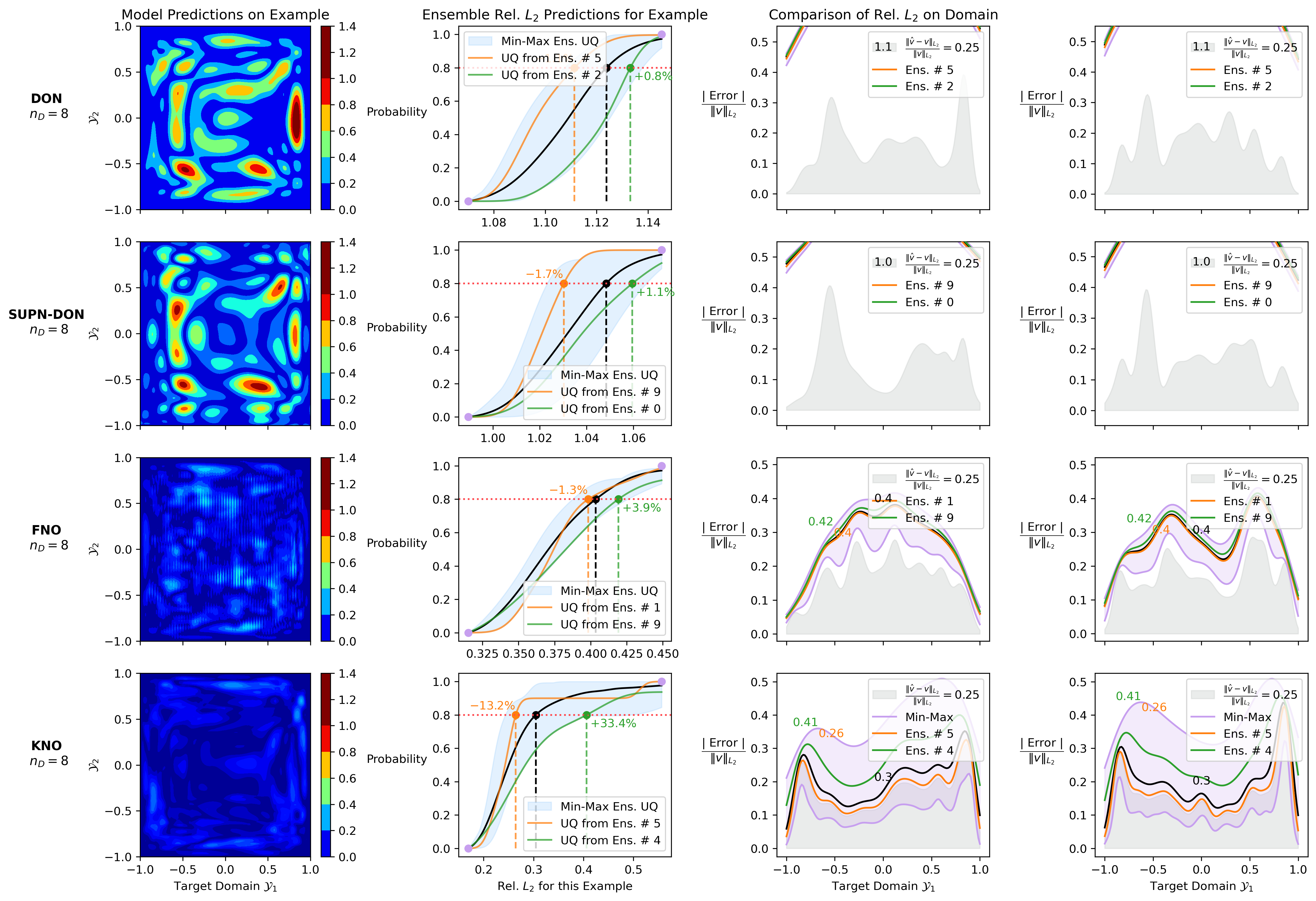}
    \caption{Ensemble estimates of the relative $L_2$ error
distribution for a representative Allen--Cahn test example
with $n_D=8$.}
    \label{fig:acmod_uqSPL2-8}
\end{figure}

\begin{figure}[H]
    \centering
    \captionsetup{width=0.9\textwidth} \includegraphics[width=0.9\linewidth,keepaspectratio]{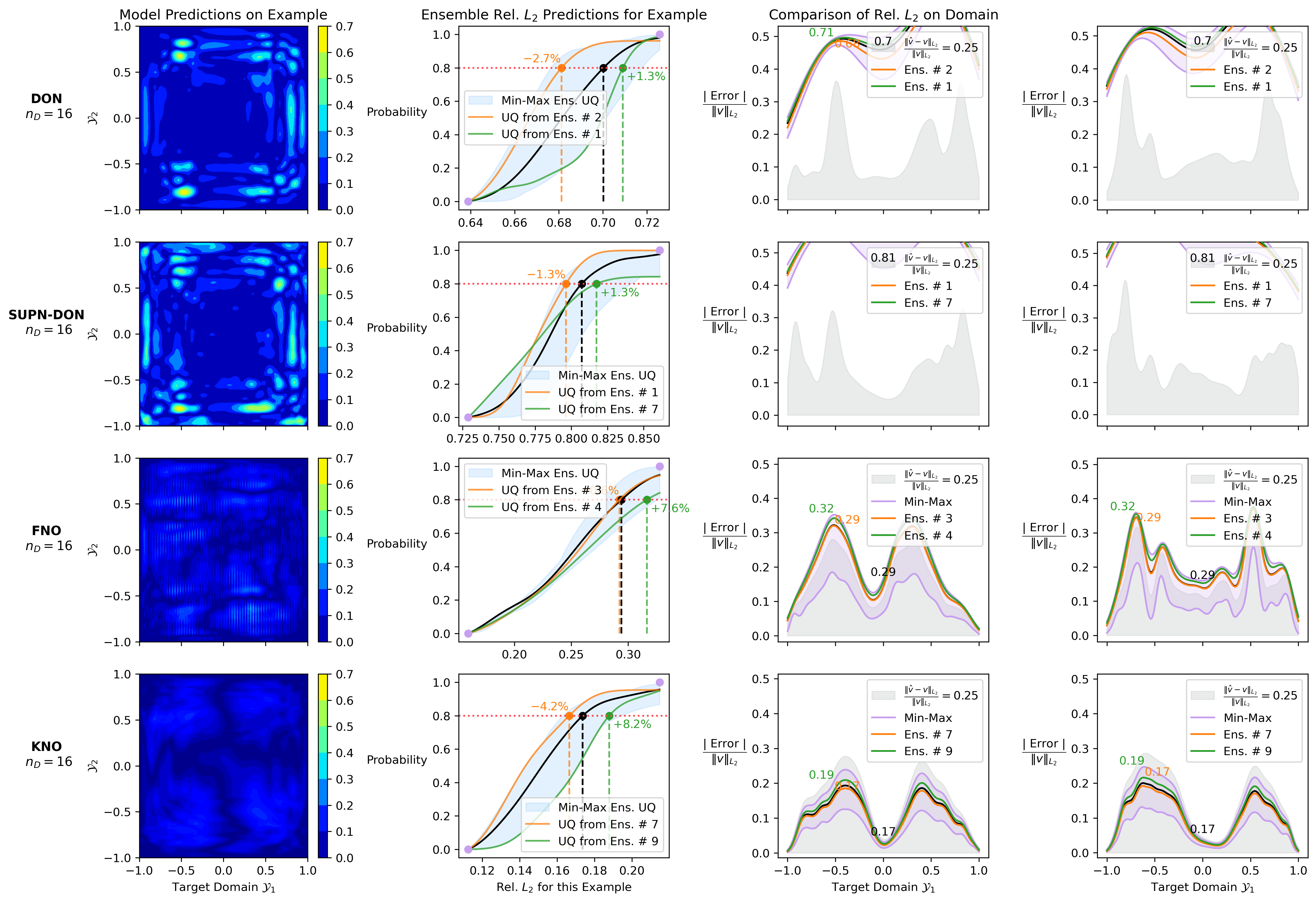}
\caption{Ensemble estimates of the relative $L_2$ error
distribution for a representative Allen--Cahn test example
with $n_D=16$.}
    \label{fig:acmod_uqSPL2-16}
\end{figure}

\begin{figure}[H]
    \centering
    \captionsetup{width=0.9\textwidth} \includegraphics[width=0.9\linewidth,keepaspectratio]{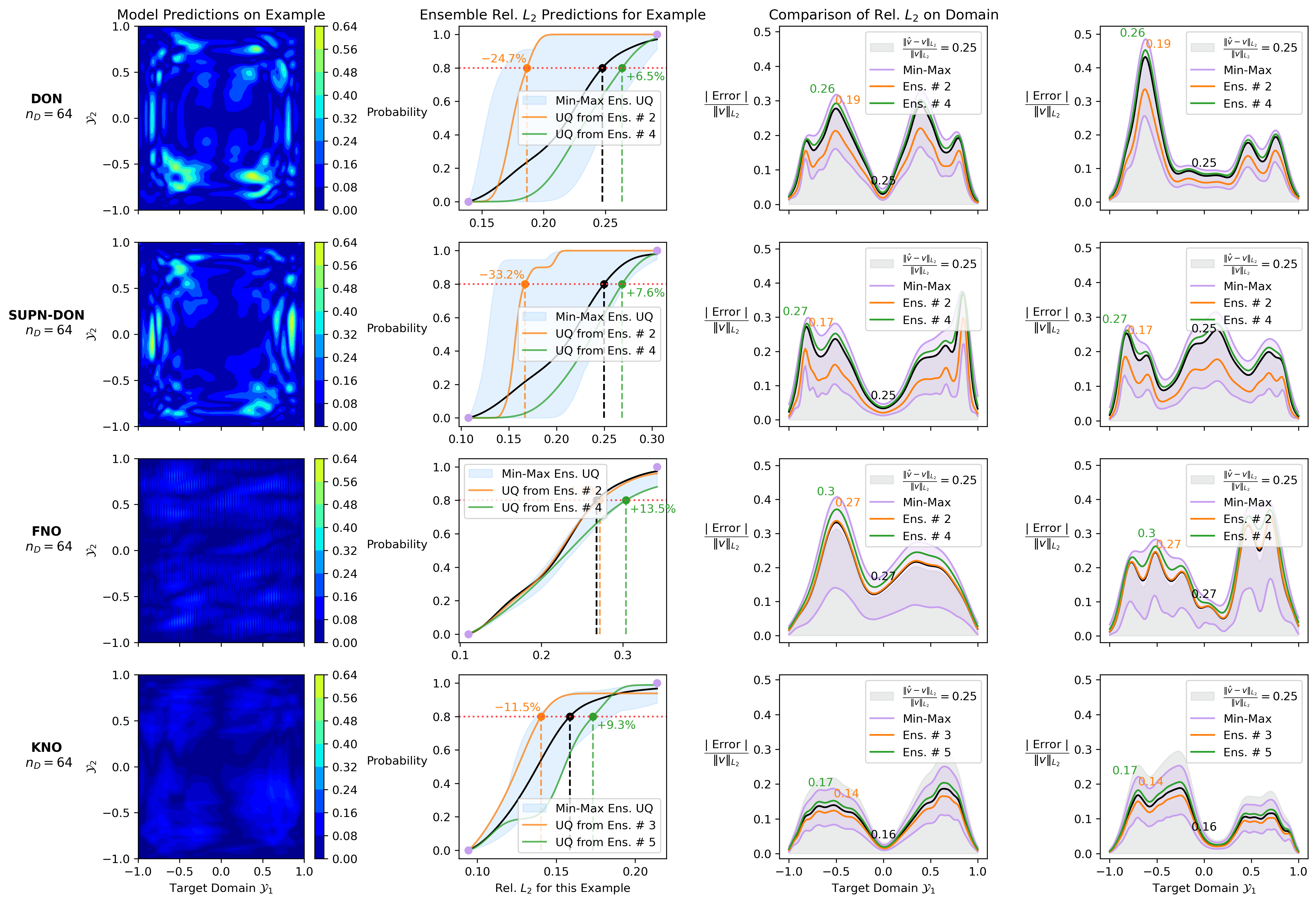}
\caption{Ensemble estimates of the relative $L_2$ error
distribution for a representative Allen--Cahn test example
with $n_D=64$.}
    \label{fig:acmod_uqSPL2-64}
\end{figure}

\end{document}